\documentclass{article}
\usepackage[final, nonatbib]{meta_album_2022}
\usepackage[square,numbers]{natbib}
\usepackage[utf8]{inputenc} 
\usepackage[T1]{fontenc}    
\usepackage{url}            
\usepackage{booktabs}       
\usepackage{amsfonts}       
\usepackage{nicefrac}       
\usepackage{microtype}      
\usepackage{xcolor}         
\usepackage{amsmath} 
\usepackage{makecell}
\usepackage{adjustbox}
\usepackage{caption}
\usepackage{tabularx}
\usepackage{subcaption}
\usepackage{xr-hyper}
\usepackage{hyperref}       
\hypersetup{
  colorlinks   = true, 
  urlcolor     = purple, 
  linkcolor    = blue, 
  citecolor    = red 
}

\newcommand{\eg}{{\em e.g.,~}}
\newcommand{\ie}{{\em i.e.,~}}

\usepackage{graphics, color, graphicx, wrapfig}
\graphicspath{ {./fig/} }
\usepackage{float}
\usepackage{todonotes}
\usepackage{comment}
\usepackage{multirow}

\usepackage{pifont}

\usepackage{xspace}
\newcommand{\mathsymbol}[2]{\newcommand{#1}{\ensuremath{\mathit{#2}}\xspace}}

\mathsymbol{\dataspace}{\mathcal{D}}
\mathsymbol{\dataset}{D}
\mathsymbol{\Tau}{\mathcal{T}}
\mathsymbol{\loss}{\mathcal{L}}
\mathsymbol{\datasethalf}{\dataset_\mathit{half}}
\mathsymbol{\datasetfull}{\dataset_\mathit{full}}
\mathsymbol{\instancespace}{\mathcal{X}}
\mathsymbol{\labelspace}{\mathcal{Y}}
\mathsymbol{\hypospace}{\mathcal{H}}
\mathsymbol{\learner}{\mathcal{A}}
\mathsymbol{\metalearner}{\learner_{meta}}
\mathsymbol{\metadataset}{\dataset_{meta}}
\mathsymbol{\risk}{\mathcal{R}}
\mathsymbol{\metarisk}{\risk_{meta}}
\mathsymbol{\benchmark}{\mathcal{B}}

\newcommand{\cmark}{\ding{51}}%
\newcommand{\xmark}{\ding{55}}%
\newcommand*\rot{\rotatebox{90}}

\title{Meta-Album: Multi-domain Meta-Dataset for \\ Few-Shot Image Classification}

\author{%
Ihsan Ullah\textsuperscript{*}, 
Dustin Carrión-Ojeda\textsuperscript{*¶‖},
Sergio Escalera\textsuperscript{\#\%},
Isabelle Guyon\textsuperscript{*\%}, 
{\bf Mike Huisman}\textsuperscript{+}, \\
{\bf Felix Mohr}\textsuperscript{‡},
{\bf Jan N. van Rijn}\textsuperscript{+},
{\bf Haozhe Sun}\textsuperscript{*},
{\bf Joaquin Vanschoren}\textsuperscript{§},
{\bf Phan Anh Vu}\textsuperscript{*}\\
\texttt{\%} ChaLearn, USA\\
\texttt{‖} hessian.AI, Germany\\
\texttt{\#} Universitat de Barcelona, Spain\\
\texttt{‡} Universidad de La Sabana, Colombia\\
\texttt{¶} Technische Universität Darmstadt, Germany\\
\texttt{*} LISN/CNRS/INRIA, Université Paris-Saclay, France\\
\texttt{§} TU/e Eindhoven University of Technology, The Netherlands\\
\texttt{+} Leiden Institute of Advanced Computer Science (LIACS), Leiden University, the Netherlands\\
\texttt{\url{https://meta-album.github.io/}
}
}

\begin{document}

\maketitle
%
%

\begin{abstract}

We introduce Meta-Album, an image classification meta-dataset designed  to facilitate few-shot learning, transfer learning, meta-learning, among other tasks. It includes 40 open datasets, each having at least 20 classes with 40 examples per class, with verified licences. They stem from diverse domains, such as ecology (fauna and flora), manufacturing (textures, vehicles), human actions, and optical character recognition, featuring various image scales (microscopic, human scales, remote sensing). All datasets are preprocessed, annotated, and formatted uniformly, and come in 3 versions (Micro $\subset$ Mini $\subset$ Extended) to match users' computational resources. We showcase the utility of the first 30 datasets on few-shot learning problems. The other 10 will be released shortly after. 
Meta-Album is already more diverse and larger (in number of datasets) than similar efforts, and we are committed to keep enlarging it via a series of competitions. As competitions terminate, their test data are released, thus creating a rolling benchmark, available through OpenML.org. Our website \url{https://meta-album.github.io/} contains the source code of challenge winning methods, baseline methods, data loaders, and instructions for contributing either new datasets or algorithms to our expandable meta-dataset.\footnote{All authors except for the first two authors (equal contributions) are in alphabetical order of last name.}

\begin{figure}[H]
    \centering
    \includegraphics[width=\textwidth]{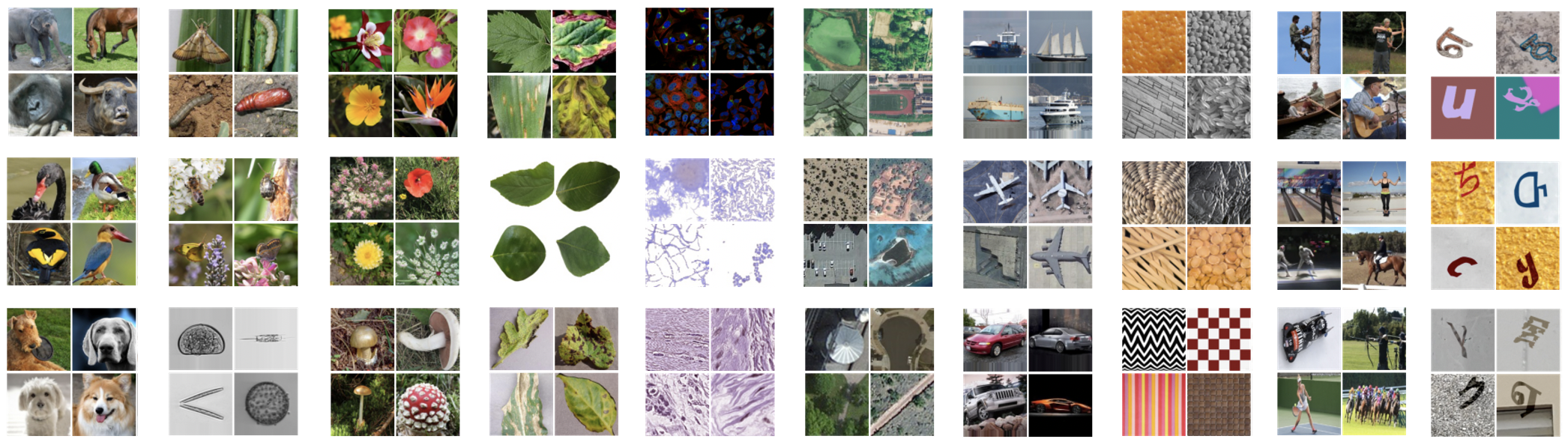}
    \caption{Meta-Album sample images. Each column represents one domain and each row one set. Domains are arranged in the same order as in Table~\ref{tab:datasets_summary_public}.}
    \label{fig:sample_images}
\end{figure}

\end{abstract}


%
%

\section{Introduction}
\label{sec:introduction}

\subsection{Background}
\label{sec:background}

Machine learning has progressed rapidly in recent years and has enabled breakthroughs in various domains. 
The success of most machine learning techniques hinges on the availability of large amounts of data~\cite{lecun2015deep,sun2017revisiting}, limiting their applicability in domains where only little data is available.
Enabling machine learning algorithms to learn new tasks from only a few examples is studied within the field of \emph{few-shot learning}~\cite{naik1992meta, schmidhuber1987evolutionary, thrun1998lifelong}.
Novel meta-learning algorithms have recently been proposed targeting few-shot learning, triggering a surge of popularity for such problems~\cite{brazdil2022metalearning,hospedales2020meta,huisman2021survey,wang2020fewshotsurvey}.
Despite the popularity of the field, progress is held back by a lack of good, challenging, and computationally feasible meta-datasets, that enable us to accurately assess the generalization abilities of few-shot learning algorithms. To remedy this, we introduce
\textbf{Meta-Album} (Figure~\ref{fig:sample_images}), an extensible {\bf multi-domain meta-dataset}, including (so far) 40 image classification datasets from 10 different domains: 30 of these are currently available through our website, and the remaining 10 will be released in spring 2023. This is part of a long-term effort to create a publicly available and growing meta-dataset, in conjunction with a meta-learning challenge series (the 2021 and 2022 editions are both parts of the NeurIPS competition track~\cite{carrion2022neurips,elbaz2021metadl,elbaz2022lessons}). As competitions terminate, older datasets get released, thus refreshing a rolling benchmark, made available through OpenML.org~\cite{openml}.  We check that all datasets are free for use in academic research and provide their original licenses.

Meta-Album was specifically designed to facilitate meta-learning research in the cross-domain few-shot setting, which is more realistic than commonly used evaluation protocols.
Traditionally, few-shot learning algorithms (\eg~\cite{finn2017model,huisman2022stateless,snell2017protonets}) have been evaluated by taking an existing benchmark dataset from a particular ``domain'' (\eg handwriting recognition) with a large number of classes, and then breaking it down into smaller classification {\bf tasks}, each including a {\em random} subset of classes (\eg a few specific characters). Algorithms are then tested for their ability to solve such tasks ``quickly'' from a small number of examples, after being trained on many other tasks. Typically the number of classes $N$ and examples per class $k$ are both {\em fixed} in what is known as an \textit{N-way k-shot} learning problem. While this setting has served research well, it is not very representative of practical real-world applications where tasks may come from various domains, include classes not drawn at random, but stemming from a class hierarchy, and include any number of classes and/or examples per class. By providing data from a wide variety of domains, including datasets with many classes and a minimum number of examples per class, and retaining class hierarchy annotations, Meta-Album
enables benchmarking according to a variety of more realistic settings.

\subsection{Related work}
\label{sec:related}

In this section, we review meta-datasets previously proposed to benchmark few-shot learning and meta-learning, as well as large-scale multi-class datasets, and then contrast them with Meta-Album.

\textbf{Single dataset benchmarks}: 
Omniglot~\cite{omniglot} is often used as a starting benchmark for few-shot learning and meta-learning. MiniImageNet~\cite{matchingnetwork} and Tiered-ImageNet~\cite{metasemisupervised} are adapted for few-shot image classification from ImageNet~\cite{imagenet}. CIFAR-FS~\cite{metadifferentiable} and FC100~\cite{tadam} are remodeled from CIFAR-100~\cite{cifar} for few-shot settings.

\textbf{Multi-dataset benchmarks}: 
A recent trend is to assemble numerous datasets from different domains in the same benchmark.
Visual Decathlon~\cite{visualdecathlon} gathers 10 diverse datasets. The focus is on finding a model with a universal representation capacity for use in many tasks.
VTAB (Visual Task Adaptation Benchmark)~\cite{vtab} assembles 19 image classification tasks across various domains. These tasks are grouped into 3 partitions: natural, specialized, and structured.
Meta-Dataset~\cite{metadataset} includes 10 image classification datasets from several application domains in one collection. Meta-Dataset also leverages the label hierarchy in ImageNet and Omniglot to organize the tasks.

\textbf{Transfer learning and meta-learning benchmarks}: 
VTAB + MD~\cite{comparetransfermeta} attempts to unify common transfer and meta-learning datasets in a single benchmark. The authors also provide a comparison of popular meta- and transfer learning methods.
BSCD-FSL (Broader Study of Cross-Domain Few-Shot Learning)~\cite{crossdomainfewshot} gathers 4 real-world tasks to compare few-shot, meta-learning, and transfer learning methods. WILDS~\cite{pmlr-v139-koh21a} is a benchmark of 10 datasets of various modalities (images, graphs, and text; 6 of them are image datasets), reflecting a diverse range of distribution shifts that naturally arise in real-world applications, and hence useful to evaluate meta-learning and transfer learning techniques.  CTrL is a continual transfer-learning benchmark~\cite{VeniatDR21} including 7 commonly used datasets in image classification. In reinforcement learning, sets of simulation environments exist for meta- and transfer learning, such as Meta-World~\cite{yu2021metaworld}.

\textbf{Outside few-shot and meta-learning}: The AutoDL challenge~\cite{autodl} features a series of 66 datasets from numerous domains. These datasets cover a wide range of modalities: image, video, audio, text, and tabular. This competition focuses on finding a universal algorithm, which can solve many tasks without human supervision.

\begin{table}[t]
  \caption{Comparison between Meta-Album and other large-scale or (meta-) datasets
  }
 
  \label{tab:datasets_comparison}
  \centering
  \begin{adjustbox}{width=\linewidth}
  \begin{tabular}{l r r r r r l c c c c c}
    \toprule
    \makecell{Dataset/\\Meta-Dataset}  
    & \rot{\# of domains} & \rot{\# of datasets}  
    & \rot{\# of images}  & \rot{\makecell{min/max classes\\per domain}}  
    & \rot{\makecell{min/max images\\per class}} & \rot{size on disk} 
    & \rot{multi-domain} & \rot{\makecell{lightweight\\($<$20GB)}} 
    & \rot{\makecell{uniform \# of\\images per class}} 
    & \rot{\makecell{uniform\\image size}} & \rot{repeated extensions} 
    \\
    \midrule
    
    Meta-Dataset    & 7 & 10 & $53\,068\,000$       
    & 43/$1\,696$ & 3/$140\,000$ & 210 GB & \cmark & \xmark & \xmark & \xmark  & \xmark
    \\
    VTAB    & 3 & 19 & $2\,244\,000$     & 2/397       
    & 40/$1\,000$ & 100 GB & \cmark & \xmark & \xmark & \xmark & \xmark
    \\
    MS-COCO & 1 & 1 & $328\,000$ & 80/80 & 9/$10\,777$ & 44 GB
    & \xmark & \xmark & \xmark & \xmark & \xmark
    \\
    Mini Imagenet   & 1 & 1 & $60\,000$ & 100/100 & 600/600   
    & 1 GB & \xmark & \cmark & \cmark & \cmark & \xmark
    \\
    Omniglot    & 1 & 1 & $32\,000$    & $1\,623/1\,623$ & 20/20      
    & 148 MB & \xmark & \cmark & \cmark & \cmark & \xmark
    \\
    CUB-200  & 1 & 1 & $6\,000$         & 200/200 & 20/39    
    & 647 MB & \xmark & \cmark & \xmark & \xmark & \xmark
    \\
    CIFAR-100   & 3 & 1 & $60\,000$        & 15/50 & 600/600   
    & 161 MB & \xmark & \cmark & \cmark & \cmark & \xmark
    \\
    
    \midrule
    \textbf{Meta-Album {\it Micro}} & \textbf{10} & \textbf{40} & \textbf{32\,000} & \textbf{19/20} & \textbf{40/40} 
    & \textbf{380 MB} & \cmark & \cmark & \cmark & \cmark & \cmark
    \\
    \textbf{Meta-Album {\it Mini}} & \textbf{10} & \textbf{40} & \textbf{220\,950} & \textbf{19/706} & \textbf{40/40} 
    & \textbf{3.9 GB} & \cmark & \cmark & \cmark & \cmark & \cmark
    \\
    \textbf{Meta-Album {\it Extended}} & \textbf{10} & \textbf{40} & \textbf{1\,583\,624} & \textbf{19/706} & \textbf{1/187\,384} 
    & \textbf{15 GB} & \cmark & \cmark & \xmark & \cmark & \cmark
    \\
    \bottomrule
    
  \end{tabular}
  \end{adjustbox}
\end{table}

We compare Meta-Album with previous benchmarks/datasets in Table~\ref{tab:datasets_comparison}, and provide further details in Appendix~\ref{appendix:benchmark_comparison}. 
Meta-Album covers a variety of domains, including ecology, manufacturing, textures, object classification, and character recognition, as well as a variety of scales: microscopic, macroscopic (human scale), or distant (remote sensing). While mostly re-purposing public datasets from heterogeneous sources to maximally vary recording conditions, we also introduce a few fresh datasets in OCR and ecology domains. Meta-Album comprises 3 different versions, Micro $\subset$ Mini $\subset$ Extended: {\bf Micro} includes 20 classes and 40 images per class for ease of running sample code, {\bf Mini} retains all original classes but also includes only 40 examples per class, while {\bf Extended} includes all classes and examples.

The variety of versions positions Meta-Album anywhere amongst small-scale datasets such as Omniglot~\cite{omniglot}, miniImageNet~\cite{matchingnetwork, ravi2017optimization} and CUB~\cite{cub}, which usually have at most 70\,000 images in total and weigh at most a few GB, or very large-scale benchmarks such as Meta-dataset~\cite{metadataset} and VTAB~\cite{comparetransfermeta}, which have more than 50 million images, weigh at least a few hundreds GB, and require high-end super-computer clusters.
Its principal distinguishing feature is that it has, by far, the {\bf largest number of domains and datasets}, collected in different conditions, and that it is designed to be {\bf continually extended by either adding new domains or new datasets} in already existing domains, making it a tool of choice for cross-domain, domain-independent, and continual learning studies. 
Secondly, while other benchmarks usually provide only raw data, we {\bf format all images uniformly as $128\times 128$ pixel maps}, which has two benefits: reducing the storage/memory footprint and facilitating the benchmarking of methods independent of preprocessing steps.
To that end, we optimized cropping and resizing to reduce dimensions as much as possible without degrading performance too much. 
In addition, Meta-Album includes datasets that have a {\bf large number of classes} and class hierarchy annotations when available, with a {\bf minimum number of classes and examples per class}: at least 20 classes (except one dataset having only 19 classes) with a minimum of 40 examples per class. This facilitates benchmark design, allowing us to vary the number of classes and the number of training examples per class over a large range of values. 
Finally and importantly, we selected datasets that are {\bf not typically used in transfer-learning or meta-learning benchmarks}, \eg for pre-training backbone networks, such as ImageNet (which is included in \eg Meta-Dataset), or for conducting other meta-learning or transfer-learning experiments, such as Omniglot, CIFAR-100, SVHN, or MNIST (which are included in \eg VTAB and CTrL). This avoids giving an unfair advantage to methods that were developed using such commonly used datasets.

\subsection{Contributions and recommended use}
\label{sec:contributions}

In summary, the contributions of our work are the following. 
\begin{itemize}
    \item We provide a {\bf new meta-dataset for few-shot learning and meta-learning} consisting of 40 uniformly formatted datasets from 10 domains, which facilitates research in cross-domain meta-learning as well as practical and realistic evaluation of few-shot algorithms. 
    \item We provide 3 versions of each dataset: Micro, Mini, and Extended to  {\bf facilitate usage by researchers with access to different amounts of computational power}.
    
    \item We {\bf uniformly preprocessed and formatted data}, but also provide {\bf instructions to retrieve the corresponding raw data} on our aforementioned website.

    \item We stimulate {\bf community-driven benchmarking}, in conjunction with our challenge series, by welcoming new contributors and providing software and instructions to create additional datasets for Meta-Album, with strict quality control and review processes.
    \item  We showcase our new meta-dataset by performing an {\bf experimental evaluation} for several use cases, including transfer learning, few-shot meta-learning, and cross-domain few-shot meta-learning tasks, using a variety of algorithms, and we {\bf open-source the code used}.
\end{itemize}

The recommended use of Meta-Album is to conduct fundamental research on machine learning algorithms and perform benchmarks, particularly in few-shot learning, meta-learning, continual learning, transfer learning, and image classification. Meta-Album is not recommended to create products, whether commercial or not, or to derive scientific findings outside benchmarking. 

%
%
\section{Meta-Album design and initial release}
\label{sec:design_and_datasets}

In this section, we explain the motivations behind the design of Meta-Album and present the 30 datasets included in the initial NeurIPS 2022 release. 10 more datasets are kept private, and will be released in spring 2023.

\subsection{Motivation}\label{sec:motivation}

Meta-Album emerged from a sequence of few-shot meta-learning benchmarks, following the problem formulations described in Section \ref{sec:problem}. The first of these was the 2020 MetaDL-mini challenge, which was run in conjunction with AAAI 2021~\cite{elbaz2021metadl}. It followed the ``within domain few-shot learning'' protocol, and algorithms were evaluated with small-scale public datasets (Omniglot and CIFAR-100). Subsequently, we designed a first version of Meta-Album, including 15 datasets, for a larger-scale ``within domain few-shot learning'' challenge (MetaDL @ NeurIPS 2021~\cite{elbaz2022lessons}). Here, algorithms were meta-trained and meta-tested on tasks extracted from a single dataset at a time, and performances were averaged over 5 datasets, both in the feedback phase and the final evaluation phase, to obtain a more robust evaluation. The 5 extra datasets were provided for practice purposes. The results of this challenge (further detailed in Section \ref{sec:use_cases}) indicated that these tasks were well within reach of state-of-the-art methods. This motivated us to move to the ``cross-domain few-shot learning'' setting. The design of this new challenge (part of an official NeurIPS 2022 challenge~\cite{carrion2022neurips}) motivated us to grow Meta-Album to 30 datasets spanning multiple domains. We intend to continue growing Meta-Album and already have 10 more datasets lined up, in preparation for the next challenge. This will constitute a {\em rolling benchmark}: with each new challenge, previous feedback datasets are publicly released, previous final evaluation datasets become feedback datasets, and fresh datasets become final evaluation datasets.

Existing meta-datasets did not allow us to carry out our challenge program for several reasons: (1)~they included datasets too familiar to the meta-learning community; (2)~they did not include enough datasets to robustly evaluate participants (particularly in the cross-domain setting); (3)~their datasets had a large variance in number of classes and examples per class, introducing bias in our experimental design. This required us to source new datasets. Furthermore, since these challenges include code submission, and providing the same resources to all participants, we needed to limit computational resources. Therefore, we had to downscale images while taking care that this does not significantly degrade performance.

\subsection{Data search}\label{sec:search}

Many people were involved in the sourcing of all datasets and their preparation, and they are gratefully acknowledged in our acknowledgements. This collaboration followed precise instructions to identify datasets that are: (i)~from the same domain; (ii)~freely available for academic research; (iii)~having at least 20 classes with at least 40 examples per class; (iv)~with images of good enough quality by visual inspection and with no offensive material (we excluded ``deprecated'' datasets); (v)~with baseline performance within a given range.

The last criterion was needed to ensure the success of our challenges, since tasks that are too easy or too hard do not allow us to separate challenge participants. 
For the purpose of designing Meta-Album, we defined a ``domain'' according to four characteristics: (1)~application domain; (2)~pattern recognition problem (texture or object classification); (3)~scale: micro, human scale, or distant; (4)~input channels. 
We ended up with 10 domains (see Table~\ref{tab:datasets_summary_public}): Large animals, small animals, plants, plant diseases, microscopy, remote sensing, vehicles, manufacturing, human actions, and optical character recognition (OCR). Data sources were very varied, and mostly came from internet searches, but we also produced our own optical character recognition datasets and obtained novel donated data.

\subsection{Data preparation}\label{sec:preparation}

We performed several iterations of preprocessing, experiments, and analyses to prepare the datasets. This workflow included identifying and, when possible, correcting bias and artifacts (including artifacts we may have introduced by resizing and cropping images), and making sure that images are recognizable by human eye inspection. 

As we work with datasets from diverse sources, each dataset requires a different preprocessing strategy, \eg the small animals' datasets, plant-diseases datasets, manufacturing, and remote sensing datasets have images in different resolutions and orientations. However, usually, the object of interest lies in the middle of the image, which facilitated cropping images horizontally or vertically to get squared images. 
In some cases, \eg the plankton dataset, the image orientation depends on the shape of the plankton and the way it is photographed, \ie images have vertical or horizontal orientations based on the plankton in the image. 
Cropping images is not useful in this case because a big part of the plankton would be cropped. 
As such, we added a matching background to the images (either horizontally or vertically) by extending the top and bottom 3 rows or left and right columns respectively.
In order to make sure that we do not introduce artifacts in the data, afterwards we applied a Gaussian kernel of size (29, 29) using open-cv~\cite{opencv_library} to the newly constructed background.  
In other cases, the area of interest was not necessarily centered, \eg human action datasets, and we had to use a human face detector to locate the subject, and then we cropped the upper body.
For all datasets except for the optical character recognition datasets, we resized the images to a $128\times 128$ resolution using an anti-aliasing filter~\cite{opencv_library}. 
The optical character recognition datasets are synthetically generated directly to the correct dimension by OmniPrint~\cite{sun2021omniprint} (MIT license), and do not need further processing. 
The preprocessed data was formatted in a  \href{https://github.com/ihsanullah2131/meta-album/tree/master/DataFormat}{data format} conserving as much meta-data as possible.
For the micro and the mini version, only classes with at least 40 examples are kept for each dataset to maintain a balance between a large number of classes and sufficient examples per class while for the extended version, all classes and all images are kept. 
More details about data preparation and formatting can be found in Appendix~\ref{appendix:data_preparation}.

\begin{table}[t]
  \caption{Meta-Album: Datasets summary (\textit{Mini versions})
  }
  \label{tab:datasets_summary_public}
  \centering
  \begin{adjustbox}{width=\linewidth}
  \begin{tabular}{l l l l l r r l}
    \toprule
    {\bf Domain ID} & {\bf Domain Name} &  {\bf Set \#}  & {\bf Dataset ID} &
    {\bf Dataset Name} & {\bf \# Categories} & {\bf \# Images} &
    {\bf Original source}
    \\
    \midrule
    
    \multirow{3}{*}{{\it LR\_AM}} & \multirow{3}{*}{Large Animals} 
    
    & 0 & {\it BRD} & Birds & 315 & 12\,600 & Birds 400~\cite{birds}
    \\
    & & 1 & {\it DOG} & Dogs & 120 & 4\,800 & Stanford Dogs~\cite{dogs}
    \\
    & & 2 & {\it AWA} & Animals with Attributes & 50 & 2\,000 & AWA~\cite{awa}
    \\
    \midrule
    
    \multirow{3}{*}{{\it SM\_AM}} & \multirow{3}{*}{Small Animals} 
    & 0 &{\it PLK} & Plankton & 86 & 3\,440 & WHOI~\cite{whoiplankton}
    \\
    & & 1 & {\it INS\_2} & Insects 2 & 102 & 4\,080 & Pest Insects~\cite{Wu2019Insect}
    \\
    & & 2 & {\it INS} & Insects & 104 & 4\,160 & SPIPOLL~\cite{insects}
    \\

    \midrule
    
    \multirow{3}{*}{{\it PLT}} & \multirow{3}{*}{Plants} 
    & 0 & {\it FLW} & Flowers & 102 & 4\,080 & Flowers~\cite{Nilsback08}
    \\
    & & 1 & {\it PLT\_NET} & PlantNet & 25 & 1\,000 & PlantNet~\cite{plantnet}
    \\
    & & 2 & {\it FNG} & Fungi & 25 & 1\,000 &  Danish Fungi~\cite{danish-fungi}
    \\
    \midrule
    
    \multirow{3}{*}{{\it PLT\_DIS}} & \multirow{3}{*}{Plant Diseases} 
    & 0 & {\it PLT\_VIL} & PlantVillage & 38 & 1\,520 & PlantVillage~\cite{plantvillage, plantvillagedata}
    \\
    & & 1 & {\it MED\_LF} & Medicinal Leaf & 25 & 1\,000 & Medicial Leaf~\cite{medicinalleaf}
    \\
    & & 2 & {\it PLT\_DOC} & PlantDoc & 27 & 1\,080 & Plant Doc~\cite{plant-doc}
    \\
    \midrule
    
    \multirow{3}{*}{{\it MCR}} & \multirow{3}{*}{Microscopy} 
    & 0 & {\it BCT} & Bacteria & 33 & 1\,320 & DiBas~\cite{diBas}
    \\
    & & 1 & {\it PNU} & PanNuke & 19 & 760 & PanNuke~\cite{pannuke, pannuke-dataset}
    \\
    & & 2 & {\it PRT} & Subcel. Human Protein & 21 & 840 & Protein Atlas~\cite{thul2017subcellular} 
    \\
    
    \midrule
    
    \multirow{3}{*}{{\it REM\_SEN}} & \multirow{3}{*}{Remote Sensing} 
    & 0 & {\it RESISC} & RESISC & 45 & 1\,800 & RESISC45~\cite{resisc}
    \\
    & & 1 & {\it RSICB} & RSICB & 45 & 1\,800 & RSICB128~\cite{rsicb}
    \\
    & & 2 & {\it RSD} & RSD & 38 & 1\,520 & RSD46~\cite{rsd46classify, rsd46localize}
    \\
    \midrule
    
    \multirow{3}{*}{{\it VCL}} & \multirow{3}{*}{Vehicles} 
    & 0 & {\it CRS} & Cars & 196 & 7\,840 & Cars~\cite{cars}
    \\
    & & 1 & {\it APL} & Airplanes & 21 & 840 & Multi-type Aircraft~\cite{airplanes}
    \\
    & & 2 & {\it BTS} & Boats & 26 & 1\,040 & MARVEL~\cite{boats}
    \\
    \midrule
    
    \multirow{3}{*}{{\it MNF}} & \multirow{3}{*}{Manufacturing} 
    & 0 & {\it TEX} & Textures & 64 & 2\,560 & KTH-TIPS~\cite{Fritz2004THEKD, Mallikarjuna2006THEK2} Kylberg~\cite{Kylberg2011c} UIUC~\cite{lazebnik:inria-00548530}
    \\
    & & 1 & {\it TEX\_DTD} & Textures DTD & 47 & 1\,880 & Texture DTD~\cite{cimpoi2013describing}
    \\
    & & 2 & {\it TEX\_ALOT} & Textures ALOT & 250 & 10\,000 & Texture ALOT~\cite{texture_alot}
    \\
    \midrule

    \multirow{3}{*}{{\it HUM\_ACT}} & \multirow{3}{*}{Human Actions} 
    & 0 & {\it SPT} & 100 Sports & 73 & 2\,920 & 100 Sports~\cite{100-sports}
    \\
    & & 1 & {\it ACT\_40} & Stanford 40 Actions & 39 & 1\,560 & Stanford 40 Actions~\cite{act-40}
    \\
    & & 2 & {\it ACT\_410} & MPII Human Pose & 29 & 1\,160 & MPII Human Pose~\cite{act-410}
    \\

    \midrule
    
    \multirow{3}{*}{{\it OCR}} & \multirow{3}{*}{Optical Char. Recog.} 
    & 0 & {\it MD\_MIX} & OmniPrint-MD-mix & 706 & 28\,240 & 
    \\
    & & 1 & {\it MD\_5\_BIS} & OmniPrint-MD-5-bis & 706 & 28\,240 & OmniPrint~\cite{sun2021omniprint}
    \\
    & & 2 & {\it MD\_6} & OmniPrint-MD-6 & 703 & 28\,120 & 
    \\
    \bottomrule

  \end{tabular}
  \end{adjustbox}
\end{table}

\subsection{Initial Meta-Album release}

The initial release of Meta-Album consists of 3 datasets for each of the 10 domains. Each dataset has 3 versions controlling the size, as explained in Section~\ref{sec:related}.
All datasets are annotated with class labels and other meta-data.
All 30 datasets were chosen after careful and critical analysis during the data preparation and quality control steps as described in Appendix~\ref{appendix:data_preparation}. 
Table~\ref{tab:datasets_summary_public} provides statistics on the various versions; Figure~\ref{fig:sample_images}
shows sample images from each dataset. More details about datasets and their meta-data are listed in Appendix~\ref{appendix:datasets}. License information for all datasets can be found in Appendix~\ref{appendix:license}. Meta-Album datasets are being used in the \href{https://github.com/DustinCarrion/cd-metadl}{NeurIPS Cross-domain meta-learning Challenge 2022}. 
The first 30 datasets are available on \href{https://www.openml.org/}{OpenML}~\cite{openml}, and later in spring 2023, 10 more datasets will be released, followed by other releases as our challenge program unfolds. Details about how to access Meta-Album datasets, contribute to the open meta-dataset, prepare new datasets with quality control, and submit these datasets for inclusion in Meta-Album can be found on the \href{https://meta-album.github.io/}{Meta-Album Website}. 
This web page will also inform on software updates and revisions or new releases of our meta-dataset.

%
%
\section{Use cases and baselines}
\label{sec:use_cases}

This section illustrates how Meta-Album can be used for a variety of purposes. The code of all experiments is provided in our GitHub repository \url{https://github.com/ihsaan-ullah/meta-album}, and can serve as a basis to benchmark new algorithms against the baseline methods we investigate here.
The problems investigated range from few-shot learning (for which Meta-Album was designed) to multi-class image classification, transfer learning, hierarchical classification, and continual learning. Because of lack of space, we only report few-shot learning experiments.

\subsection{Problem setting}
\label{sec:problem}

\mathsymbol{\task}{\mathcal{T}}
\mathsymbol{\datatrain}{\dataset^{\mathit{meta\text{-}train}}}
\mathsymbol{\datatest}{\dataset^{\mathit{meta\text{-}test}}}
\mathsymbol{\querysetj}{\dataset^{\mathit{te}}_{\task_j}}
\mathsymbol{\supportsetj}{\dataset^{\mathit{tr}}_{\task_j}}
\mathsymbol{\queryset}{\dataset^{\mathit{te}}_{\task}}
\mathsymbol{\supportset}{\dataset^{\mathit{tr}}_{\task}}

In this paper, we focus on {\bf few-shot image classification}, where the goal is to {\bf learn to perform new classification tasks from a limited number of examples}.  
Here, every task $\mathcal{T}_j=( \mathcal{D}_{\mathcal{T}_j}^{train}, \mathcal{D}_{\mathcal{T}_j}^{test} )$ consists of a \textit{support set} $\mathcal{D}_{\mathcal{T}_j}^{train}$ with training examples and a \textit{query set} $\mathcal{D}_{\mathcal{T}_j}^{test}$ with test examples.\footnote{The nomenclature {\em support set} instead of {\em training set}, and {\em query set} instead of {\em test set} is common in the meta-learning literature. It highlights the fact that, while meta-training is done on {\em tasks = \{support set, query set\}}, no actual test-data is presented to the classifier. The meta-test data also includes pairs of support and query sets, from which the ground truth of query set samples is hidden from the classifier.}
In \textit{N}-way \textit{k}-shot classification, we require that every \textit{support set} contain exactly $N$ classes with $k$ examples per class ($kN = |\mathcal{D}_{\mathcal{T}_j}^{train}|$).
Another requirement is that the classes in the query set must occur in the support set.

Few-shot learning does not necessarily require meta-learning. As in other ``regular'' learning problems, a {\em learner}, having available a set of training examples $\mathcal{D}_{\mathcal{T}_j}^{train}$ for a given task, can just return a {\em trained model} (classifier). But meta-learning is frequently used to enhance few-shot learning. 

In a meta-learning problem, a {\em meta-learner}, having available a set of $m$ training {\em tasks} $\mathcal{M}_{\mathcal{D}}^{train} = \{{\mathcal{T}}_j\}_{j=1}^m$, returns a meta-trained {\em learner}. In order to develop a meta-trained few-shot {\em learner}, available data organized in tasks $\mathcal{M}_{\mathcal{D}}$ (coming either from one or multiple datasets) are split into three ``meta-splits'' containing {\em disjoint sets of classes}: \textit{meta-training} split $\mathcal{M}_{\mathcal{D}}^{train}$, \textit{meta-validation} split $\mathcal{M}_{\mathcal{D}}^{valid}$, and \textit{meta-testing} split $\mathcal{M}_{\mathcal{D}}^{test}$. The {\em learner} is meta-trained with $\mathcal{M}_{\mathcal{D}}^{train}$. During meta-training, the {\em learner} is evaluated with $\mathcal{M}_{\mathcal{D}}^{valid}$ every few meta-training cycles, to monitor progress. The final product of meta-training when the time budget has elapsed, is the {\em learner} with the highest performance on $\mathcal{M}_{\mathcal{D}}^{valid}$ tasks. It is then evaluated on tasks from $\mathcal{M}_{\mathcal{D}}^{test}$.

Within the realm of few-shot learning, we distinguish two cases. \textbf{Within domain few-shot learning} refers to the problem where data from the meta-validation and meta-test splits come from the same domain as meta-training data. 
Here, domain refers to one single dataset
of Meta-Album $\mathcal{D}_i, \;i\in\{1,\dots,30\}$.
We enforce that $\mathcal{D}_i$ is partitioned into $\mathcal{M}_{\mathcal{D}_i}^{train}$, $\mathcal{M}_{\mathcal{D}_i}^{valid}$, and $\mathcal{M}_{\mathcal{D}_i}^{test}$, using three disjoint sets of classes. In this setting, the goal of \emph{learners} is to learn tasks including classes coming from the same original domain/dataset. 
If the \emph{learner} has been meta-trained, \textbf{test tasks include new classes unseen during meta-training}.
\textbf{Cross-domain few-shot learning}, in contrast, is a setting for which meta-split is performed at {\em dataset level} instead of {\em class level}. 
Once the {\em learner} has been meta-trained, {\bf test tasks come from \underline{new datasets} unseen during meta-training}. 
Note that as a consequence, there is still a slight domain overlap between the meta-train, meta-validation and meta-test dataset. For example, the meta-train dataset can contain observations from ten different datasets, including the `Fungi' dataset, whereas the learner will be evaluated on a meta-test dataset constructed from ten different datasets, including the `Flowers' dataset. 
This introduces two important challenges for the meta-learning algorithms whenever confronted with a given task in the meta-test set:
1)~it has to deal with several classes in the meta-train set that are not related to the concepts from the task at hand.
2)~while there are indeed observations in the meta-train set that are related to concept of the current task, these come from a different dataset, and might be sampled according to different conditions (different camera, lightning, geographical area, etc.).
This aligns with the cross-domain setting introduced in the NeurIPS'22 meta-learning challenge~\cite{carrion2022neurips}.
Beyond the cross-domain setting, one can imagine a `domain independent' setting, where each of the meta-train, meta-validation and meta-test datasets contain classes from different domains, and therefore no domain knowledge from the meta-train phase can be exploited. 

We also distinguish between {\bf fixed N-way k-shot} evaluations and {\bf any-way any-shot} evaluations. The former requires fixing the value of $N$ and $k$ for the entire benchmark. The latter requires randomly choosing $N$ and $k$ for each task, within pre-defined ranges. Meta-Album allows us to choose $N \in [2, 20]$ and $k \in [1, 20]$.

\subsection{Experiments}
\label{sec:experiments}

The first motivational use of Meta-Album has been the NeurIPS 2021 MetaDL challenge~\cite{elbaz2022lessons}. This was a meta-learning challenge with code submission, aiming at evaluating {\bf few-shot learning methods in the \underline{within domain} setting}, as described in Section~\ref{sec:problem}. The evaluation was carried out with 600 tasks in the {\bf 5-way 5-shot setting}, using a subset of Meta-Album (see Table~\ref{tab:metadl2021}).

\begin{table}[tb!]
\caption{Datasets used in the NeurIPS 2021 MetaDL challenge~\cite{elbaz2022lessons}.\label{tab:metadl2021}}
\begin{tabularx}{\textwidth}{lX}
\toprule
Phase & Datasets according to Table~\ref{tab:datasets_summary_public}\\
\midrule
  Feedback Phase & SM\_AM.PLK,
MDN.MLD,
MNF.TEX\_DTD,
REM\_SEN.RSICB,
OCR.MD\_MIX\\
Final test phase &
SM\_AM.INS,
PLT\_DIS.PLT\_VIL,
MNF.TEX,
REM\_SEN.RESISC,
OCR.MD\_5\_BIS\\
\bottomrule
\end{tabularx}
\end{table}

The solutions of the top participants have been \href{https://metalearning.chalearn.org/metadlneurips2021}{open-sourced}.
In a paper, authored collaboratively between the competition organizers
and the top-ranked participants~\cite{elbaz2022lessons}, we analyse the results of the competition. The lessons learned include
that learning good representations is essential for effective transfer learning. The winner's solution MetaDelta++~\cite{chen2019closer}, based on a combination of pre-trained backbone networks, performed best on all final 5 test phase datasets, with high accuracy scores (0.98, 0.94, 0.99, 0.92, 0.94). This indicates that, in future challenges, we are ready to tackle harder tasks, and motivated us to move to {\bf cross-domain few-shot learning}, in the {\bf any-way any-shot setting} for the NeurIPS 2022 challenge~\cite{carrion2022neurips}.
Fine-tuning backbones on meta-training data turned out to be important, though
there are indications that off-the-shelf backbones pre-trained with self-supervised learning
on massive datasets might become the way of the future, essentially making meta-learning
unnecessary for image classification problems. Thus, meta-learning should be benchmarked
in {\bf de novo training conditions}, in the future, to prepare for scenarios (in other domains) in
which such backbones are not available. The NeurIPS 2022 challenge encourages {\em de novo} training in a dedicated league. Appendix~\ref{appendix:dataset-difficulty} contains a detailed analysis of the difficulty of all Meta-Album datasets following the NeurIPS 2021 MetaDL challenge protocol.

\subsubsection*{Difficulty of cross-domain few-shot learning}\label{sec:fewshotlearning}

To evaluate the gap in difficulty between ``within domain'' and ``cross-domain'' few-shot learning problems (Section~\ref{sec:problem}), we carried out first experiments in the 5-way [1, 5, 10, 20]-shot setting. For all experiments, we use Meta-Album Mini, single PNY GeForce RTX 2080TI GPUs with 11GB of VRAM or a single NVIDIA V100 with 16GB of VRAM. Each experimental run took at most 24 hours on the former GPU (for details, please see Appendix~\ref{appendix:fslappendix} and Appendix~\ref{appendix:cd-fs-additional}).

Although several methods have been proposed in the state-of-the-art to tackle the cross-domain few-shot learning problem~\cite{SUR2020,RDC2022,TSA2021,URL2021,URT2020,iMAML2019,FLUTE2021}, they require too much time or are not compatible with our fully supervised setting. Therefore, we investigated the few-shot learning performance of popular meta-learning methods: MAML~\cite{finn2017model}, Matching networks~\cite{matchingnetwork}, and Prototypical networks~\cite{snell2017protonets}.  
We compared them against two baseline methods: TrainFromScratch (learning every task starting from a random initialization at meta-test time, \ie no meta-learning) and FineTuning, which is pre-trained on the classification problem arising from concatenating all meta-training classes and corresponding data and only fine-tunes the last layer at meta-test time~\cite{chen2019closer}.
All techniques use a ResNet-18 backbone~\cite{resnet} and are trained from scratch on Meta-Album (not using any pre-trained feature extractors) using the best-reported hyperparameters by the original authors on $5$-way $5$-shot miniImageNet (\ie for FineTuning the backbone is pre-trained with Meta-Album meta-training data only). It is worth mentioning that the purpose of our baseline methods is to give a set of ``classical'' and ``representative'' techniques, not to be exhaustive.

For a given dataset, all meta-learning techniques are meta-trained on $60\,000$ tasks. However, the backbone used for FineTuning is meta-trained (pre-trained) on $60\,000$ randomly sampled batches of size 16.
The performance of trainers is validated every $2\,500$ tasks (or batches in case of pre-training the FineTuning backbone). 
The query set for every task contains 16 examples per class, following~\cite{chen2019closer}.
The learning algorithm with the best validation performance is evaluated on $600$ meta-test tasks randomly sampled from the meta-testing split, which has information from unseen classes during training and validation. We average the results over 3 runs with different random seeds. Error bars are $95\%$ confidence intervals of the mean overall meta-test tasks in all runs ($1\,800$ tasks per dataset).

\begin{figure}
     \centering
     \begin{subfigure}{0.44\textwidth}
         \centering
         \includegraphics[width=\textwidth]{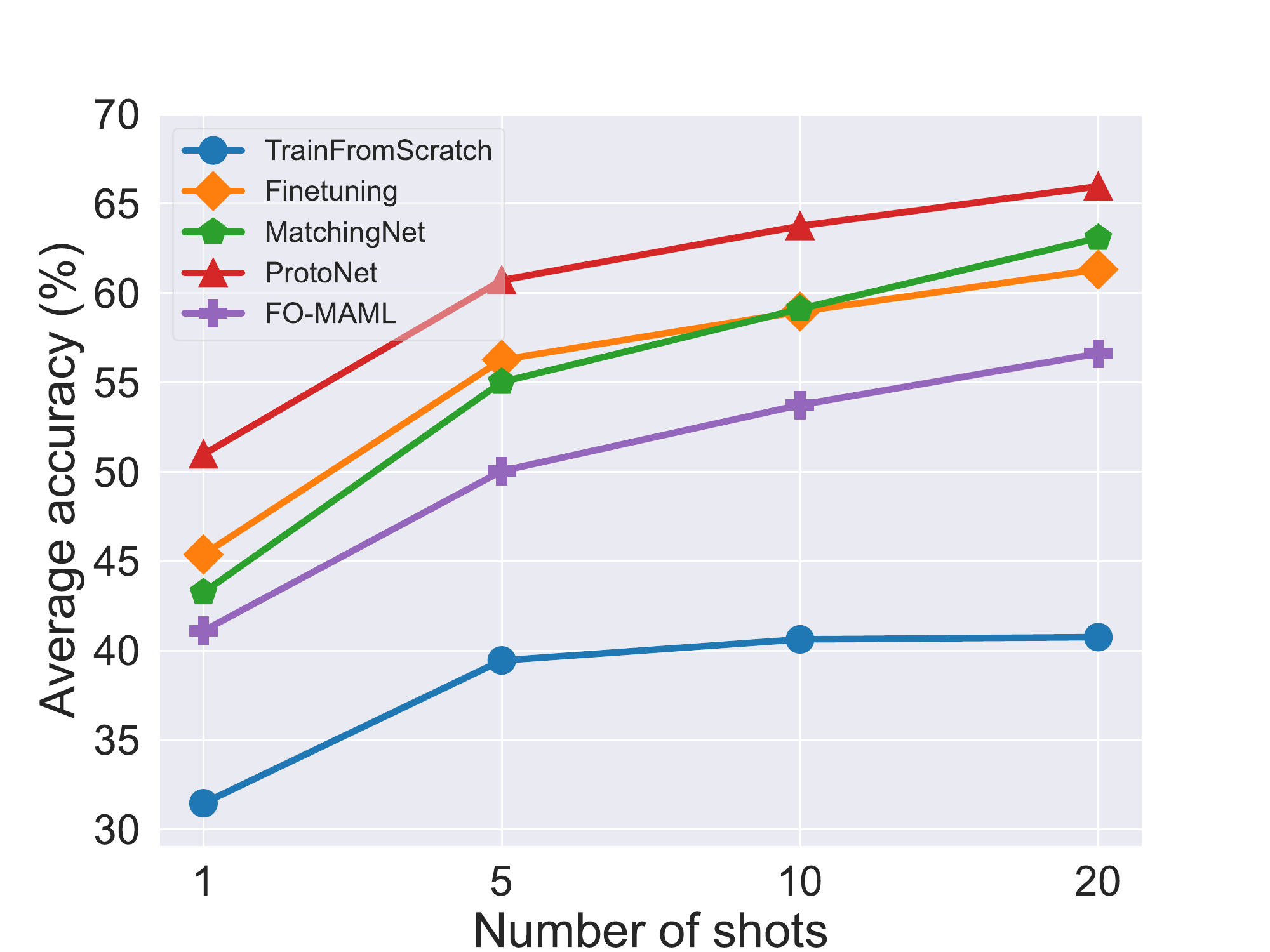}
         \caption{Within domain}
     \end{subfigure}
     \hfill
     \begin{subfigure}{0.44\textwidth}
         \centering
         \includegraphics[width=\textwidth]{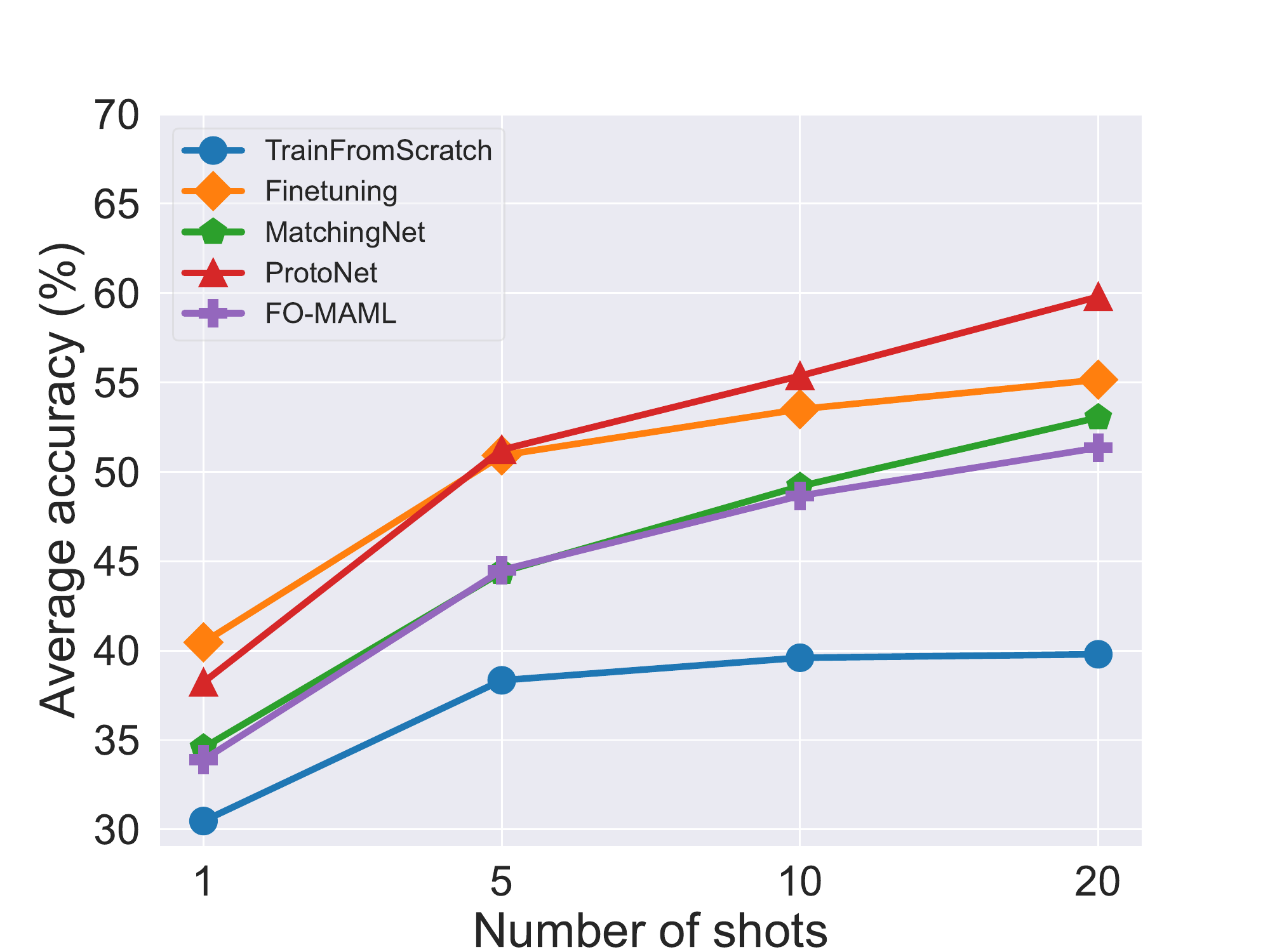}
         \caption{Cross-domain}
     \end{subfigure}
     \caption{\label{fig:fsl-plots} {\bf Comparison of ``within domain'' and ``cross-domain'' few-shot learning.} We plot 5-way [1, 5, 10, 20]-shot learning meta-test mean task accuracy, averaged over $1\,800$ tasks drawn from the 30 released Meta-Album datasets. Corresponding 95\% confidence intervals are within the size of the symbols.}
     \vspace{-0.5cm}
 \end{figure}

Results are shown in Figure~\ref{fig:fsl-plots}. A first observation is that Prototypical Networks (ProtoNet) dominate other algorithms (both within domain and cross-domain) and that the ranking of algorithms does not significantly change with the number of shots. However, the exception is FineTuning for 1-shot learning in the cross-domain configuration, which outperforms ProtoNet by a small margin. Moreover, we observe that FineTuning outperforms MAML and Matching Networks (the other episodic meta-learning algorithm we tried), corroborating findings showing that finetuning yields excellent few-shot learning performance without using episodic meta-learning~\cite{chen2019closer,huisman2021preliminary,tian2020rethinking,triantafillou2021learning}. We also see that the naive baseline TrainFromScratch yields the worst performance, indicating that meta-learning actually helps transfer knowledge to new tasks.
Furthermore, we observe that the performances improve with the number of shots (training examples per class). Lastly, the details provided in Appendix~\ref{appendix:fslappendix} and Appendix~\ref{appendix:cd-fs-additional} show that FineTuning is the fastest method at training time while ProtoNet and MatchingNet are the fastest methods at inference time with less than 1 second per task.

For cross-domain few-shot learning, as can be expected, the accuracy is lower since the problem is more complex. However, it does not dramatically decrease compared to within domain few-shot learning, which leads us to speculate that such a new problem is within reach of the current state-of-the-art. This gives rise to new opportunities for improvement in this more complicated and more realistic setting.

\subsubsection*{Difficulty of ``any''-way ``any''-shot learning}

Moving to yet more realistic and harder tasks, we also investigated the performance in the ``any''-way ``any''-shot setting, where tasks at meta-test time include a varying number of classes between 2 to 20 and a varying number of examples per class between 1 to 20. For example, at meta-test and meta-validation time, some carved out tasks might be as follows: \textbf{Test task 1:} 5-way 1-shot task from Dataset 9; \textbf{Test task 2:} 3-way 15-shot task from Dataset 3; \textbf{Test task 3:} 12-way 4-shot task from Dataset 8; etc. However, during the meta-training phase, we kept the number of classes constant (specifically, we used 5-way any-shot tasks). This facilitates using off-the-shelf meta-learning techniques. All other experimental conditions (hyper-parameters, computational resources) are the same as in the previous section.

\begin{figure}
     \centering
     \begin{subfigure}{0.48\textwidth}
         \centering
         \includegraphics[width=\textwidth]{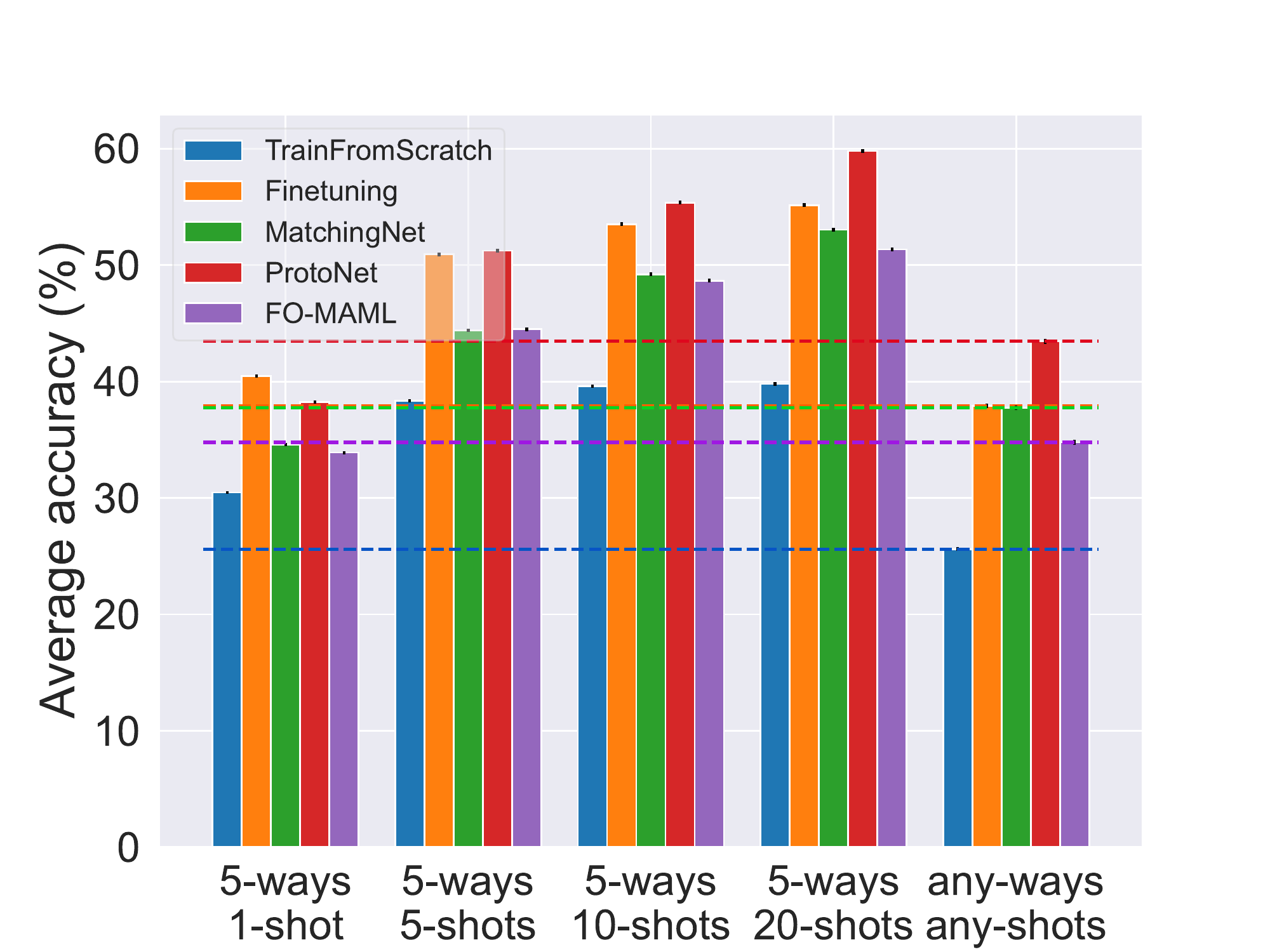}
         \caption{\label{fig:AnyWayAnyShotComparison:a} Difficulty of fixed and variable \# of ways and shots}
     \end{subfigure}
     \hfill
     \begin{subfigure}{0.48\textwidth}
         \centering
         \includegraphics[width=\textwidth]{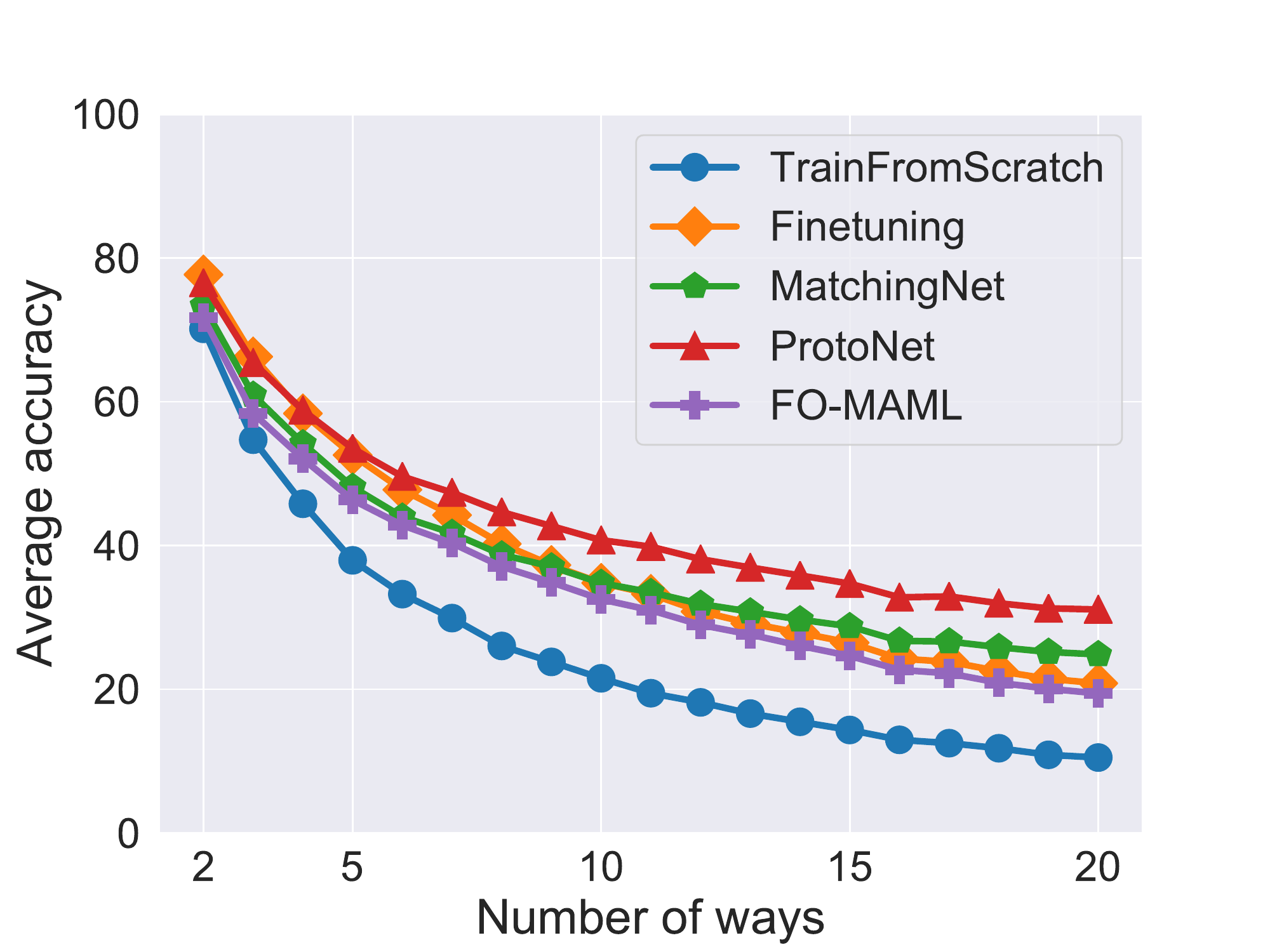}
         \caption{\label{fig:AnyWayAnyShotComparison:b} Influence of \# of ways}
     \end{subfigure}
     \caption{\label{fig:AnyWayAnyShotComparison} {\bf Comparison of ``cross-domain'' few-shot learning using fixed and variable number of ways and shots, and influence of number of ways on performance.} We plot few-shot learning meta-test mean task accuracy, averaged over test tasks drawn from the 30 released Meta-Album datasets. Corresponding 95\% confidence intervals are almost imperceptible as they are around $\pm0.15$.}
     \vspace{-0.5cm}
 \end{figure}

In Figure~\ref{fig:AnyWayAnyShotComparison:a} we can observe that the complexity of the any-way any-shot setting is similar to the 5-way 1-shot setting. Nevertheless, the meta-learning approaches (ProtoNet, MatchingNet, MAML) adapt better to this novel setting since their performance is better than the one achieved in the 5-way 1-shot setting, while the performance of FineTuning and TrainFromScractch is worse compared to the same setting. Additionally, the results presented in Figure~\ref{fig:AnyWayAnyShotComparison:b} and
Appendix~\ref{appendix:cd-fs-additional} show that the dominant difficulty factor in any-way any-shot learning is the variability in the number of ways since as it can be seen, the performance of the evaluated methods is highly affected by the increment in this number. 
This is supported by the fact that the absolute Pearson correlation between the number of ways and the test accuracy is larger (r=-0.55, p$<$0.05) than the correlation between the number of shots and the accuracy (r=0.1, p$<$0.05). Therefore, we anticipate that this new setting of any-way any-shot learning will deliver new interesting results in the upcoming challenge.

\section{Discussion and conclusion}
\label{sec:conclusion}

We introduce Meta-Album, a new meta-dataset for few-shot image classification, which is both practical and extensive: it includes many datasets from a wide variety of domains, all preprocessed to allow training according to different settings on commodity GPUs. It is especially amenable to evaluating meta-learning and transfer learning techniques. It can also be used for hierarchical classification as well as domain adaptation, due to the presence of overlapping classes between datasets, 
and continual learning, where algorithms are progressively trained across datasets. 

We evaluate the utility of Meta-Album using a range of few-shot learning experiments. Our findings include that Prototypical Networks and the FineTuning baseline perform quite well. This corroborates the results of the NeurIPS'21 challenge, in which the winners capitalized on the use of pre-trained backbones, obtaining results in the high 90\% classification accuracy in the ``within domain'' 5-way 5-shot setting~\cite{elbaz2022lessons}. Meta-Album will further challenge the research community by being considerably larger and by mixing tasks from multiple domains, in [2-20]-way [1-20]-shot settings. Furthermore, Meta-Album allows {\em de novo} training. 
We tested and compared this new framework to that of previous challenges and demonstrated an increased difficulty on all our baseline methods. 

In preparing the datasets we identified several types of biases, including correlations between class labels and nuisance variables (\eg background, luminosity, contrast, colour spectrum, position and orientation of objects). In this first release, we avoided correcting such biases, to avoid introducing yet more bias, and opted to homogenize the datasets by shuffling the examples. We documented our findings to facilitate the creation of challenges that study the problem of bias, in which the (meta-)training data and (meta-)test data will have distribution shifts.  

In future work, we want to make tasks more challenging by ensuring that every task consists of related concepts, belonging to a same super-class. 
For example, insects from the Coleoptera order have more resemblance with one another than with insects coming from another order, \eg butterflies. 
As such, it would be more challenging to have a task where the goal is to classify various insects from the Coleoptera order, rather than a tasks where insects from various orders are combined into one classification task. 
This requires datasets in which class hierarchies are provided, and we currently have only a few of those. 
Further work also includes introducing the even more difficult ``domain independent'' settings, in which meta-training and meta-testing are performed on entirely different domains.
Indeed, a problem setting where the goal is to learn how to learn on tasks that are not related to the meta-test data, would truly challenge a meta-learning system.

%
%
\begin{ack}
\label{sec:acknowledgement}

We gratefully acknowledge the data owners/creators: 

\textbf{LR\_AM.BRD}: Gerald Piosenka, \textit{Scottsdale, Arizona, United States};
\textbf{LR\_AM.DOG}: Aditya Khosla, Nityananda Jayadevaprakash, Bangpeng Yao and Li Fei-Fei from \textit{Stanford University};
\textbf{LR\_AM.AWA}: Christoph H. Lampert, Bernt Schiele and Zeynep Akata.

\textbf{SM\_AM.PLK}: Heidi M. Sosik, Emily E. Peacock, Emily F. Brownlee and Eric Orenstein from \textit{Woods Hole Oceanographic Institution, United States}; 
\textbf{SM\_AM.INS}: Grégoire Loïs, Colin Fontaine and Jean-Francois Julien from \textit{National Museum of Natural History Paris, France} and \textit{SPIPOLL Science project}; 
\textbf{SM\_AM.INS\_2}: Xiaoping Wu, Chi Zhan, Yukun Lai, Ming-Ming Cheng and Jufeng Yang.

\textbf{PLT.FLW}: Maria-Elena Nilsback and Andrew Zisserman from \textit{University of Oxford, England}; 
\textbf{PLT.PLT\_NET}: Garcin Camille, Joly Alexis, Bonnet Pierre, Lombardo Jean-Christophe, Affouard Antoine, Chouet Mathias, Servajean Maximilien, Salmon Joseph and Lorieul Titouan;
\textbf{PLT.FNG}: Lukáš Picek, Milan Šulc, Jiří Matas, Jacob Heilmann-Clausen, Thomas S. Jeppesen, Thomas Læssøe and Tobias Frøslev.

\textbf{PLT\_DIS.PLT\_VIL}: Sharada Mohanty, David Hughes, and Marcel Salathé, from \textit{EPFL Switzerland} and \textit{Penn State University}, J. Arun Pandian and G. Geetharamani, from \textit{Department of Mathematics, University College of Engineering, Anna University - BIT Campus and Department of Computer Science and Engineering, M.A.M. College of Engineering and Technology, Tiruchirappalli, India};
\textbf{PLT\_DIS.MED\_LF}: S Roopashree, J Anitha, from \textit{Visvesvaraya Technological University, R V Institute of Management, Dayananda Sagar University, India};
\textbf{PLT\_DIS.PLT\_DOC}: Sharada Mohanty, David Hughes, and Marcel Salathé.

\textbf{MCR.BCT}: Bartosz Zieliński, Anna Plichta, Krzysztof Misztal, Przemysław Spurek, Monika Brzychczy-Włoch and Dorota Ochońska from \textit{Uniwersytet Jagielloński};
\textbf{MCR.PRT}: Peter J Thul, Lovisa Akesson, Mikaela Wiking, Diana Mahdessian, Aikaterini Geladaki, Hammou Ait Blal, Tove Alm, Anna Asplund, Lars Björk, Lisa Breckels, and others from \textit{Protein Atlas};
\textbf{MCR.PNU}: Gamper Jevgenij , Koohbanani Navid Alemi , Benet Ksenija , Khuram Ali and Rajpoot Nasir from \textit{University of Warwick}.

\textbf{REM\_SEN.RESISC}: Gong Cheng, Junwei Han, and Xiaoqiang Lu from \textit{Northwestern Polytechnical University, Xi’an, China}; 
\textbf{REM\_SEN.RSICB}: Haifeng Li, Xin Dou, Chao Tao, Zhixiang Hou, Jie Chen, Jian Peng, Min Deng, Ling Zhao from \textit{Central South University, Changsha, China};
\textbf{REM\_SEN.RSD}: Yang Long, Yiping Gong, Zhifeng Xiao, and Qing Liu, Deren Li, Chunshan Wei, Gefu Tang and Junyi Liu from \textit{State Key Laboratory of Information Engineering in Surveying Mapping and Remote Sensing, Wuhan University, Wuhan 430079, China}.

\textbf{VCL.CRS}: Jonathan Krause, Michael Stark, Jia Deng and Li Fei-Fei from \textit{Stanford University};
\textbf{VCL.APL}: Wu Zhize;
\textbf{VCL.BTS}: Gundogdu E., Solmaz B, Yucesoy V., Koc A.

\textbf{MNF.TEX}: 
 Eric Hayman, Barbara Caputo, Mario Fritz, P. Mallikarjuna and Alireza Tavakoli Targhi from \textit{KTH Royal Institute of Technology in Stockholm} (for KTH TIPS and KTH TIPS 2); 
Gustaf Kylberg from \textit{Uppsala University, Sweden} (for Kylberg Texture); 
Jean Ponce, Svetlana Lazebnik and Cordelia Schmid from \textit{University of Illinois Urbana-Champaign} (for UIUC Textures);
\textbf{MNF.TEX\_DTD}: Mircea Cimpoi, Subhransu Maji, Iasonas Kokkinos, Sammy Mohamed and Andrea Vedaldi, the Authors of Describable Textures Dataset (DTD);
\textbf{MNF.TEX\_ALOT}: Gertjan Burghouts and Jan-Mark Geusebroek from \textit{University of Amsterdam, Netherlands}.

\textbf{HUM\_ACT.SPT}: Gerald Piosenka, \textit{Scottsdale, Arizona, United States};
\textbf{HUM\_ACT.ACT\_40}: B. Yao, X. Jiang, A. Khosla, A.L. Lin, L.J. Guibas, and L. Fei-Fei from \textit{Stanford University};
\textbf{HUM\_ACT.ACT\_410}: Mykhaylo Andriluka and Leonid Pishchulin and Peter Gehler and Schiele Bernt.

\textbf{OCR.MD\_MIX, OCR.MD\_5\_BIS, OCR.MD\_6}: Generated by Haozhe Sun (co-author).

We acknowledge the efforts of Philip Boser, Maria Belen Guaranda Cabezas, Jilin He, Felix Heron, Gabriel Lauzzana, Romain Mussard, and Manh Hung Nguyen for datasets and datasheets preparation. We also received useful input from many members of the TAU team of the LISN laboratory, Wei Wei Tu from 4Paradigm Inc, China, and the MetaDL technical crew: Adrian El Baz, Zhengying Liu, Adrien Pavao, Jennifer (Yuxuan) He, Yui Man Lui, S\'ebastien Treguer, Benjia Zhou, and Jun Wan, who participated in identifying datasets and contributed to discussions. 
We also would like to thank the hundreds of volunteers involved in the SPIPOLL citizen science program who pictured and identified insects.

This work was supported by ChaLearn, the ANR (Agence Nationale de la Recherche, National Agency for Research) under AI chair of excellence HUMANIA, grant number ANR-19-CHIA-0022, TAILOR EU Horizon 2020 grant 952215 and Labex Digicosme project ANR11LABEX0045DIGICOSME operated by ANR as part of the program Investissement d’Avenir Idex Paris Saclay (ANR11IDEX000302). 

In addition, some experiments were performed using the compute resources from the Academic Leiden Interdisciplinary Cluster Environment (ALICE) provided by Leiden University. This research was partially supported by TAILOR, a project funded by EU Horizon 2020 research and innovation programme under GA No 952215.

\end{ack}

\bibliography{ref}

\appendix

\clearpage
\newpage
\section{Datasets and meta-data}
\label{appendix:datasets}

Meta-Album datasets, their original sources and how the datasets are curated are explained separately in the following sections. All these datasets come from 10 domains:
\begin{itemize}
    \item Large Animals
    \item Small Animals
    \item Plants
    \item Plant Diseases
    \item Microscopy
    \item Remote Sensing
    \item Vehicles
    \item Manufacturing
    \item Human Actions
    \item Optical Character Recognition
\end{itemize}
We have 3 versions of the datasets:
\begin{enumerate}
    \item {\bf Micro}: a minimal version with 20 randomly selected classes and 40 images per class;
    \item {\bf Mini}: a medium version with all classes having at least 40 images per class, including 40 randomly selected images per class;
    \item {\bf Extended}: a full version that consists of all classes and all images per class.
\end{enumerate}

\begin{figure}[tbh!]
    
    \captionsetup[subfloat]{labelformat=empty}
    \centering
    \subfloat[LR\_AM.BRD]{\includegraphics[width=0.2\textwidth]{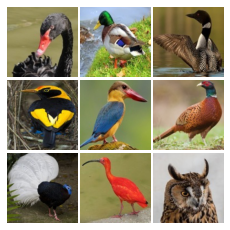}}
    \subfloat[LR\_AM.DOG]{\includegraphics[width=0.2\textwidth]{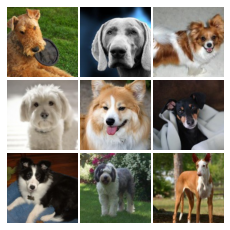}}
    \subfloat[LR\_AM.AWA]{\includegraphics[width=0.2\textwidth]{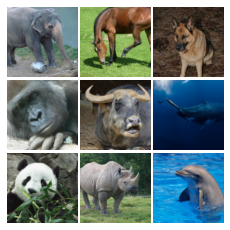}}
    \subfloat[SM\_AM.PLK]{\includegraphics[width=0.2\textwidth]{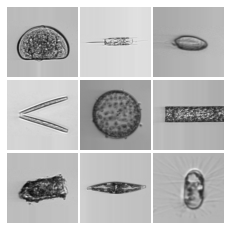}}
    \subfloat[SM\_AM.INS\_2]{\includegraphics[width=0.2\textwidth]{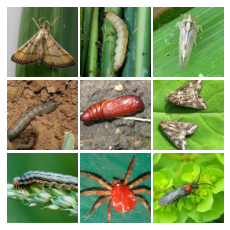}}
    \\
    \subfloat[SM\_AM.INS]{\includegraphics[width=0.2\textwidth]{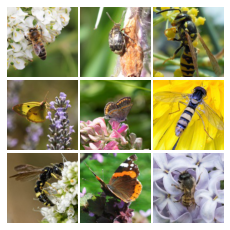}}
    \subfloat[PLT.FLW]{\includegraphics[width=0.2\textwidth]{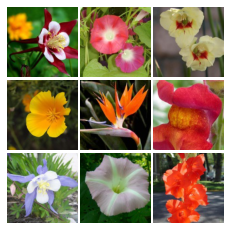}}
    \subfloat[PLT.PLT\_NET]{\includegraphics[width=0.2\textwidth]{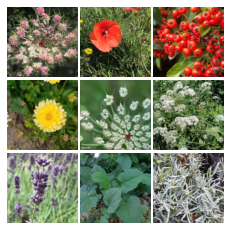}}
    \subfloat[PLT.FNG]{\includegraphics[width=0.2\textwidth]{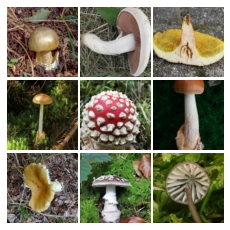}}
    \subfloat[PLT\_DIS.PLT\_VIL]{\includegraphics[width=0.2\textwidth]{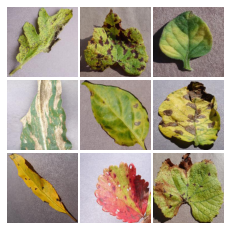}}
    \\
    \subfloat[PLT\_DIS.MED\_LF]{\includegraphics[width=0.2\textwidth]{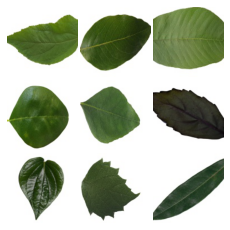}}
    \subfloat[PLT\_DIS.PLT\_DOC]{\includegraphics[width=0.2\textwidth]{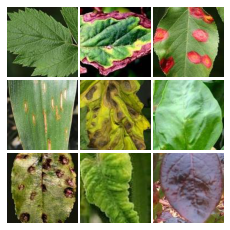}}
    \subfloat[MCR.BCT]{\includegraphics[width=0.2\textwidth]{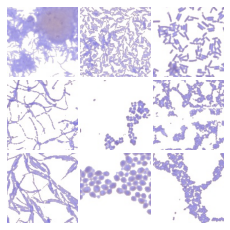}}
    \subfloat[MCR.PNU]{\includegraphics[width=0.2\textwidth]{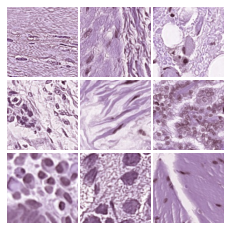}}
    \subfloat[MCR.PRT]{\includegraphics[width=0.2\textwidth]{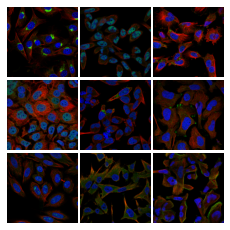}}
    \\
    \subfloat[REM\_SEN.RESISC ]{\includegraphics[width=0.2\textwidth]{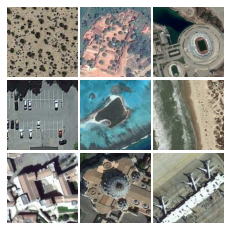}}
    \subfloat[REM\_SEN.RSICB]{\includegraphics[width=0.2\textwidth]{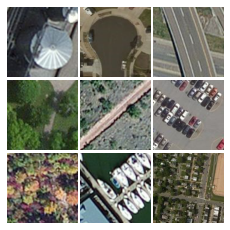}}
    \subfloat[REM\_SEN.RSD]{\includegraphics[width=0.2\textwidth]{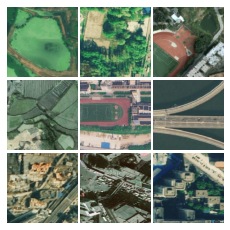}}
    \subfloat[VCL.CRS]{\includegraphics[width=0.2\textwidth]{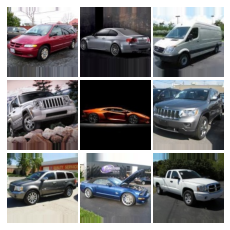}}
    \subfloat[VCL.APL]{\includegraphics[width=0.2\textwidth]{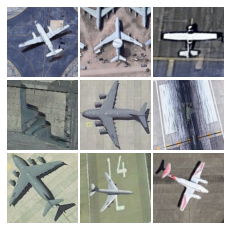}}
    \\
    \subfloat[VCL.BTS]{\includegraphics[width=0.2\textwidth]{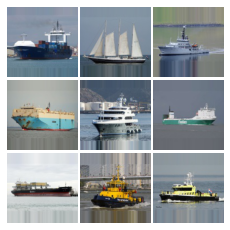}}
    \subfloat[MNF.TEX]{\includegraphics[width=0.2\textwidth]{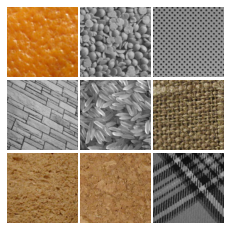}}
    \subfloat[MNF.TEX\_DTD]{\includegraphics[width=0.2\textwidth]{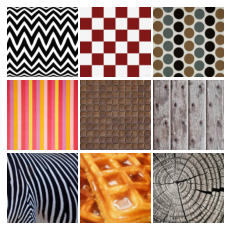}}
    \subfloat[MNF.TEX\_ALOT]{\includegraphics[width=0.2\textwidth]{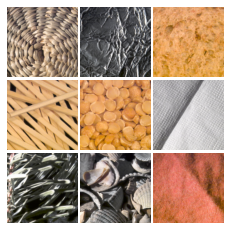}}
    \subfloat[HUM\_ACT.SPT]{\includegraphics[width=0.2\textwidth]{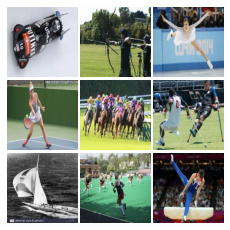}}
    \\
    \subfloat[HUM\_ACT.ACT\_40]{\includegraphics[width=0.2\textwidth]{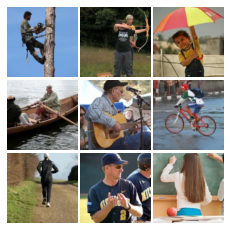}}
    \subfloat[HUM\_ACT.ACT\_410]{\includegraphics[width=0.2\textwidth]{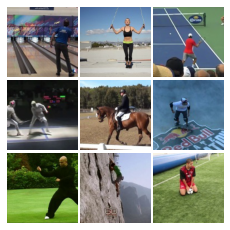}}
    \subfloat[OCR.MD\_MIX]{\includegraphics[width=0.2\textwidth]{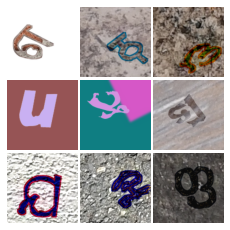}}
    \subfloat[OCR.MD\_5\_BIS]{\includegraphics[width=0.2\textwidth]{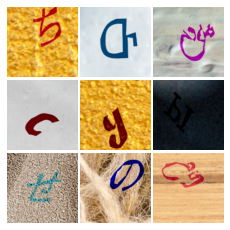}}
    \subfloat[OCR.MD\_6]{\includegraphics[width=0.2\textwidth]{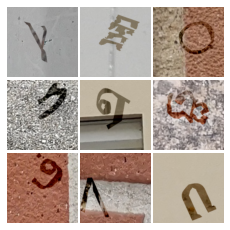}}
    
    \caption{Sample images from Meta-Album datasets.}
    \label{fig-sampleimages-more}
    
\end{figure}

Sample images from Meta-Album are provided in Figure~\ref{fig-sampleimages-more}. The formatted Meta-Album datasets are referred to as ``preprocessed" versions. Such preprocessed images are used in all Meta-Album versions: Micro, Mini, or Extended.

\subsection{Large Animals \textit{(LR\_AM)}}

\subsubsection*{BRD}
When Meta-Album was created, the \href{https://www.kaggle.com/datasets/gpiosenka/100-bird-species}{Birds dataset}~\cite{birds} contained images of 315 bird species, but now it has increased the number of species to 450. It has more than 49\,000 images, each with a resolution of 224x224 px. All the images have their natural background, which can lead to bias since, for example, some birds are frequently found in water backgrounds. Additionally, the dataset is imbalanced regarding the ratio of male species images to female species images. The preprocessed version distributed in Meta-Album is made from the original dataset by resizing all the images to a resolution of 128x128 px using an anti-aliasing filter.

\subsubsection*{DOG}
Researchers from Stanford University created the original \href{http://vision.stanford.edu/aditya86/ImageNetDogs/}{Dogs dataset}~\cite{dogs}. It contains more than 20\,000 images belonging to 120 breeds of dogs worldwide. The images and annotations came from ImageNet for the task of fine-grained image categorization. The number of images per class and the resolution is not balanced. Each class can have 148 to 252 images with a resolution from 100x105 to 2\,448x3\,264 px. This dataset has a little inter-class variation and a large intra-class variation due to color, pose, and occlusion. Most of the images in this dataset are taken in man-made environments leading to a significant background variation. The preprocessed version of this dataset is prepared from the original dataset by cropping the images from either side to make squared images. In case an image has a resolution lower than 128 px, the squared images are done by either duplicating the top and bottom-most 3 rows or the left and right most 3 columns based on the orientation of the original image. These square images are then resized into 128x128 px using an anti-aliasing filter.

\subsubsection*{AWA}
The original \href{https://cvml.ist.ac.at/AwA2/}{Animals with Attributes 2} (AWA) dataset~\cite{awa} was designed to benchmark transfer-learning algorithms, in particular attribute base classification and zero-shot learning. It has more than 37\,000 images from 50 animals, where each animal corresponds to a class. The images of this dataset were collected from public sources, such as Flickr, in 2016, considering only images licensed for free use and redistribution. Each class can have 100 to 1\,645 images with a resolution from 100x100 to 1\,893x1\,920 px. To preprocess this dataset, we cropped the images from either side to make them square. In case an image has a resolution lower than 128 px, the squared images are done by either duplicating the top and bottom-most 3 rows or the left and right most 3 columns based on the orientation of the original image. Lastly, the square images are resized into 128x128 px using an anti-aliasing filter.
\subsection{Small Animals \textit{(SM\_AM)}}

\subsubsection*{PLK}
The Plankton dataset is created by researchers at the \href{https://www.whoi.edu/}{Woods Hole Oceanographic Institution}. Imaging FlowCytobot (IFCB) was used for the data collection. The Complete process and mechanism are described in the paper~\cite{whoiplankton}. Each image in the dataset contains one or multiple planktons. The images are captured in a controlled environment and have different orientations based on the flow of the fluid in which the images are captured and the size and shape of the planktons. The preprocessed plankton dataset is prepared from the original WHOI Plankton dataset. The preprocessing of the images is done by creating a background squared image by either duplicating the top and bottom-most 3 rows or the left and right most 3 columns based on the orientation of the original image to match the width or height of the image respectively. A Gaussian kernal of size 29x29 is applied to the background image to blur the image. Finally, the original plankton image is pasted on the background image at the center of the image. The squared background image with the original plankton image on top of it as one image is then resized into 128x128 with anti-aliasing.

\subsubsection*{INS\_2}
The pest insects dataset~\cite{Wu2019Insect} was originally created as a large scale benchmark dataset for \href{https://github.com/xpwu95/IP102}{Insect Pest Recognition}. It contains more than 75\,000 images belongs to 102 categories. It also has a hierarchical taxonomy and the insect pests which mainly affect one specific agricultural product are grouped into the same upper-level category. The preprocessed version is made from the original dataset by cropping the images in perfect squares and then resizing them into the required images size of 128x128.

\subsubsection*{INS}
The original Insects dataset~\cite{insects} is created by the \href{https://www.mnhn.fr/fr}{National Museum of Natural History, Paris}. It has more than 290\,000 images in different sizes and orientations. The dataset has hierarchical classes which are listed from top to bottom as Order, Super-Family, Family, and Texa. Each image contains an insect in its natural environment or habitat, \textit{i.e}, either on a flower or near to vegetation. The images are collected by the researchers and hundreds of volunteers from \href{https://www.spipoll.org/}{SPIPOLL Science project}. The images are uploaded to a centralized server either by using the \href{https://www.spipoll.org/}{SPIPOLL website}, \href{https://play.google.com/store/apps/details?id=fr.eneo.spipoll&hl=fr}{Android application} or \href{https://apps.apple.com/fr/app/spipoll/id1495843067}{IOS application}. The preprocessed insect dataset is prepared from the original Insects dataset by carefully preprocessing the images, \ie cropping the images from either side to make squared images. These cropped images are then resized into 128x128 using Open-CV with an anti-aliasing filter.

\subsection{Plants \textit{(PLT)}}

\subsubsection*{FLW}
The \href{https://www.robots.ox.ac.uk/~vgg/data/flowers/102/index.html}{Flowers dataset}~\cite{Nilsback08} consists of a variety of flowers gathered from different websites and some are photographed by the original creators. These flowers are commonly found in the UK. The images generally have large scale, pose and light variations. Some categories of flowers in the dataset has large variations of flowers while other have similar flowers in a category. The dataset was created back in 2008 at Oxford University by Nilsback, M-E. and Zisserman, A.\cite{Nilsback08}. The Flowers dataset in the Meta-Album meta-dataset is a preprocessed version of the original flowers dataset. The images are first cropped and made into squared images which are then resized into 128x128 with anti-aliasing filter.

\subsubsection*{PLT\_NET}
Meta-Album PlantNet dataset is created by sampling the \href{https://openreview.net/forum?id=eLYinD0TtIt}{Pl@ntNet-300k dataset}, itself a sampling of the \href{https://plantnet.org/en/}{Pl@ntNet Project}'s repository. The images and labels which enter this database are sourced by citizen botanists from around the world, then confirmed using a weighted reliability score from others users, such that each image has been reviewed by 2.03 citizen botanists on average. Of the 1\,081 classes in the original Pl@ntNet-300k dataset, PLT\_NET retains the 25 most populous classes, belonging to 21 genera, for a total of 120\,688 images total, with min 2\,914, max 9\,011 image distribution per class. Each image contains a colored 128x128 image of a plant or a piece or a plant from the corresponding class (or in some cases sketches of plants or plant cells on microscope slides), scaled from the initial variable width using the INTER\_AREA anti-aliasing filter from \href{https://pypi.org/project/opencv-python/}{Open-CV}~\cite{opencv_library}. Almost all images were initially square; cropping by taking the largest  possible square with center at the middle of the initial image was applied otherwise.

\subsubsection*{FNG}
Meta-Album Fungi dataset is created by sampling the \href{https://arxiv.org/abs/2103.10107}{Danish Fungi 2020 dataset}, itself a sampling of the \href{https://svampe.databasen.org/}{Atlas of Danish Fungi}'s repository. The images and labels which enter this database are sourced by a group consisting of 3\,300 citizen botanists, then verified by their peers using a ranking of each person's reliability, then finally verified by experts working at the Atlas. Of the 128 classes in the original Danish Fungi 2020 dataset, FNG retains the 25 most populous classes, belonging to six genera, for a total of 15\,122 images total, with min 372, and max 1\,221 images per class. Each image contains a colored 128x128 image of a fungus or a piece of a fungus from the corresponding class. Because the initial data were of widely varying sizes, we needed to crop a significant portion of the images, which we implemented by taking the largest  possible square with center at the middle of the initial image. We then scaled each squared image to the 128x128 standard using  the INTER\_AREA anti-aliasing filter from \href{https://pypi.org/project/opencv-python/}{Open-CV}~\cite{opencv_library}.

\subsection{Plant Diseases \textit{(PLT\_DIS)}}

\subsubsection*{PLT\_VIL} 
The \href{https://data.mendeley.com/datasets/tywbtsjrjv/1}{Plant Village dataset}~\cite{plantvillage, plantvillagedata} contains camera photos of 17 crop leaves. The original image resolution is 256x256 px. This collection covers 26 plant diseases and 12 healthy plants. The leaves are removed from the plant and placed on gray or black background, in various lighting conditions. All images are captured on a variety of gray backgrounds, except Corn Common rust which has a black background. For the curated version, we exclude the irrelevant Background and Corn Common Rust classes from the original collection. Plant Village has a 2-level label hierarchy, the supercategory is the crop type and the category is the disease type. We have preprocessed Plant Village for Meta-Album by resizing a subset from the original dataset to 128x128 image size.

\subsubsection*{MED\_LF}
The \href{https://data.mendeley.com/datasets/nnytj2v3n5/1}{Medicinal Leaf Database}~\cite{medicinalleaf} gathers 30 species of healthy and mature medicinal herbs. The leaves are plucked from different plants of the same species, then placed on a white uniform background. There are around 1\,800 images in total, captured with a mobile phone camera. The original resolution is 1\,600x1\,200 px.
We create Medleaf for Meta-Album by cropping them at the center and resize to 128x128 px.

\subsubsection*{PLT\_DOC}
The \href{https://github.com/pratikkayal/PlantDoc-Dataset}{PlantDoc dataset}~\cite{plant-doc} is made up of images of leaves of healthy and unhealthy plants. The images were downloaded from Google Images and Ecosia, and later cropped by the authors, so generally,  one complete leaf fits in one image. The original, uncropped images are generally different in scale, light conditions, and pose. However, within one category, images of leaves that came from the same original image can be found. The images correspond to 27 classes, including plant disease names and plant species names, {\em e.g.}: Corn Leaf Blight and Cherry Leaf respectively. The dataset was created for a benchmarking classification model work, published in 2020 by Singh et al. The PlantDoc dataset in the Meta-Album benchmark is extracted from a preprocessed version of the original PlantDoc dataset. First, to get \textit{i.i.d.} samples, only one leaf image per each original image is randomly picked. Then, leaves images are cropped and made into squared images which are then resized into 128x128 with anti-aliasing filter.

\subsection{Microscopy  \textit{(MCR)}}

\subsubsection*{BCT}
The \href{https://github.com/gallardorafael/DIBaS-Dataset}{Digital Images of Bacteria Species dataset (DIBaS)}~\cite{diBas} is a dataset of 33 bacterial species with around 20 images for each species. 
For the Meta-Album, since the images were large (2\,048x1\,532) with very few samples in each class, we decided to split each image into several smaller images before resizing them to 128x128. We then obtained a preprocessed dataset of 4\,060 images with at least 108 images for each class. This dataset was also preprocessed with blob normalization techniques, which is quite unusual for this type of image. The goal of this transformation was to reduce the importance of color in decision-making for a bias-aware challenge.

\subsubsection*{PRT}
This dataset is a subset of the Subcellular dataset~\cite{thul2017subcellular} in the \href{https://www.proteinatlas.org/}{Protein Atlas project}.
The original dataset, which stems from the \href{https://www.kaggle.com/competitions/human-protein-atlas-image-classification}{Human Protein Atlas Image Classification Kaggle competition}, comprises 31\,072 RGBY images of size 512x512 px, each of which belongs to one or more out of 28 classes. The labels correspond to protein organelle localizations. 
For Meta-Album, we performed two modifications:
(1), to turn the dataset into a multi-class dataset, we dropped all images belonging to more than a single class and also those images that belong to classes with less than 40 members;
(2) we converted the remaining images into RGB simply by dropping the yellow channel; this was also a common practice in the competition.
Finally, and as for all datasets in Meta-Album, the images from the original dataset were resized to 128x128 image size.

\subsubsection*{PNU}
The \href{https://jgamper.github.io/PanNukeDataset}{PanNuke dataset}~\cite{pannuke, pannuke-dataset} is a semi-automatically generated segmentation and classification task of nuclei. The dataset contains 7\,753 images of 19 different tissue types. For the Meta-Album meta-dataset, even though this dataset was designed as a segmentation task, we were able to transform it into a tissue classification task since we had the tissue type for each sample in the dataset. We also resized the images to 128x128 pixels and applied stain normalization to avoid bias and remove some spurious features.

\subsection{Remote Sensing \textit{(REM\_SEN)}}

\subsubsection*{RESISC}
\href{https://gcheng-nwpu.github.io/}{RESISC45 dataset}~\cite{resisc} gathers 700 RGB images of size 256x256 px for each of 45 scene categories. The data authors strive to provide a challenging dataset by increasing both within-class diversity and between-class similarity, as well as integrating many image variations. Even though RESISC45 does not propose a label hierarchy, it can be created from other common aerial image label organization scheme.
We have preprocessed RESISC for Meta-Album by resizing the dataset to 128x128 px.

\subsubsection*{RSICB}
\href{https://github.com/lehaifeng/RSI-CB}{RSICB128 dataset}~\cite{rsicb} covers 45 scene categories, assembling in total 36\,000 images of resolution 128x128 px. The data authors select various locations around the world, and follow China's land-use classification standard. This collection has 2-level label hierarchy with 6 super-categories: agricultural land, construction land  and  facilities, transportation  and  facilities,  water  and  water conservancy  facilities, woodland, and other lands. 
The preprocessed version of RSICB is created by resizing the images into 128x128 px using an anti-aliasing filter.

\subsubsection*{RSD}
\href{https://github.com/RSIA-LIESMARS-WHU/RSD46-WHU}{RSD46 dataset}~\cite{rsd46localize, rsd46classify} is collected from Google Earth and Tianditu. The collection contains 46 scene categories, with a total of 117\,000 images. Each scene category has between 500-3\,000 images. The original resolution are 256x256 px or 512x512 px. We have created preprocessed version of RSD for Meta-Album by resizing the original dataset to 128x128 px.

\subsection{Vehicles \textit{(VCL)}}

\subsubsection*{APL}
The original \href{https://zenodo.org/record/3464319}{Airplanes dataset}~\cite{airplanes} comprises more than 9\,000 remote sensing images acquired from Google Earth satellite imagery, including 21 different types of aircraft from around 36 airports. All the images were carefully labeled by seven specialists in the field of remote sensing image interpretation. Each class can have 230 to 846 images, where each image contains only one complete aircraft, and they have variable resolutions. To preprocess this dataset, we cropped the images from either side to make them square, and then we resized them into 128x128 px using an anti-aliasing filter.

\subsubsection*{CRS}
The original \href{http://ai.stanford.edu/~jkrause/cars/car_dataset.html}{Cars dataset}~\cite{cars} was collected in 2013, and it contains more than 16\,000 images from 196 classes of cars. Most images are on the road, but some have different backgrounds, and each image has only one car. Each class can have 48 to 136 images of variable resolutions. The preprocess version for this dataset was obtained by creating square images either duplicating the top and bottom-most 3 rows or the left and right most 3 columns based on the orientation of the original image. In this case, cropping was not applied to create the square images since following this technique results in losing too much information from the cars. Then, the square images were resized into 128x128 px using an anti-aliasing filter.

\subsubsection*{BTS}
The original version of the Meta-Album boats dataset is called \href{https://github.com/avaapm/marveldataset2016}{MARVEL dataset}~\cite{boats}. It has more than 138\,000 images of 26 different maritime vessels in their natural background. Each class can have 1\,802 to 8\,930  images of variable resolutions. To preprocess this dataset, we either duplicate the top and bottom-most 3 rows or the left and right most 3 columns based on the orientation of the original image to create square images. No cropping was applied because the boats occupy most of the image, and applying this technique will lead to incomplete images. Finally, the square images were resized into 128x128 px using an anti-aliasing filter.

\subsection{Manufacturing \textit{(MNF)}}

\subsubsection*{TEX}
The original Textures dataset is a combination of 4 texture datasets: \href{https://www.csc.kth.se/cvap/databases/kth-tips/index.html}{KTH-TIPS and KTH-TIPS 2}~\cite{Fritz2004THEKD, Mallikarjuna2006THEK2}, 
\href{http://www.cb.uu.se/~gustaf/texture/}{Kylberg Textures Dataset}~\cite{Kylberg2011c} and 
{UIUC Textures Dataset}~\cite{lazebnik:inria-00548530}. The data in all four datasets is collected in laboratory conditions, \ie images were captured in a controlled environment with configurable brightness, luminosity, scale and angle. The \textit{KTH-TIPS} dataset was collected by Mario Fritz and \textit{KTH-TIPS 2} dataset was collected by P. Mallikarjuna and Alireza Tavakoli Targhi, created in 2004 and 2006 respectively. Both of these datasets were prepared under the supervision of Eric Hayman and Barbara Caputo. The data for \textit{Kylberg Textures Dataset} and \textit{UIUC Textures Dataset} data was collected by the original authors of these datasets in September 2010 and August 2005 respectively.

The Meta-Album Textures dataset is a preprocessed version of the original dataset (combination of 4 datasets). All the images are preprocessed by first cropping into perfect squared images and then resized into 128x128 with an anti-aliasing filter.

\subsubsection*{TEX\_DTD}
The \href{https://www.robots.ox.ac.uk/~vgg/data/dtd/index.html}{Textures DTD dataset} is a large textures dataset which consists of 5\,640 images. The data is collected from \href{https://images.google.com/}{Google} and \href{https://www.flickr.com/}{Flicker} by the original authors of the paper ``Describing Textures in the Wild''\cite{cimpoi2013describing}. The data was annotated using \href{https://www.mturk.com/}{Amazon Mechanical Turk}. The data collection process is mentioned on the \href{https://www.robots.ox.ac.uk/~vgg/data/dtd/index.html}{dataset overview page} For  Meta-Album meta-dataset, this dataset is preprocessed by cropping the images to square images and then resizing them to 128x128 using Open-CV with an anti-aliasing filter. This dataset has 47 class labels.

\subsubsection*{TEX\_ALOT}
\href{https://aloi.science.uva.nl/public_alot/}{Textures ALOT dataset}~\cite{texture_alot} consists of 27\,500 images from 250 categories. The images in the dataset are captured in controlled environment by the creators of the dataset. The images have different viewing angle, illumination angle, and illumination color for each material of texture.

A preprocessed version of Textures-ALOT is used in the Meta-Album meta-dataset. The images are first cropped into square images and then resized to 128x128 with anti-aliasing filter.

\subsection{Human Actions \textit{(HUM\_ACT)}}

\subsubsection*{SPT}
The \href{https://www.kaggle.com/datasets/gpiosenka/sports-classification}{100-Sports dataset}~\cite{100-sports} is a collection of sports images covering 73 different sports. Images are 224x224x3 in size and in .jpg format. Images were gathered from internet searches. The images were scanned with a duplicate image detector program and all duplicate images were removed. For Meta-Album, the dataset is preprocessed and images are resized into 128x128 pixels using Open-CV~\cite{opencv_library}

\subsubsection*{ACT\_40}
The \href{http://vision.stanford.edu/Datasets/40actions.html}{Stanford 40 Actions dataset}~\cite{act-40} contains images of humans performing 40 actions. There are 9\,532 images in total with 180-300 images per action class. The dataset is designed for understanding human actions in still images. For Meta-Album, the dataset is preprocessed and images are resized into 128x128 pixels using Open-CV~\cite{opencv_library}.

\subsubsection*{ACT\_410}
The \href{http://human-pose.mpi-inf.mpg.de/#download}{MPII Human Pose dataset}~\cite{act-410} is a state of the art benchmark for evaluation of articulated human pose estimation. It includes around 25\,000 images containing over 40\,000 people with annotated body joints. The images were systematically collected using an established taxonomy of every day human activities. Overall the dataset covers 410 human activities and each image is provided with an activity label. Each image was extracted from a YouTube video. Like other Meta-Album datasets, this dataset is preprocessed and all images are resized into 128x128 pixels.

\subsection{Optical Character Recognition \textit{(OCR)}}

\subsubsection*{MD\_MIX}
OmniPrint-MD-mix dataset consists of 28\,240 images (128x128, RGB) from 706 categories. The images are synthesized with OmniPrint~\cite{sun2021omniprint}, and no further processing was done. The OmniPrint synthesis parameters are stated as follows: font size is 192, image size is 128, the strength of random perspective transformation is 0.04, left/right/top/bottom margins are all \(20\%\) of the image size, the strength of pre-rasterization elastic transformation is 0.035, random translation is activated both horizontally and vertically, rotation is within \(-60\) and \(60\) degrees, horizontal shear is within \(-0.5\) and \(0.5\), brightness is within \(0.8333\) and \(1.2\), contrast is within \(0.8333\) and \(1.2\), color enhancement is within \(0.8333\) and \(1.2\). The other parameters vary between images. We designed 20 settings, each setting is used to synthesize 2 images. The 20 settings are described below:

\begin{enumerate}
    \item plain white background, random color foreground, trivial image blending (pasting)
    \item plain white background, random color foreground, trivial image blending (pasting), random color outline
    \item plain white background, random color foreground, trivial image blending (pasting), morphological gradient operation with elliptical kernel (kernel size is 5) 
    \item plain white background, image/texture foreground, trivial image blending (pasting)
    \item plain white background, image/texture foreground, image/texture outline, trivial image blending (pasting)
    \item random color background, trivial image blending (pasting), random color foreground
    \item random color background, trivial image blending (pasting), random color foreground, random color outline
    \item random color background, trivial image blending (pasting), random color foreground, morphological gradient operation with elliptical kernel (kernel size is 5) 
    \item random color background, trivial image blending (pasting), image/texture foreground
    \item random color background augmented with random polygons, trivial image blending (pasting), random color foreground
    \item image/texture background, Poisson image blending~\cite{poissonimageediting}, random color foreground
    \item image/texture background, Poisson image blending~\cite{poissonimageediting}, random color foreground, random color outline
    \item image/texture background, Poisson image blending~\cite{poissonimageediting}, random color foreground, morphological gradient operation with elliptical kernel (kernel size is 5) 
    \item image/texture background, Poisson image blending~\cite{poissonimageediting}, image/texture foreground
    \item image/texture background, Poisson image blending~\cite{poissonimageediting}, image/texture foreground, morphological gradient operation with elliptical kernel (kernel size is 5) 
    \item image/texture background, Poisson image blending~\cite{poissonimageediting}, image/texture foreground, random color outline
    \item image/texture background, Poisson image blending~\cite{poissonimageediting}, image/texture foreground, random color outline, morphological gradient operation with elliptical kernel (kernel size is 5) 
    \item image/texture background, Poisson image blending~\cite{poissonimageediting}, image/texture foreground, image/texture outline
    \item image/texture background, Poisson image blending~\cite{poissonimageediting}, image/texture foreground, image/texture outline, morphological gradient operation with elliptical kernel (kernel size is 5) 
    \item image/texture background, Poisson image blending~\cite{poissonimageediting}, image/texture foreground, image/texture outline, outline size is 10
\end{enumerate}

All images/textures consists of photos taken by a personal mobile phone~\cite{sun2021omniprint}.

The 706 categories are characters from: 

\begin{itemize}
    \item Armenian
    \begin{itemize}
        \item lowercase letters
    \end{itemize}
    \item ASCII digits
    \item Balinese
    \begin{itemize}
        \item consonants
        \item digits
    \end{itemize}
    \item Basic Latin
    \begin{itemize}
        \item lowercase letters
    \end{itemize}
    \item Devanagari
    \begin{itemize}
        \item digits
    \end{itemize}
    \item Georgian
    \item Gujarati
    \begin{itemize}
        \item consonants
    \end{itemize}
    \item Hebrew
    \item Hiragana
    \item khmer
    \begin{itemize}
        \item consonants
    \end{itemize}
    \item Mongolian
    \begin{itemize}
        \item basic letters
    \end{itemize}
    \item Myanmar
    \begin{itemize}
        \item digits
    \end{itemize}
    \item N Ko
    \begin{itemize}
        \item letters
        \item digits
    \end{itemize}
    \item Oriya
    \begin{itemize}
        \item consonants
    \end{itemize}
    \item Russian
    \item Sinhala
    \begin{itemize}
        \item independent vowels
    \end{itemize}
    \item Tamil
    \begin{itemize}
        \item consonants
        \item digits
    \end{itemize}
    \item Telugu
    \begin{itemize}
        \item digits
    \end{itemize}
    \item Thai
    \begin{itemize}
        \item consonants
    \end{itemize}
    \item Tibetan
    \begin{itemize}
        \item digits
    \end{itemize}
\end{itemize}

\subsubsection*{MD\_5\_BIS}
OmniPrint-MD-5-bis dataset consists of 28\,240 images (128x128, RGB) from 706 categories. The images are synthesized with OmniPrint~\cite{sun2021omniprint}, and no further processing was done. The OmniPrint synthesis parameters are stated as follows: font size is 192, image size is 128, the strength of random perspective transformation is 0.04, left/right/top/bottom margins are all \(20\%\) of the image size, the strength of pre-rasterization elastic transformation is 0.035, random translation is activated both horizontally and vertically, image blending method is Poisson Image Editing~\cite{poissonimageediting}, rotation is within \(-60\) and \(60\) degrees, horizontal shear is within \(-0.5\) and \(0.5\), the foreground is filled with a random color, the background consists of images downloaded from \href{https://www.pexels.com/}{Pexels}.

The 706 categories are characters from: 

\begin{itemize}
    \item Armenian
    \begin{itemize}
        \item lowercase letters
    \end{itemize}
    \item ASCII digits
    \item Balinese
    \begin{itemize}
        \item consonants
        \item digits
    \end{itemize}
    \item Basic Latin
    \begin{itemize}
        \item lowercase letters
    \end{itemize}
    \item Devanagari
    \begin{itemize}
        \item digits
    \end{itemize}
    \item Georgian
    \item Gujarati
    \begin{itemize}
        \item consonants
    \end{itemize}
    \item Hebrew
    \item Hiragana
    \item khmer
    \begin{itemize}
        \item consonants
    \end{itemize}
    \item Mongolian
    \begin{itemize}
        \item basic letters
    \end{itemize}
    \item Myanmar
    \begin{itemize}
        \item digits
    \end{itemize}
    \item N Ko
    \begin{itemize}
        \item letters
        \item digits
    \end{itemize}
    \item Oriya
    \begin{itemize}
        \item consonants
    \end{itemize}
    \item Russian
    \item Sinhala
    \begin{itemize}
        \item independent vowels
    \end{itemize}
    \item Tamil
    \begin{itemize}
        \item consonants
        \item digits
    \end{itemize}
    \item Telugu
    \begin{itemize}
        \item digits
    \end{itemize}
    \item Thai
    \begin{itemize}
        \item consonants
    \end{itemize}
    \item Tibetan
    \begin{itemize}
        \item digits
    \end{itemize}
\end{itemize}

\subsubsection*{MD\_6}
OmniPrint-MD-6 dataset consists of 28\,120 images (128x128, RGB) from 703 categories. The images are synthesized with OmniPrint~\cite{sun2021omniprint}, no further processing was done. The OmniPrint synthesis parameters are stated as follows: font size is 192, image size is 128, the strength of random perspective transformation is 0.04, left/right/top/bottom margins are all \(20\%\) of the image size, the strength of pre-rasterization elastic transformation is 0.035, random translation is activated both horizontally and vertically, image blending method is Poisson Image Editing~\cite{poissonimageediting}, rotation is within \(-60\) and \(60\) degrees, horizontal shear is within \(-0.5\) and \(0.5\), both foreground and background are images taken from a personal mobile phone~\cite{sun2021omniprint}.

The 703 categories are characters from: 

\begin{itemize}
    \item Arabic
    \item Armenian
    \begin{itemize}
        \item uppercase letters
        \item lowercase letters
    \end{itemize}
    \item Balinese
    \begin{itemize}
        \item independent vowels
    \end{itemize}
    \item Basic Latin
    \begin{itemize}
        \item uppercase letters
    \end{itemize}
    \item Bengali
    \begin{itemize}
        \item consonants
        \item digits
        \item independent vowels
    \end{itemize}
    \item punctuation symbols
    \item Devanagari
    \begin{itemize}
        \item consonants
        \item independent vowels
    \end{itemize}
    \item Ethiopic
    \begin{itemize}
        \item digits
    \end{itemize}
    \item Greek
    \item Gujarati
    \begin{itemize}
        \item digits
        \item independent vowels
    \end{itemize}
    \item Katakana
    \item Khmer
    \begin{itemize}
        \item digits
        \item independent vowels
    \end{itemize}
    \item Lao
    \begin{itemize}
        \item consonants
        \item digits
    \end{itemize}
    \item Mongolian
    \begin{itemize}
        \item digits
    \end{itemize}
    \item Myanmar
    \begin{itemize}
        \item consonants
        \item independent vowels
    \end{itemize}
    \item Oriya
    \begin{itemize}
        \item digits
        \item independent vowels
    \end{itemize}
    \item Sinhala
    \begin{itemize}
        \item astrological digits
        \item consonants
    \end{itemize}
    \item Tamil
    \begin{itemize}
        \item independent vowels
    \end{itemize}
    \item Telugu
    \begin{itemize}
        \item consonants
        \item independent vowels
    \end{itemize}
    \item Thai
    \begin{itemize}
        \item digits
    \end{itemize}
    \item Tibetan
    \begin{itemize}
        \item consonants
    \end{itemize}
\end{itemize}

\subsection{Meta-data}
Meta-Album datasets have the following meta-data files:
\begin{itemize}
    \item \textbf{labels.csv} : The meta-data in \textit{labels.csv} consists of the \textit{FILE\_NAME}, \textit{CATEGORY} and \textit{SUPER\_CATEGORY} (if any). In the case of OCR datasets, it contains some extra information about the images of characters in the data, \eg shear, stretch, rotation, font, etc.
    \item \textbf{info.json} : It consists of meta-data (dataset name, description, number of categories and super-categories, column names, etc.) which is useful in reading the data, specially for data quality control. It also gives important statistics about the dataset.
    
    \item \textbf{DATASET\_info.json} : This file is a detailed overview of the dataset which includes:
    \begin{itemize}
        \item Meta-Album ID
        \item Domain ID
        \item Domain name
        \item Dataset ID
        \item Dataset name
        \item Dataset description
        \item Data format
        \item Image size
        \item License name
        \item License URL
        \item Source
        \item Source URL
        \item Original author
        \item Original contact 
        \item Citation
        \item Meta-Album dataset creator
        \item Created data
        \item Contact details
        \item Meta-Album website
        \item Dataset download links (Exntended, Mini, Micro)
        \item Data statistics (Exntended, Mini, Micro)
        
    \end{itemize}
\end{itemize}

\clearpage
\newpage
\section{License information of Meta-Album datasets}
\label{appendix:license}

In this section, we provide details about the licenses of all Meta-Album datasets.

We researched the ownership of all datasets to determine the type of usage permission. We contacted original owners by email to clarify permissions if needed, when no explicit statement of license or permission was found. 

Unless otherwise stated, all data are re-distributed by us under a non-commercial license. When no original license could be identified, we use CC BY NC 4.0 by default. For commercial use, users should contact the original owners, whose contact information is found in the datasheets,  in Appendix~\ref{appendix:datasets} and on Meta-Album website (\url{https://meta-album.github.io/}).

\begin{table}[h]
  \caption{Meta-Album: Datasets license information
  }
  \label{tab:datasets_licenses}
  \centering
  \begin{adjustbox}{width=\linewidth}
  \begin{tabular}{l l l l l}
    \toprule
    {\bf Domain Name} & {\bf Set \#} &  {\bf Dataset ID} & {\bf License} & {\bf Original source}
    \\
    \midrule
    
    \multirow{3}{*}{Large Animals} 
    & 0 & {\it BRD} & CC0 Public Domain & \href{https://www.kaggle.com/gpiosenka/100-bird-species}{Birds 400}~\cite{birds}
    \\
    & 1 & {\it DOG} & CC BY-NC 4.0 & \href{http://vision.stanford.edu/aditya86/ImageNetDogs/}{Stanford Dogs}~\cite{dogs}
    \\
    & 2 & {\it AWA} & Creative Commons & \href{https://cvml.ist.ac.at/AwA2/}{AWA}~\cite{awa}
    \\
    \midrule
    
    \multirow{3}{*}{Small Animals} 
    & 0 & {\it PLK} & MIT License & \href{https://github.com/hsosik/WHOI-Plankton}{WHOI}~\cite{whoiplankton}
    \\
    & 1 & {\it INS\_2} & CC BY-NC 4.0 & \href{https://github.com/xpwu95/IP102}{Pest Insects}~\cite{Wu2019Insect}
    \\
    & 2 & {\it INS} & CC BY-NC 2.0 & \href{https://www.spipoll.org/}{SPIPOLL}~\cite{insects}
    \\

    \midrule
    
    \multirow{3}{*}{Plants} 
    & 0 & {\it FLW} & GNU General Public License Version 2 & \href{https://www.robots.ox.ac.uk/~vgg/data/flowers/}{Flowers}~\cite{Nilsback08}
    \\
    & 1 & {\it PLT\_NET} & Creative Commons Attribution 4.0 International & \href{https://plantnet.org/en/2021/03/30/a-plntnet-dataset-for-machine-learning-researchers/}{PlantNet}~\cite{plantnet}
    \\
    & 2 & {\it FNG} & BSD-3-Clause License &  \href{https://sites.google.com/view/danish-fungi-dataset}{Danish Fungi}~\cite{danish-fungi}
    \\
    \midrule
    
    \multirow{3}{*}{Plant Diseases} 
    & 0 & {\it PLT\_VIL} & CC0 1.0 & \href{https://github.com/spMohanty/PlantVillage-Dataset}{PlantVillage}~\cite{plantvillage, plantvillagedata}
    \\
    & 1 & {\it MED\_LF} & CC BY 4.0 & \href{https://data.mendeley.com/datasets/nnytj2v3n5/1}{Medicial Leaf}~\cite{medicinalleaf}
    \\
    & 2 & {\it PLT\_DOC} & Creative Commons Attribution 4.0 International & \href{https://github.com/pratikkayal/PlantDoc-Object-Detection-Dataset}{Plant Doc}~\cite{plant-doc}
    \\
    \midrule
    
    \multirow{3}{*}{Microscopy} 
    & 0 & {\it BCT} & CC BY-NC 4.0 & \href{https://github.com/gallardorafael/DIBaS-Dataset}{DiBas}~\cite{diBas}
    \\
    & 1 & {\it PNU} & Attribution-NonCommercial-ShareAlike 4.0 International & \href{https://jgamper.github.io/PanNukeDataset/}{PanNuke}~\cite{pannuke, pannuke-dataset}
    \\
    & 2 & {\it PRT} & CC BY-SA 3.0 & \href{https://proteinatlas.org}{Protein Atlas}~\cite{thul2017subcellular} 
    \\
    
    \midrule
    
    \multirow{3}{*}{Remote Sensing} 
    & 0 & {\it RESISC} & CC-BY-NC 4.0 & \href{https://gcheng-nwpu.github.io/}{RESISC45}~\cite{resisc}
    \\
    & 1 & {\it RSICB} & CC BY-NC 4.0 & \href{https://github.com/lehaifeng/RSI-CB}{RSICB128}~\cite{rsicb}
    \\
    & 2 & {\it RSD} & CC BY-NC 4.0 & \href{https://github.com/RSIA-LIESMARS-WHU/RSD46-WHU}{RSD46}~\cite{rsd46classify, rsd46localize}
    \\
    \midrule
    
    \multirow{3}{*}{Vehicles} 
    & 0 & {\it CRS} & ImageNet License & \href{https://ai.stanford.edu/~jkrause/cars/car_dataset.html}{Cars}~\cite{cars}
    \\
    & 1 & {\it APL} & Creative Commons Attribution 4.0 International & \href{https://zenodo.org/record/3464319}{Multi-type Aircraft}~\cite{airplanes}
    \\
    & 2 & {\it BTS} & CC BY-NC 4.0 & \href{https://github.com/avaapm/marveldataset2016}{MARVEL}~\cite{boats}
    \\
    \midrule
    
    \multirow{3}{*}{Manufacturing} 
    & 0 & {\it TEX} & CC BY-NC 4.0 & \href{https://www.csc.kth.se/cvap/databases/kth-tips/index.html}{KTH-TIPS}~\cite{Fritz2004THEKD, Mallikarjuna2006THEK2} \href{https://www.cb.uu.se/~gustaf/texture/}{Kylberg}~\cite{Kylberg2011c} \href{https://github.com/abin24/Textures-Dataset}{UIUC}~\cite{lazebnik:inria-00548530}
    \\
    & 1 & {\it TEX\_DTD} & CC BY-NC 4.0 & \href{https://www.robots.ox.ac.uk/~vgg/data/dtd/}{Texture DTD}~\cite{cimpoi2013describing}
    \\
    & 2 & {\it TEX\_ALOT} & CC BY-NC 4.0 & \href{https://aloi.science.uva.nl/public_alot/}{Texture ALOT}~\cite{texture_alot}
    \\
    \midrule

    \multirow{3}{*}{Human Actions} 
    & 0 & {\it SPT} & CC0 1.0 Public Domain & \href{https://www.kaggle.com/gpiosenka/sports-classification}{100 Sports}~\cite{100-sports}
    \\
    & 1 & {\it ACT\_40} & CC BY-NC 4.0 & \href{http://vision.stanford.edu/Datasets/40actions.html}{Stanford 40 Actions}~\cite{act-40}
    \\
    & 2 & {\it ACT\_410} & Simplified BSD License & \href{http://human-pose.mpi-inf.mpg.de/#download}{MPII Human Pose}~\cite{act-410}
    \\

    \midrule
    
    \multirow{3}{*}{Optical Char. Recog.} 
    & 0 & {\it MD\_MIX} & CC BY 4.0 & 
    \\
    & 1 & {\it MD\_5\_BIS} & CC BY 4.0 & \href{https://github.com/SunHaozhe/OmniPrint-datasets}{OmniPrint}~\cite{sun2021omniprint}
    \\
    & 2 & {\it MD\_6} & CC BY 4.0 & 
    \\
    \bottomrule

  \end{tabular}
  \end{adjustbox}
\end{table}

\clearpage
\newpage

\section{Data preparation}
\label{appendix:data_preparation}

Data preparation process is done in many steps which include, depending on the dataset:
\begin{itemize}
  \item Visual inspection and data cleaning 
  \item Cropping images to a square around the region of interest (eventually background padding)
  \item Resizing squared images to 128x128 size (with anti-aliasing filter)
  \item Sub-sampling (to obtain the Mini and Micro versions).
\end{itemize}

\subsection{Visual inspection and data cleaning}

The goal of this step is to remove non-suitable images.

Most Meta-Album datasets were downloaded from the Internet; a few were directly obtained from their original owners/creators.
Cleaning steps included the removal of duplicate images, corrupted images, and images that included watermarks or identifiers. After cleaning, each dataset is saved in a folder including a sub-folder with the images, in a standard jpg format, and a meta-data file including file names, categories/classes/labels, super-categories, etc. 

\subsection{Image cropping}

The goal of the step is to obtain square images.

Most of the datasets come with images of different sizes and resolutions. Usually, the object of interest (\eg bird, leaf, car, etc.) is already mostly centered in the middle of the image, with comfortable margins on all sides. In this case, the image is cropped to obtain a square by keeping the smallest dimension (width or height) and eliminating symmetrically spurious margins in the largest dimension.

Some datasets, \eg Plankton and Boats, are preprocessed in a different way because the rectangular images tightly fit the object of interest. 
In that case, the images are padded with extra background or either side in the smallest dimension, to obtain a square of the largest dimension.
The padding is obtained from the 3 extreme rows (or columns) on either side, which is smoothed with a Gaussian kernel of size 29x29.
We found that this technique minimally introduces image artifacts.

Some of the datasets, \eg RESISC and Textures datasets, are already square, hence the only preprocessing needed is resizing (see next sub-section).  All OCR datasets are generated by the OmniPrint software~\cite{sun2021omniprint}. The images are exactly in 128x128 size and hence no preprocessing is required at all. 

Datasets in Human Actions domain are preprocessed in a different way as they contain humans performing some actions, \eg playing a sport, etc. For these datasets, a face detection software is used to detect a face in the image and then a reasonable area around the face is cropped in order to make sure that the human body along with the face is cropped. After cropping, like other datasets, the images are resized into 128x128 pixel sizes.

\begin{figure}[htb]
    \centering

    \subfloat[Insects ]{\includegraphics[width=0.46\textwidth]{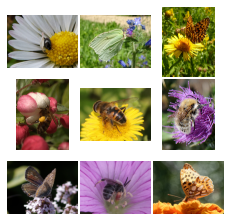}}
    \subfloat[Plankton]{\includegraphics[width=0.46\textwidth]{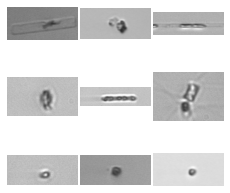}}
    \\
    \subfloat[Flowers ]{\includegraphics[width=0.46\textwidth]{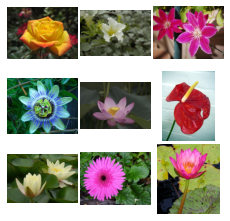}}
    \subfloat[Texture DTD ]{\includegraphics[width=0.46\textwidth]{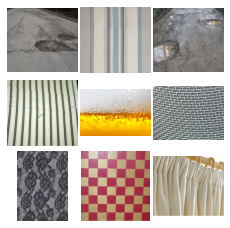}}
    
    \caption{Meta-Album datasets raw sample images.}
    \label{fig:raw-sampleimages}
\end{figure}

See Figure~\ref{fig:raw-sampleimages} for sample raw images from some of the datasets.

\begin{figure}[th]
\centering
\includegraphics[width=0.8\textwidth]{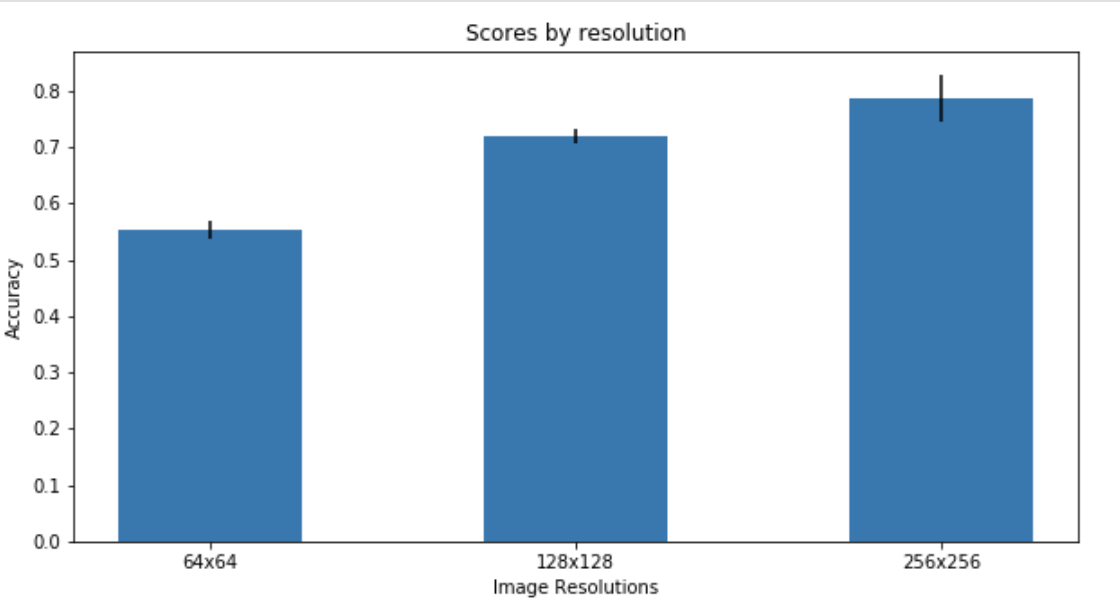}
\caption{Image resolution comparison (64x64 vs 128x128 vs 256x256).}
\label{fig:resolution_comparison}
\end{figure}

\begin{figure}[th]
\centering
\includegraphics[width=0.8\textwidth]{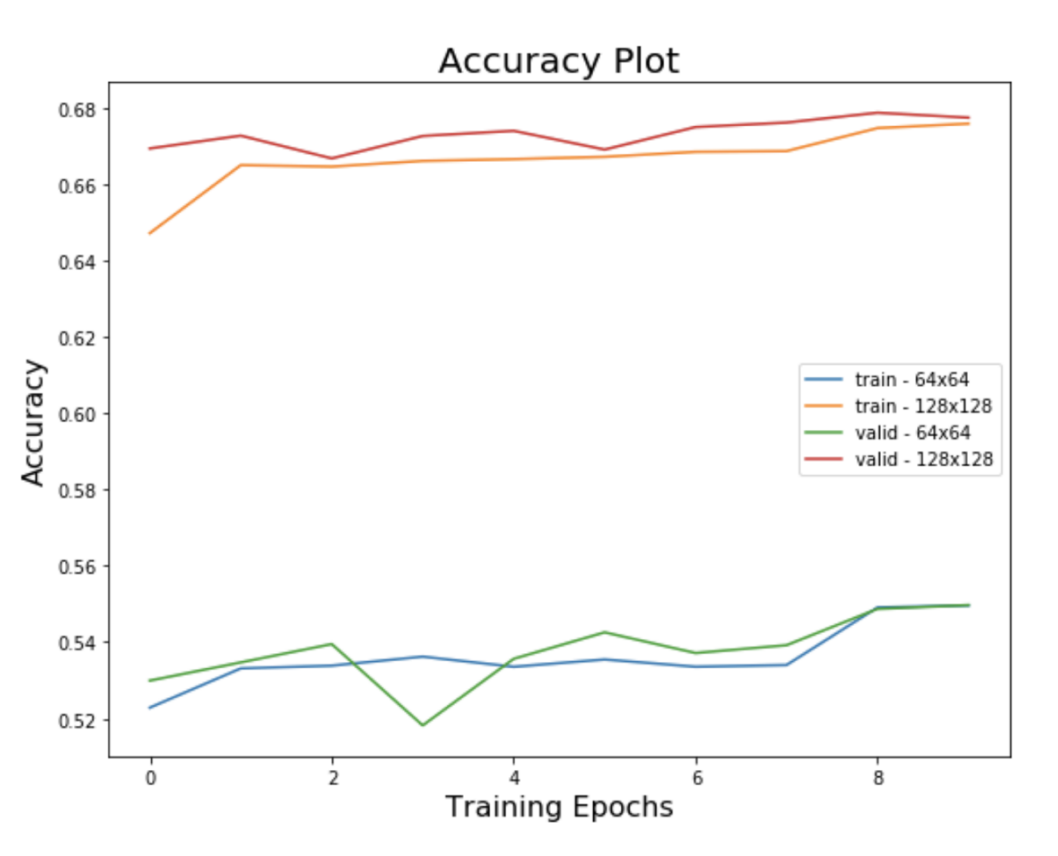}
\caption{Image resolution comparison (64x64 vs 128x128 vs 256x256).}
\label{fig:resolution_comparison_accuracy}
\end{figure}

\subsection{Image resizing}

The goal of this step is to reduce the image size and obtain images of the same dimension for all datasets.

Except for the OCR datasets (images already in the correct size), all other dataset images are resized to 128x128 using Open-CV~\cite{opencv_library}, with an anti-aliasing filter. 
This avoids the aliasing effect, often visible in the form of jagged edges, introduced by the loss of information during down-sampling. 

The image resolution 128x128 is chosen based on preliminary experiments and analyses done for different resolutions of images on the insect dataset, presenting fine features necessary for recognition to the human eye: 64x64, 128x128 and 256x256. For this experiment (See Figure~\ref{fig:resolution_comparison}), a ResNet 152V2 architecture from Keras is used. It can be observed that 256x256 gives the highest scores among the observed resolutions but the dataset size also increases by a factor of 2. The images with resolution 128x128 are observed to be clearly recognizable to human eyes.

\begin{figure}[htb]
\centering

    \subfloat[Sample 128x128 images ]{\includegraphics[width=\textwidth]{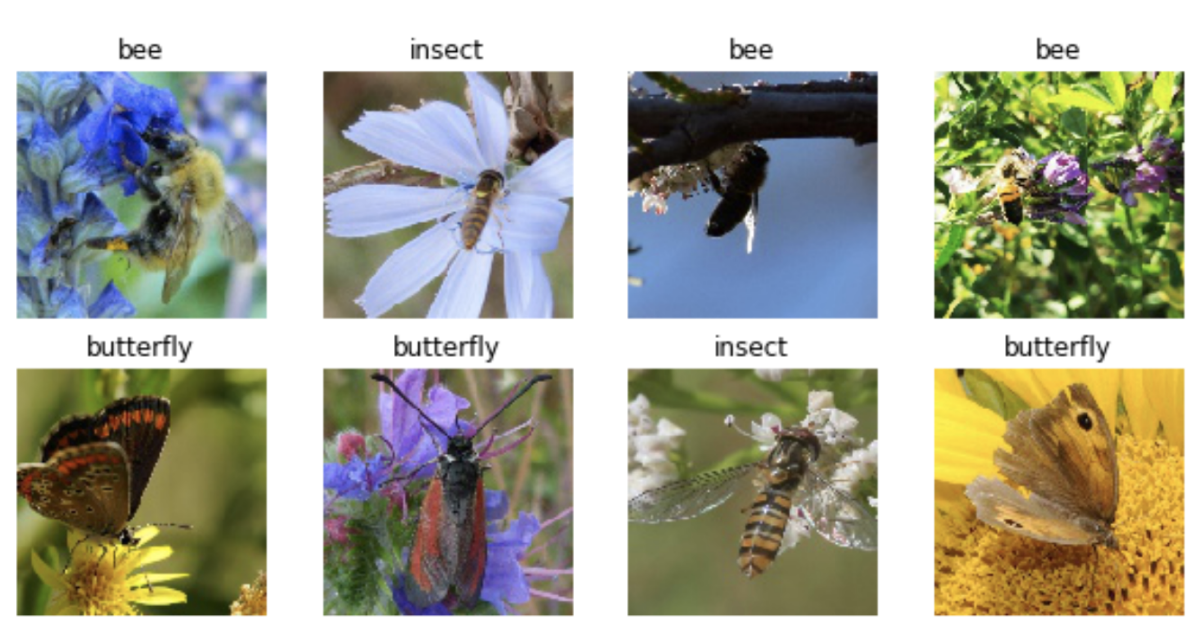}}
    \\
    \subfloat[Sample 64x64 images ]{\includegraphics[width=\textwidth]{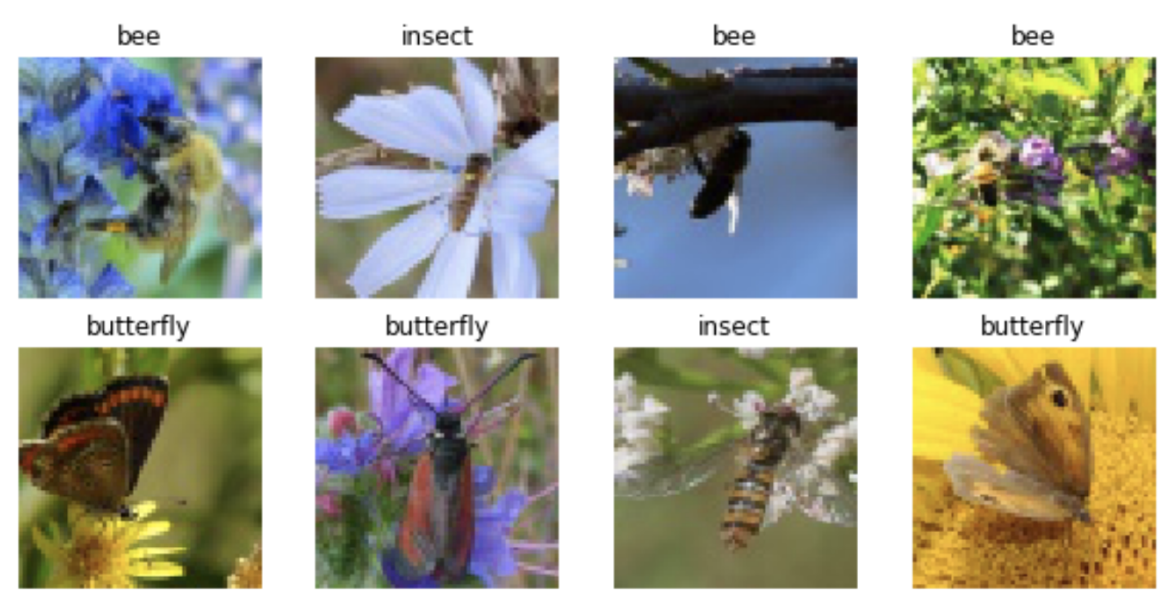}}
    
\caption{Sample images in 128x128 and 64x64 resolutions.}
\label{fig:resolution_comparison_samples}
\end{figure}

Another experiment on the 
insect dataset, comparing 64x64 and 128x128 resolution, is also carried out. The number of images used for the experiment is 180\,000, 135\,000 of which are used for training, and  the remaining for validation. The accuracy plot in Figure~\ref{fig:resolution_comparison_accuracy} shows clearly that 128x128 produces much better results.  The experimental settings for this experiment include: ResNet-18 architecture, SGD optimizer, learning rate of 0.001 and 10 epochs. The better quality of 128x128 images is also clearly visually observable (See Figure~\ref{fig:resolution_comparison_samples}).

\subsection{Sub-sampling}

The goal of this step is to keep only classes sufficiently populated (more than 40 examples per class) and to balance the dataset (even number of examples per class).

After the previous steps, we obtain what we call the ``Extended version'' of Meta-Album. The preprocessed datasets in this version conserve the proportions of samples of the original data, with different number of classes and images per class.

To facilitate experimentation and reduce the need for large computational resources, we have prepared the Mini version, including all classes with at least 40 images/class, but only 40 randomly chosen images from each class. We have also prepared the Micro version, including only 20 randomly selected classes from the Mini version, with 40 images per class.

The software to prepare the datasets, \ie add backgrounds, crop images, resize images and extract a subset from a dataset, is available in the Meta-Album GitHub Repository (\url{https://github.com/ihsaan-ullah/meta-album}).

\subsection{Quality control}
In order to ensure a high quality of datasets, we put a strong emphasis on quality control. We describe two ways to ensure this quality: factsheets and GradCam visualizations. We also define an easy to read Data format for all our datasets.

\subsubsection*{Data format}

In order to generate 
\href{https://github.com/ihsanullah2131/meta-album/tree/master/Factsheets}{Factsheets} for quality control, the data is first formatted in the recommended \href{https://github.com/ihsanullah2131/meta-album/tree/master/DataFormat}{Data format}. This makes the data easy-to-read and understand. The data format has a root directory (ideally with the dataset name and it consists of an \textit{images} folder with all the required images for the fact-sheet generation, a \textit{labels.csv} file with at least two columns, \ie \textit{FILE\_NAME} and \textit{CATEGORY} (category is synonym of class/label) which shows the name of the image instance and the label of that image respectively. One more important file is a \textit{info.json} file in the root directory, which has important information about the dataset and how the data should be read for factsheets. The Data format repository has \href{https://github.com/ihsaan-ullah/meta-album/tree/master/DataFormat/mini_insect_1}{sample dataset} for reference. To verify the data format, a \href{https://github.com/ihsanullah2131/meta-album/blob/master/DataFormat/check_data_format.py}{python script} is also provided and complete details of the formatting and testing are given in the README of the data format repository.

\subsubsection*{Factsheets}

For each dataset, we generate a factsheet which is an automatically generated report showing basic information and experimental results on the dataset. 
These experiments are run on the formatted datasets (Mini version).
The experiments are flexible and the configuration can be changed easily by changing the parameters of the software. 
For each dataset, we train a randomly initialized ResNet-18 architecture~\cite{resnet} on various 5-way 20-shot classification tasks, \ie the network is presented with training sets consisting of 20 images for each of the 5 classes. 
The remaining 20 images per class are used as query set to measure the generalization ability of the network. 
Note that the network is trained ``from scratch'' (randomly initialized weights), on every task. 
From the experiments, we construct ROC curves for every task for all classes. 

Classes that are too hard to separate (indicating non-learnable tasks) can be identified easily with this experiment. AUC score below 0.7 indicates the difficulty in separation and hence could be studied further to construct a fair task.

In addition to this automatic quality check, we also generate Imagesheets for the datasets, which are PDF files with images of every class. This way, we can also check the data quality visually.

\subsubsection*{GradCam visualization}

In order to detect potential bias stemming from the data and our curating process, we inspect where the baseline model focuses with GradCAM~\cite{gradcam}. GradCAM is a post-hoc attention method, showing the contribution of each input element to the output. 

\begin{figure}[tbh]
\centering
\includegraphics[width=\textwidth]{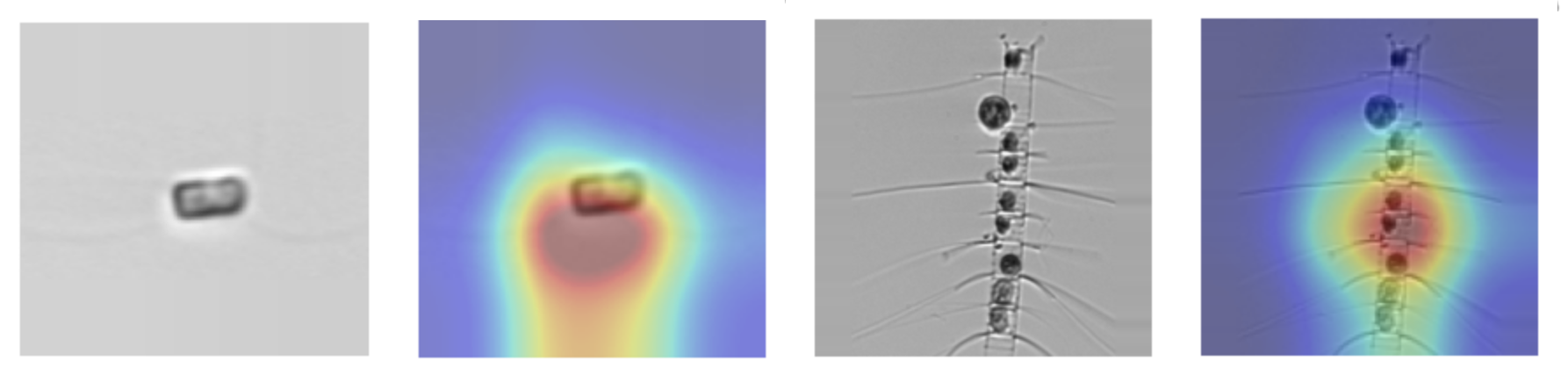}
\caption{GradCAM activation maps to detect artifacts that could be introduced by image preprocessing.}
\label{fig:gradcamplankton}
\end{figure}

By displaying the activation map on top of the original input image, we can observe the impact of each pixel on the prediction output.
As such, we can verify whether the model focuses on the relevant subject.
In this way, it can also be used to detect whether artifacts are introduced by the image processing in the picture or on the background and other artifacts. For example, we use GradCAM for detecting any artifacts introduced by preprocessing the backgrounds of plankton images to make squared images using padding (Figure~\ref{fig:gradcamplankton}). 
Other methods should also be considered to diagnose data~\cite{visual_interpretability_survey}.

\clearpage
\newpage
\section{Dataset difficulty analysis} \label{appendix:dataset-difficulty}

\begin{figure}[H]
\centering

\begin{subfigure}{.5\textwidth}
  \centering
  \includegraphics[width=\textwidth]{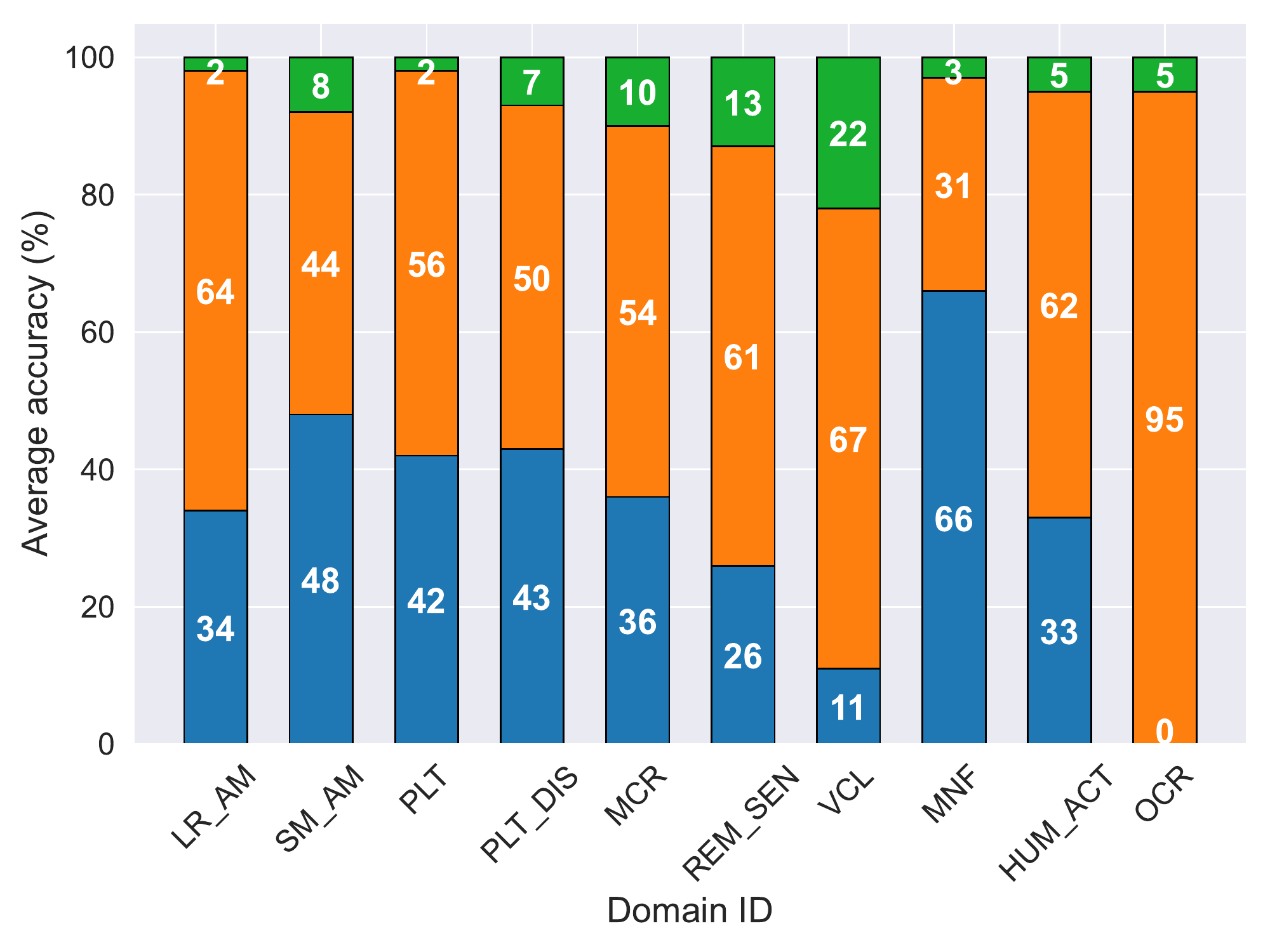}
 \caption{\label{fig:diff-set-0} Set 0}
\end{subfigure}%
\begin{subfigure}{.5\textwidth}
  \centering
  \includegraphics[width=\textwidth]{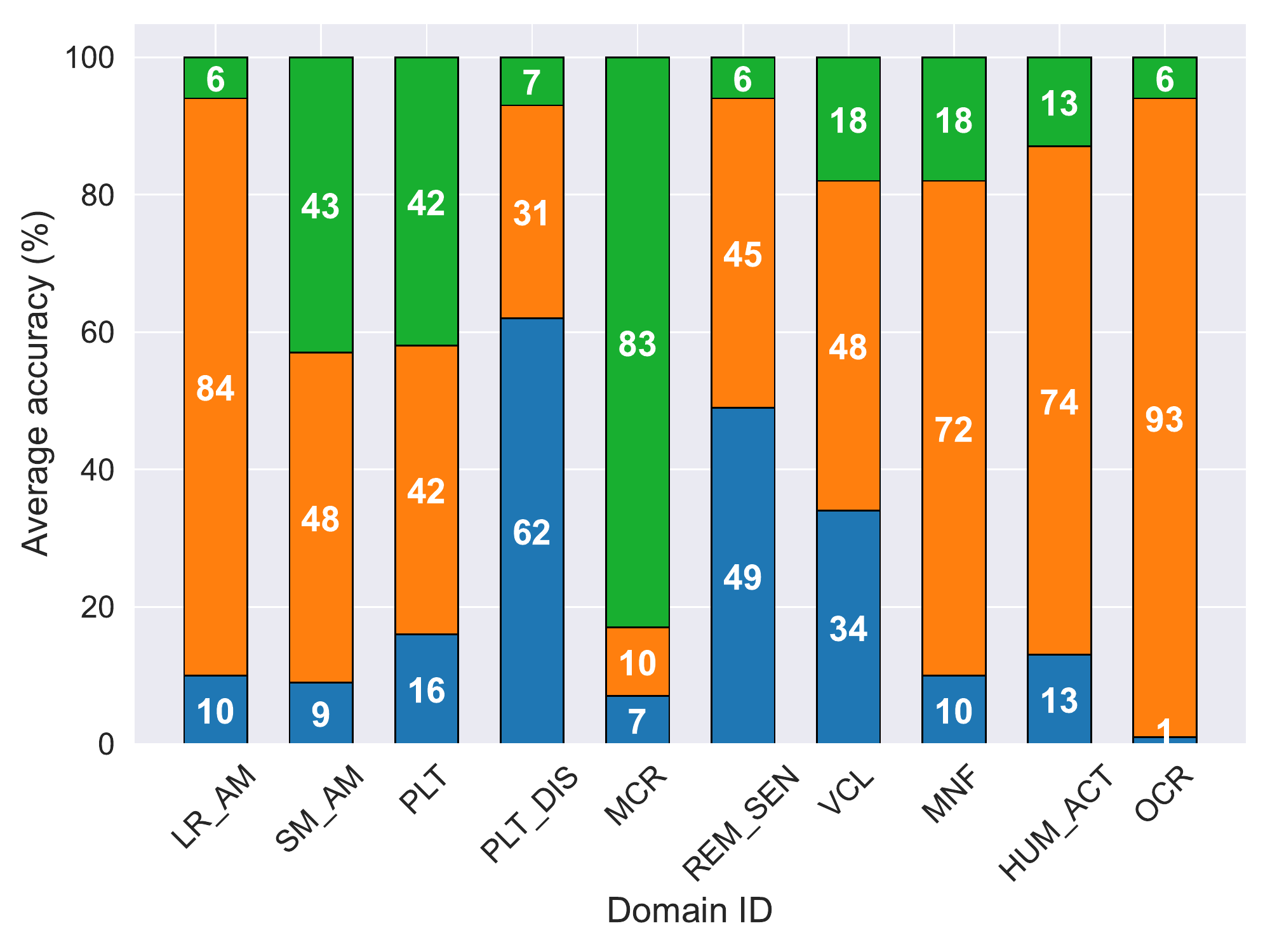}
 \caption{\label{fig:diff-set-1} Set 1}
\end{subfigure}
\\
\begin{subfigure}{.5\textwidth}
  \centering
  \includegraphics[width=\textwidth]{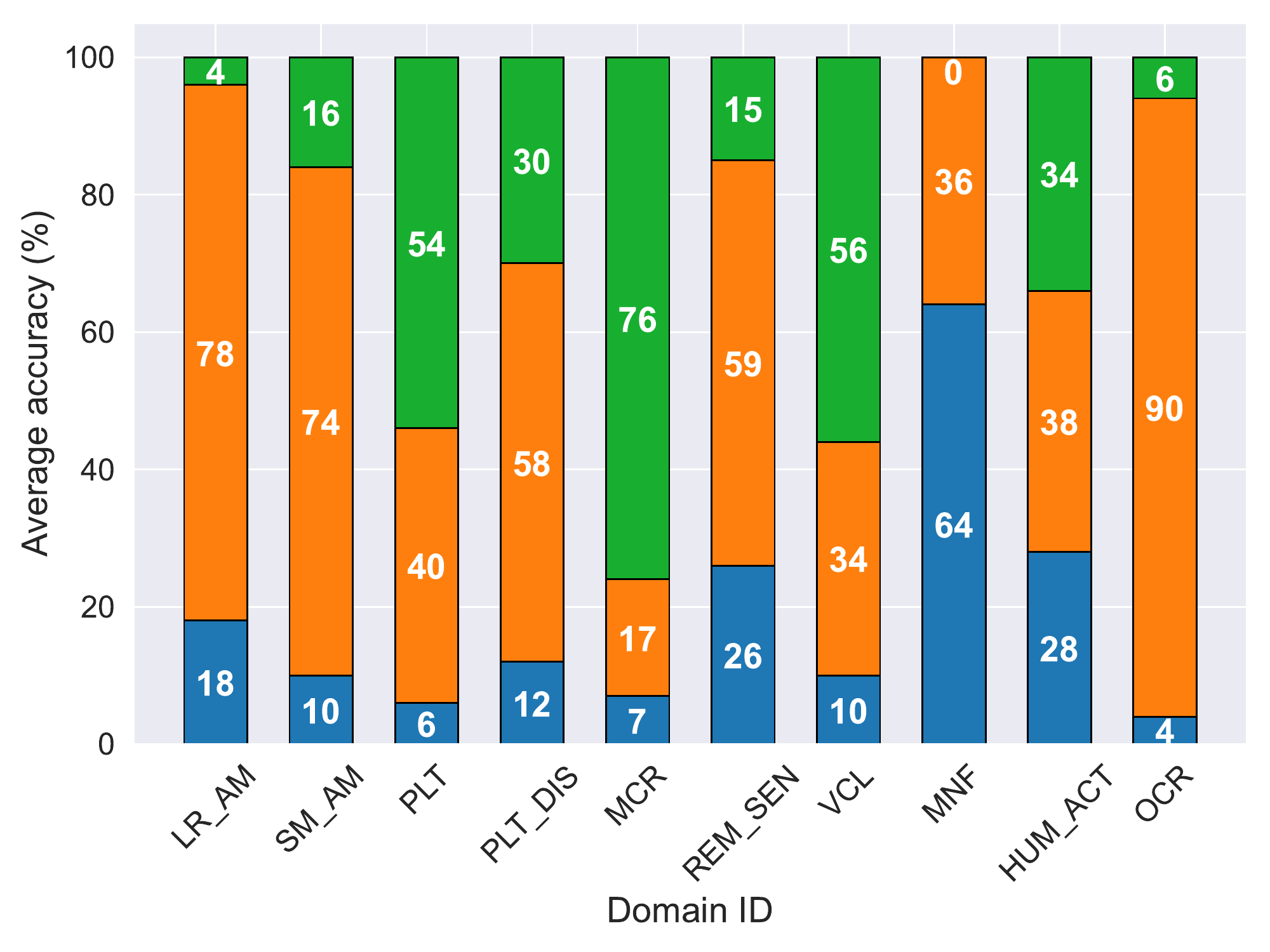}
 \caption{\label{fig:diff-set-2} Set 2}
\end{subfigure}%
\begin{subfigure}{.5\textwidth}
  \centering
  \includegraphics[width=\textwidth]{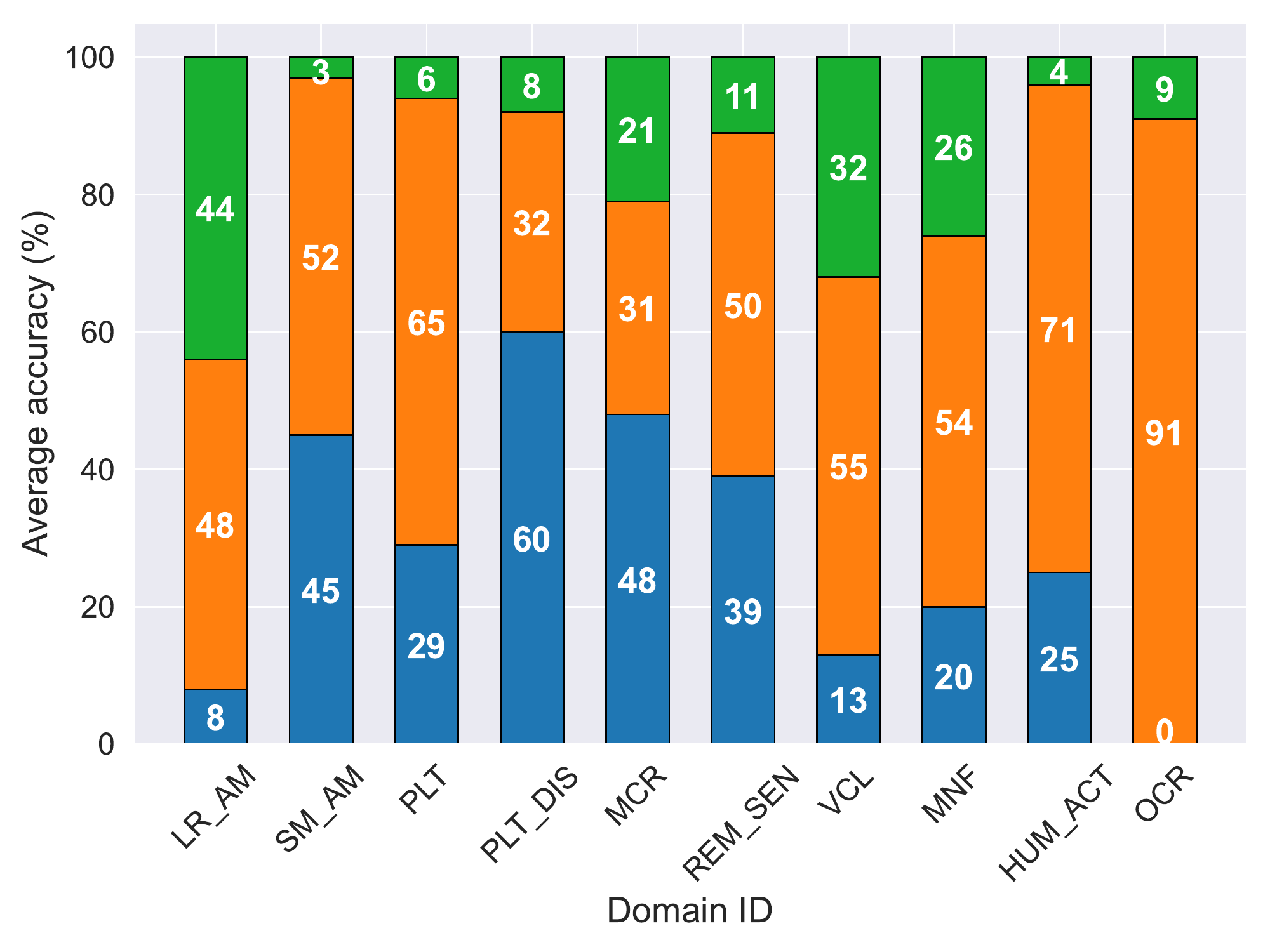}
 \caption{\label{fig:diff-set-3} Set 3}
\end{subfigure}

\caption{Difficulty analysis for 4 sets (40 datasets) of Meta-Album. Each set consists of 10 datasets from 10 domains (one dataset per domain).}
\label{fig:difficulty}
\end{figure}

In this section, we analyze the difficulty per domain of each set of Meta-Album datasets following the NeurIPS 2021 MetaDL challenge protocol (within domain setting). In Figure~\ref{fig:diff-set-0}, Figure~\ref{fig:diff-set-1}, Figure~\ref{fig:diff-set-2}, and Figure~\ref{fig:diff-set-3}, the top of the blue bar indicates the worst baseline performance (TrainFromScratch). The top of the orange bar indicates the performance of the winners of the NeurIPS 2021 challenge (MetaDelta++)~\cite{elbaz2022lessons}. The top of the green bar indicates the maximum achievable performance. The larger the green bar, the larger the \textit{ intrinsic difficulty}. The larger the orange bar, the larger the \textit{ modeling difficulty}.

\textbf{Experimental protocol.}  Each dataset was evaluated independently by splitting it into meta-training, meta-validation, and meta-testing sets with non-overlapping classes, as explained in Section 3.1. Then, for each dataset, the MetaDelta++ method was meta-trained for 30 minutes with batches of size 64. The trainers produced by MetaDelta++ were meta-validated every 50 batches on 50 5-way 5-shot tasks drawn from the corresponding meta-validation split. On the other hand, as explained in Section 3.2, the TrainFromScratch method has no meta-training and meta-validation phases; instead, it learns each task at meta-test time. Lastly, the learning algorithm with the best meta-validation performance was meta-tested on 100 5-way 5-shot tasks randomly sampled from the meta-testing split. The query set of each meta-test task contained 20 images per class. As described in Section 3.2, we averaged the results over 3 runs with different random seeds.

The experimental results show that in some domains like OCR, there is always a large modeling difficulty, as evidenced by the large orange bar, which makes them desirable to be used in challenges. On the other hand, the microscopy datasets (MCR) in Sets 1 and 2 have a large intrinsic difficulty (large green bar): even the winner's method does not get good results which may indicate that the data of these datasets are of insufficient quality. Moreover, some datasets are relatively easy even for the TrainFromScratch method, such as the manufacturing datasets (MNF) of Set 0 or 2. Finally, on average, Set 0 is the easiest while the hardest is Set 2.

\clearpage
\newpage
\section{Within Domain Few-shot learning: additional results}
\label{appendix:fslappendix}

In this section, we show additional results for the within domain few-shot learning experiments that were presented in Section 3.2.
Table~\ref{tab:app1shot}, Table~\ref{tab:app5shot}, Table~\ref{tab:app10shot}, and Table~\ref{tab:app20shot} display the average accuracy per technique and dataset in the [1, 5, 10, 20]-shot settings, respectively. 
Note that the 1-shot performance of matching networks and prototypical networks is not the same as they use different distance measures. More specifically, matching networks use cosine similarity while prototypical networks use squared euclidean distance.

\begin{figure}[t]
 \centering
 \begin{subfigure}{0.48\textwidth}
     \centering
     \includegraphics[width=\textwidth]{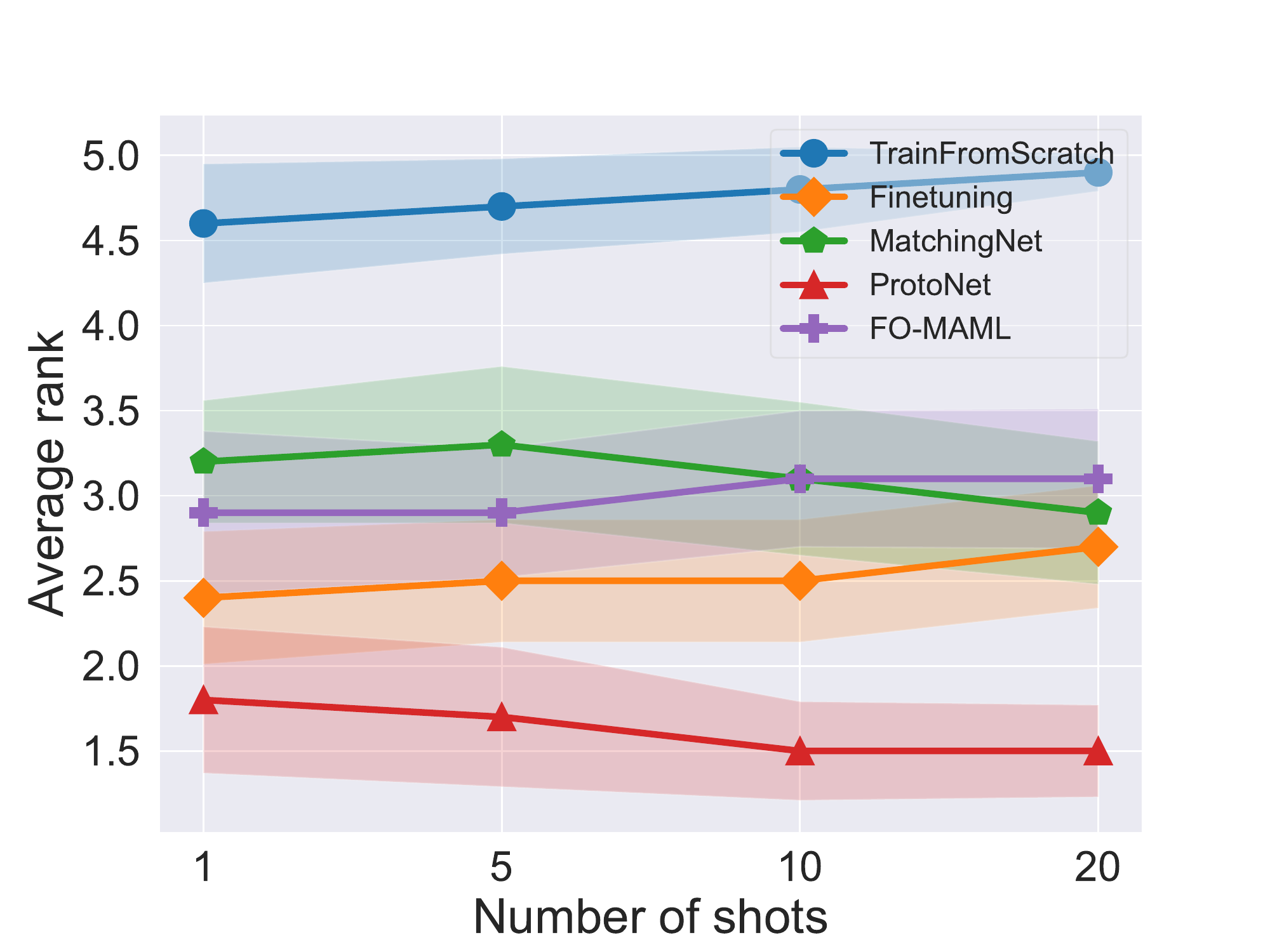}
     \caption{Average rankings}
 \end{subfigure}
 \hfill
 \begin{subfigure}{0.48\textwidth}
     \centering
     \includegraphics[width=\textwidth]{fig/fsl_experiment/fsl-avgaccs.pdf}
     \caption{Average accuracy}
 \end{subfigure}
 \caption{\label{fig:appfsl-plots} {\bf Within domain few-shot learning.} Averages over 30 released Meta-Album datasets of: (a) algorithm rank (smaller is better). The 95\% confidence intervals are computed at dataset level over 30 datasets. (b) classification accuracy. The 95\% confidence intervals are computed at per-task level over 3 runs per dataset with 600 tasks per run (total tasks: $3 \times 600 \times 30 = 54\,000$).}
\end{figure}

Table~\ref{tab:apprankings} displays the average ranks and running times per technique in different within domain settings.
The rankings follow roughly the same pattern as in the right subplot of Figure~\ref{fig:appfsl-plots}.
Note that the confidence intervals for the average ranks are larger than for the average accuracy as the former is computed over the 30 datasets, while the latter is computed at per-task level over all datasets. 
Figure~\ref{fig:appfsl-plots} displays the average rank per method and average accuracy for within domain 5-way [1, 5, 10, 20]-shot classification.

\begin{table}[ht]
\centering
\caption{Full performance results for 5-way 1-shot image classification on all datasets. The 95\% confidence intervals are computed at per-task level over 3 runs per dataset with 600 tasks per run (total tasks: $3\times600=1\,800$).}
\label{tab:app1shot}
 \begin{adjustbox}{width=\linewidth}
\begin{tabular}{lllllll}
\toprule
Domain & Dataset ID  &           TrainFromScratch &       Finetuning &      MatchingNet &         ProtoNet &             FO-MAML \\
\midrule 
\multirow{3}{*}{Large animals} & BRD         &  31.4 $\pm$ 0.37 &  62.7 $\pm$ 0.51 &  52.7 $\pm$ 0.59 &  67.7 $\pm$ 0.53 &  60.0 $\pm$ 0.58 \\
& DOG          &  24.6 $\pm$ 0.27 &  31.7 $\pm$ 0.36 &  30.2 $\pm$ 0.42 &  33.6 $\pm$ 0.40 &  29.5 $\pm$ 0.38 \\
& AWA           &  26.5 $\pm$ 0.31 &  32.5 $\pm$ 0.35 &  32.3 $\pm$ 0.42 &  31.7 $\pm$ 0.36 &  29.7 $\pm$ 0.42 \\
\midrule 
\multirow{3}{*}{Small animals} & PLK      &  44.5 $\pm$ 0.53 &  60.0 $\pm$ 0.57 &  55.8 $\pm$ 0.60 &  66.7 $\pm$ 0.59 &  59.1 $\pm$ 0.58 \\
& INS\_2      &  23.8 $\pm$ 0.28 &  27.4 $\pm$ 0.31 &  26.8 $\pm$ 0.36 &  28.2 $\pm$ 0.35 &  24.7 $\pm$ 0.31 \\
& INS       &  22.6 $\pm$ 0.25 &  34.5 $\pm$ 0.40 &  35.0 $\pm$ 0.49 &  36.1 $\pm$ 0.45 &  31.8 $\pm$ 0.51 \\
\midrule 
\multirow{3}{*}{Plants} & FLW       &  35.4 $\pm$ 0.44 &  59.7 $\pm$ 0.51 &  48.3 $\pm$ 0.56 &  63.0 $\pm$ 0.49 &  55.0 $\pm$ 0.57 \\
& PLT\_NET      &  24.8 $\pm$ 0.29 &  31.5 $\pm$ 0.35 &  33.9 $\pm$ 0.49 &  30.5 $\pm$ 0.37 &  35.2 $\pm$ 0.41 \\
& FNG         &  23.8 $\pm$ 0.25 &  23.8 $\pm$ 0.25 &  22.4 $\pm$ 0.24 &  23.7 $\pm$ 0.26 &  22.9 $\pm$ 0.26 \\
\midrule 
\multirow{3}{*}{Plant diseases} & PLT\_VIL   &  49.5 $\pm$ 0.47 &  69.0 $\pm$ 0.39 &  50.0 $\pm$ 0.46 &  64.1 $\pm$ 0.41 &  55.6 $\pm$ 0.42 \\
& MED\_LF &  57.8 $\pm$ 0.46 &  63.0 $\pm$ 0.42 &  60.3 $\pm$ 0.39 &  68.4 $\pm$ 0.46 &  66.7 $\pm$ 0.43 \\
& PLT\_DOC      &  21.5 $\pm$ 0.23 &  23.1 $\pm$ 0.25 &  22.9 $\pm$ 0.26 &  24.1 $\pm$ 0.26 &  23.8 $\pm$ 0.27 \\
\midrule 
\multirow{3}{*}{Microscopy} & BCT      &  35.3 $\pm$ 0.41 &  65.7 $\pm$ 0.45 &  70.0 $\pm$ 0.53 &  75.5 $\pm$ 0.43 &  58.7 $\pm$ 0.68 \\
& PNU       &  24.2 $\pm$ 0.29 &  23.0 $\pm$ 0.24 &  23.3 $\pm$ 0.25 &  22.9 $\pm$ 0.25 &  24.2 $\pm$ 0.27 \\
& PRT           &  22.0 $\pm$ 0.24 &  25.3 $\pm$ 0.27 &  27.9 $\pm$ 0.36 &  26.0 $\pm$ 0.30 &  28.0 $\pm$ 0.37 \\
\midrule 
\multirow{3}{*}{Remote sensing} & RESISC        &  31.9 $\pm$ 0.39 &  44.0 $\pm$ 0.44 &  37.4 $\pm$ 0.46 &  42.2 $\pm$ 0.43 &  42.1 $\pm$ 0.50 \\
& RSICB         &  47.0 $\pm$ 0.47 &  62.9 $\pm$ 0.46 &  50.2 $\pm$ 0.48 &  64.2 $\pm$ 0.46 &  60.2 $\pm$ 0.50 \\
& RSD          &  33.7 $\pm$ 0.39 &  43.7 $\pm$ 0.44 &  45.1 $\pm$ 0.58 &  49.9 $\pm$ 0.46 &  51.6 $\pm$ 0.50 \\
\midrule 
\multirow{3}{*}{Vehicles} & CRS          &  23.0 $\pm$ 0.25 &  48.9 $\pm$ 0.48 &  52.4 $\pm$ 0.61 &  63.5 $\pm$ 0.56 &  53.1 $\pm$ 0.64 \\
& APL     &  29.9 $\pm$ 0.31 &  36.4 $\pm$ 0.39 &  34.0 $\pm$ 0.40 &  41.2 $\pm$ 0.43 &  33.1 $\pm$ 0.37 \\
& BTS         &  23.6 $\pm$ 0.25 &  26.5 $\pm$ 0.28 &  23.8 $\pm$ 0.29 &  26.2 $\pm$ 0.29 &  23.7 $\pm$ 0.28 \\
\midrule 
\multirow{3}{*}{Manufacturing} & TEX     &  59.8 $\pm$ 0.59 &  81.2 $\pm$ 0.46 &  72.8 $\pm$ 0.68 &  84.1 $\pm$ 0.50 &  84.5 $\pm$ 0.51 \\
& TEX\_DTD   &  24.4 $\pm$ 0.28 &  28.6 $\pm$ 0.34 &  26.3 $\pm$ 0.29 &  27.3 $\pm$ 0.32 &  27.6 $\pm$ 0.31 \\
& TEX\_ALOT  &  57.0 $\pm$ 0.59 &  89.8 $\pm$ 0.38 &  93.1 $\pm$ 0.38 &  96.0 $\pm$ 0.23 &  93.5 $\pm$ 0.35 \\
\midrule 
\multirow{3}{*}{Human actions} & SPT           &  31.3 $\pm$ 0.41 &  45.4 $\pm$ 0.46 &  42.8 $\pm$ 0.57 &  48.2 $\pm$ 0.50 &  47.4 $\pm$ 0.55 \\
& ACT\_40         &  22.2 $\pm$ 0.23 &  24.8 $\pm$ 0.25 &  25.5 $\pm$ 0.30 &  24.7 $\pm$ 0.26 &  25.2 $\pm$ 0.30 \\
& ACT\_410        &  30.6 $\pm$ 0.36 &  33.2 $\pm$ 0.36 &  24.9 $\pm$ 0.29 &  27.6 $\pm$ 0.30 &  25.5 $\pm$ 0.30 \\
\midrule 
\multirow{3}{*}{OCR} & MD\_MIX        &  20.3 $\pm$ 0.19 &  40.8 $\pm$ 0.56 &  35.4 $\pm$ 0.78 &  89.0 $\pm$ 0.37 &  20.0 $\pm$ 0.15 \\
& MD\_5\_BIS      &  20.2 $\pm$ 0.20 &  42.6 $\pm$ 0.44 &  69.6 $\pm$ 0.66 &  90.8 $\pm$ 0.35 &  20.2 $\pm$ 0.20 \\
& MD\_6          &  21.2 $\pm$ 0.22 &  79.7 $\pm$ 0.58 &  72.3 $\pm$ 0.65 &  92.1 $\pm$ 0.31 &  20.3 $\pm$ 0.21 \\
\bottomrule
\end{tabular}
\end{adjustbox}
\end{table}

\begin{table}[ht]
\centering
\caption{Full performance results for 5-way 5-shot image classification on all datasets. The 95\% confidence intervals are computed at per-task level over 3 runs per dataset with 600 tasks per run (total tasks: $3\times600=1\,800$).}
\label{tab:app5shot}
 \begin{adjustbox}{width=\linewidth}
\begin{tabular}{lllllll}
\toprule
Domain & Dataset ID  &           TrainFromScratch &       Finetuning &      MatchingNet &         ProtoNet &             FO-MAML \\
\midrule 
\multirow{3}{*}{Large animals} &
BRD         &  45.4 $\pm$ 0.43 &  80.5 $\pm$ 0.36 &  72.0 $\pm$ 0.41 &  82.4 $\pm$ 0.33 &  73.7 $\pm$ 0.42 \\
& DOG          &  30.0 $\pm$ 0.31 &  40.1 $\pm$ 0.36 &  42.9 $\pm$ 0.39 &  41.8 $\pm$ 0.37 &  34.9 $\pm$ 0.38 \\
& AWA           &  35.5 $\pm$ 0.32 &  42.6 $\pm$ 0.36 &  37.2 $\pm$ 0.33 &  45.2 $\pm$ 0.38 &  37.2 $\pm$ 0.36 \\
\midrule
\multirow{3}{*}{Small animals} & PLK      &  57.8 $\pm$ 0.50 &  73.9 $\pm$ 0.50 &  71.0 $\pm$ 0.50 &  76.7 $\pm$ 0.44 &  69.6 $\pm$ 0.49 \\
& INS\_2      &  28.3 $\pm$ 0.31 &  34.4 $\pm$ 0.32 &  29.1 $\pm$ 0.29 &  32.2 $\pm$ 0.31 &  31.4 $\pm$ 0.32 \\
& INS       &  25.3 $\pm$ 0.25 &  43.6 $\pm$ 0.39 &  44.7 $\pm$ 0.42 &  45.0 $\pm$ 0.43 &  43.3 $\pm$ 0.43 \\
\midrule
\multirow{3}{*}{Plants} & FLW       &  50.8 $\pm$ 0.46 &  76.2 $\pm$ 0.42 &  68.0 $\pm$ 0.42 &  79.5 $\pm$ 0.33 &  72.8 $\pm$ 0.44 \\
& PLT\_NET      &  31.1 $\pm$ 0.30 &  40.5 $\pm$ 0.32 &  36.7 $\pm$ 0.29 &  41.1 $\pm$ 0.33 &  38.6 $\pm$ 0.39 \\
& FNG         &  27.4 $\pm$ 0.26 &  27.6 $\pm$ 0.25 &  25.1 $\pm$ 0.25 &  29.1 $\pm$ 0.26 &  25.1 $\pm$ 0.25 \\
\midrule
\multirow{3}{*}{Plant diseases} & PLT\_VIL   &  68.6 $\pm$ 0.41 &  85.8 $\pm$ 0.25 &  58.8 $\pm$ 0.33 &  80.9 $\pm$ 0.24 &  70.2 $\pm$ 0.35 \\
& MED\_LF &  77.6 $\pm$ 0.35 &  79.9 $\pm$ 0.30 &  75.3 $\pm$ 0.30 &  83.6 $\pm$ 0.24 &  85.6 $\pm$ 0.22 \\
& PLT\_DOC      &  23.4 $\pm$ 0.23 &  27.5 $\pm$ 0.25 &  27.8 $\pm$ 0.26 &  26.3 $\pm$ 0.27 &  26.5 $\pm$ 0.26 \\
\midrule
\multirow{3}{*}{Microscopy} & BCT      &  51.1 $\pm$ 0.42 &  78.2 $\pm$ 0.36 &  78.6 $\pm$ 0.41 &  82.8 $\pm$ 0.34 &  80.4 $\pm$ 0.40 \\
& PNU       &  28.5 $\pm$ 0.27 &  23.0 $\pm$ 0.21 &  26.5 $\pm$ 0.23 &  26.9 $\pm$ 0.24 &  31.0 $\pm$ 0.27 \\
& PRT           &  25.1 $\pm$ 0.25 &  31.0 $\pm$ 0.25 &  31.5 $\pm$ 0.28 &  30.3 $\pm$ 0.28 &  33.3 $\pm$ 0.30 \\
\midrule
\multirow{3}{*}{Remote sensing} & RESISC        &  41.4 $\pm$ 0.38 &  57.5 $\pm$ 0.36 &  48.7 $\pm$ 0.33 &  62.7 $\pm$ 0.37 &  61.7 $\pm$ 0.35 \\
& RSICB         &  60.5 $\pm$ 0.47 &  81.9 $\pm$ 0.33 &  77.2 $\pm$ 0.29 &  84.4 $\pm$ 0.26 &  79.9 $\pm$ 0.35 \\
& RSD          &  39.9 $\pm$ 0.36 &  57.1 $\pm$ 0.41 &  56.2 $\pm$ 0.43 &  61.2 $\pm$ 0.39 &  49.4 $\pm$ 0.31 \\
\midrule
\multirow{3}{*}{Vehicles} & CRS          &  28.3 $\pm$ 0.28 &  65.3 $\pm$ 0.42 &  72.4 $\pm$ 0.45 &  70.2 $\pm$ 0.41 &  71.2 $\pm$ 0.48 \\
& APL     &  40.1 $\pm$ 0.37 &  47.6 $\pm$ 0.37 &  42.7 $\pm$ 0.33 &  55.6 $\pm$ 0.38 &  45.0 $\pm$ 0.38 \\
& BTS         &  27.8 $\pm$ 0.28 &  32.9 $\pm$ 0.25 &  29.2 $\pm$ 0.27 &  33.6 $\pm$ 0.27 &  28.9 $\pm$ 0.25 \\
\midrule
\multirow{3}{*}{Manufacturing} & TEX     &  71.4 $\pm$ 0.57 &  91.7 $\pm$ 0.27 &  91.5 $\pm$ 0.30 &  93.0 $\pm$ 0.28 &  92.5 $\pm$ 0.28 \\
& TEX\_DTD   &  27.2 $\pm$ 0.29 &  38.8 $\pm$ 0.34 &  36.4 $\pm$ 0.32 &  34.6 $\pm$ 0.32 &  35.6 $\pm$ 0.30 \\
& TEX\_ALOT  &  69.2 $\pm$ 0.57 &  96.9 $\pm$ 0.18 &  97.1 $\pm$ 0.18 &  98.2 $\pm$ 0.12 &  97.7 $\pm$ 0.16 \\
\midrule
\multirow{3}{*}{Human actions} & SPT           &  42.3 $\pm$ 0.43 &  60.8 $\pm$ 0.42 &  55.0 $\pm$ 0.43 &  62.0 $\pm$ 0.43 &  55.3 $\pm$ 0.44 \\
& ACT\_40         &  25.8 $\pm$ 0.24 &  29.6 $\pm$ 0.26 &  26.5 $\pm$ 0.27 &  32.3 $\pm$ 0.26 &  26.7 $\pm$ 0.25 \\
& ACT\_410        &  41.3 $\pm$ 0.34 &  43.6 $\pm$ 0.31 &  30.1 $\pm$ 0.27 &  40.6 $\pm$ 0.30 &  42.1 $\pm$ 0.48 \\
\midrule
\multirow{3}{*}{OCR} & MD\_MIX        &  19.6 $\pm$ 0.18 &  53.1 $\pm$ 0.67 &  80.7 $\pm$ 0.42 &  95.7 $\pm$ 0.17 &  20.0 $\pm$ 0.18 \\
& MD\_5\_BIS      &  20.3 $\pm$ 0.19 &  56.1 $\pm$ 0.49 &  90.0 $\pm$ 0.33 &  96.6 $\pm$ 0.17 &  20.5 $\pm$ 0.20 \\
& MD\_6          &  22.2 $\pm$ 0.23 &  90.2 $\pm$ 0.40 &  91.7 $\pm$ 0.27 &  97.0 $\pm$ 0.12 &  21.3 $\pm$ 0.21 \\
\bottomrule
\end{tabular}
\end{adjustbox}
\end{table}

\begin{table}[ht]
\centering
\caption{Full performance results for 5-way 10-shot image classification on all datasets. The 95\% confidence intervals are computed at per-task level over 3 runs per dataset with 600 tasks per run (total tasks: $3\times600=1\,800$).}
\label{tab:app10shot}
 \begin{adjustbox}{width=\linewidth}
\begin{tabular}{lllllll}
\toprule
Domain & Dataset ID  &           TrainFromScratch &       Finetuning &      MatchingNet &         ProtoNet &             FO-MAML \\
\midrule 
\multirow{3}{*}{Large animals} & BRD         &  47.5 $\pm$ 0.44 &  83.2 $\pm$ 0.34 &  78.8 $\pm$ 0.36 &  85.6 $\pm$ 0.29 &  78.7 $\pm$ 0.37 \\
& DOG          &  31.4 $\pm$ 0.31 &  43.3 $\pm$ 0.35 &  46.5 $\pm$ 0.36 &  47.6 $\pm$ 0.37 &  38.8 $\pm$ 0.37 \\
& AWA           &  37.8 $\pm$ 0.34 &  46.5 $\pm$ 0.35 &  40.9 $\pm$ 0.31 &  52.0 $\pm$ 0.32 &  42.2 $\pm$ 0.37 \\
\midrule 
\multirow{3}{*}{Small animals} & PLK      &  59.6 $\pm$ 0.50 &  77.0 $\pm$ 0.47 &  73.4 $\pm$ 0.45 &  78.7 $\pm$ 0.42 &  71.7 $\pm$ 0.50 \\
& INS\_2      &  28.2 $\pm$ 0.31 &  37.0 $\pm$ 0.32 &  34.1 $\pm$ 0.30 &  36.1 $\pm$ 0.33 &  35.3 $\pm$ 0.33 \\
& INS       &  26.1 $\pm$ 0.26 &  47.0 $\pm$ 0.38 &  47.4 $\pm$ 0.41 &  52.5 $\pm$ 0.40 &  44.0 $\pm$ 0.43 \\
\midrule 
\multirow{3}{*}{Plants} & FLW       &  53.3 $\pm$ 0.48 &  78.6 $\pm$ 0.39 &  72.3 $\pm$ 0.38 &  82.8 $\pm$ 0.30 &  75.9 $\pm$ 0.43 \\
& PLT\_NET      &  32.7 $\pm$ 0.31 &  43.1 $\pm$ 0.31 &  41.3 $\pm$ 0.28 &  47.3 $\pm$ 0.28 &  38.0 $\pm$ 0.31 \\
& FNG         &  27.9 $\pm$ 0.25 &  28.9 $\pm$ 0.24 &  28.7 $\pm$ 0.23 &  31.6 $\pm$ 0.25 &  29.2 $\pm$ 0.24 \\
\midrule 
\multirow{3}{*}{Plant diseases} & PLT\_VIL   &  71.3 $\pm$ 0.40 &  87.3 $\pm$ 0.24 &  70.6 $\pm$ 0.29 &  87.6 $\pm$ 0.19 &  76.9 $\pm$ 0.30 \\
& MED\_LF &  81.3 $\pm$ 0.33 &  82.4 $\pm$ 0.27 &  74.9 $\pm$ 0.39 &  81.4 $\pm$ 0.23 &  88.6 $\pm$ 0.21 \\
& PLT\_DOC      &  23.7 $\pm$ 0.22 &  29.4 $\pm$ 0.25 &  31.6 $\pm$ 0.27 &  30.6 $\pm$ 0.24 &  27.4 $\pm$ 0.25 \\
\midrule 
\multirow{3}{*}{Microscopy} & BCT      &  53.5 $\pm$ 0.41 &  80.5 $\pm$ 0.35 &  82.4 $\pm$ 0.37 &  83.9 $\pm$ 0.32 &  82.8 $\pm$ 0.38 \\
& PNU       &  29.5 $\pm$ 0.27 &  22.5 $\pm$ 0.19 &  27.9 $\pm$ 0.22 &  29.3 $\pm$ 0.23 &  34.8 $\pm$ 0.25 \\
& PRT           &  25.8 $\pm$ 0.26 &  33.0 $\pm$ 0.25 &  35.0 $\pm$ 0.27 &  33.4 $\pm$ 0.28 &  32.2 $\pm$ 0.27 \\
\midrule 
\multirow{3}{*}{Remote sensing} & RESISC        &  42.4 $\pm$ 0.37 &  61.3 $\pm$ 0.36 &  54.4 $\pm$ 0.29 &  66.4 $\pm$ 0.35 &  68.3 $\pm$ 0.35 \\
& RSICB         &  61.1 $\pm$ 0.46 &  86.9 $\pm$ 0.25 &  85.5 $\pm$ 0.26 &  85.8 $\pm$ 0.26 &  87.6 $\pm$ 0.26 \\
& RSD          &  40.2 $\pm$ 0.35 &  61.2 $\pm$ 0.38 &  53.3 $\pm$ 0.57 &  64.7 $\pm$ 0.37 &  55.0 $\pm$ 0.29 \\
\midrule 
\multirow{3}{*}{Vehicles} & CRS          &  29.9 $\pm$ 0.30 &  69.4 $\pm$ 0.40 &  76.7 $\pm$ 0.39 &  72.4 $\pm$ 0.39 &  74.0 $\pm$ 0.43 \\
& APL     &  41.9 $\pm$ 0.39 &  50.1 $\pm$ 0.35 &  51.5 $\pm$ 0.41 &  55.5 $\pm$ 0.31 &  49.9 $\pm$ 0.39 \\
& BTS         &  29.1 $\pm$ 0.29 &  35.9 $\pm$ 0.25 &  33.1 $\pm$ 0.26 &  38.3 $\pm$ 0.26 &  31.2 $\pm$ 0.25 \\
\midrule 
\multirow{3}{*}{Manufacturing} & TEX     &  71.6 $\pm$ 0.58 &  93.8 $\pm$ 0.22 &  90.9 $\pm$ 0.29 &  93.4 $\pm$ 0.26 &  93.3 $\pm$ 0.25 \\
& TEX\_DTD   &  27.2 $\pm$ 0.28 &  42.4 $\pm$ 0.32 &  36.6 $\pm$ 0.33 &  37.5 $\pm$ 0.31 &  40.1 $\pm$ 0.30 \\
& TEX\_ALOT  &  70.7 $\pm$ 0.57 &  97.6 $\pm$ 0.15 &  98.5 $\pm$ 0.11 &  98.7 $\pm$ 0.09 &  98.3 $\pm$ 0.12 \\
\midrule 
\multirow{3}{*}{Human actions} & SPT           &  44.0 $\pm$ 0.42 &  64.8 $\pm$ 0.39 &  62.2 $\pm$ 0.40 &  65.7 $\pm$ 0.42 &  62.3 $\pm$ 0.46 \\
& ACT\_40         &  26.7 $\pm$ 0.24 &  31.6 $\pm$ 0.24 &  31.3 $\pm$ 0.24 &  35.3 $\pm$ 0.25 &  28.1 $\pm$ 0.23 \\
& ACT\_410        &  42.7 $\pm$ 0.35 &  47.3 $\pm$ 0.31 &  31.8 $\pm$ 0.31 &  47.1 $\pm$ 0.28 &  40.8 $\pm$ 0.50 \\
\midrule 
\multirow{3}{*}{OCR} & MD\_MIX        &  19.5 $\pm$ 0.16 &  55.6 $\pm$ 0.70 &  92.3 $\pm$ 0.24 &  96.5 $\pm$ 0.16 &  20.3 $\pm$ 0.17 \\
& MD\_5\_BIS      &  20.2 $\pm$ 0.18 &  60.0 $\pm$ 0.49 &  95.1 $\pm$ 0.21 &  96.9 $\pm$ 0.16 &  21.0 $\pm$ 0.21 \\
& MD\_6          &  22.1 $\pm$ 0.22 &  91.8 $\pm$ 0.36 &  94.5 $\pm$ 0.19 &  97.9 $\pm$ 0.10 &  45.5 $\pm$ 1.58 \\\bottomrule
\end{tabular}
\end{adjustbox}
\end{table}

\begin{table}[t]
\centering
\caption{Full performance results for 5-way 20-shot image classification on all data sets. The 95\% confidence intervals are computed at per-task level over 3 runs per dataset with 600 tasks per run (total tasks: $3\times600=1\,800$).}
\label{tab:app20shot}
 \begin{adjustbox}{width=\linewidth}
\begin{tabular}{lllllll}
\toprule
Domain & Dataset ID  &           TrainFromScratch &       Finetuning &      MatchingNet &         ProtoNet &             FO-MAML \\
\midrule 
\multirow{3}{*}{Large animals} & BRD         &  47.8 $\pm$ 0.47 &  85.1 $\pm$ 0.32 &  82.2 $\pm$ 0.32 &  86.9 $\pm$ 0.27 &  81.0 $\pm$ 0.36 \\
& DOG          &  31.7 $\pm$ 0.31 &  46.0 $\pm$ 0.36 &  48.1 $\pm$ 0.33 &  50.9 $\pm$ 0.36 &  41.7 $\pm$ 0.36 \\
& AWA           &  37.8 $\pm$ 0.33 &  49.4 $\pm$ 0.34 &  46.7 $\pm$ 0.32 &  53.4 $\pm$ 0.33 &  48.9 $\pm$ 0.33 \\
\midrule 
\multirow{3}{*}{Small animals} & PLK      &  59.9 $\pm$ 0.50 &  79.5 $\pm$ 0.44 &  77.3 $\pm$ 0.41 &  80.2 $\pm$ 0.40 &  75.8 $\pm$ 0.48 \\
& INS\_2      &  28.1 $\pm$ 0.31 &  39.3 $\pm$ 0.33 &  37.3 $\pm$ 0.32 &  41.3 $\pm$ 0.31 &  38.9 $\pm$ 0.33 \\
& INS       &  25.8 $\pm$ 0.26 &  49.8 $\pm$ 0.39 &  51.2 $\pm$ 0.37 &  54.4 $\pm$ 0.36 &  47.2 $\pm$ 0.43 \\
\midrule 
\multirow{3}{*}{Plants} & FLW       &  54.2 $\pm$ 0.50 &  80.5 $\pm$ 0.36 &  76.4 $\pm$ 0.33 &  85.7 $\pm$ 0.26 &  79.2 $\pm$ 0.35 \\
& PLT\_NET      &  32.6 $\pm$ 0.32 &  44.6 $\pm$ 0.37 &  47.7 $\pm$ 0.27 &  47.0 $\pm$ 0.32 &  41.6 $\pm$ 0.30 \\
& FNG         &  27.8 $\pm$ 0.25 &  30.1 $\pm$ 0.24 &  32.2 $\pm$ 0.27 &  34.2 $\pm$ 0.24 &  31.3 $\pm$ 0.23 \\
\midrule 
\multirow{3}{*}{Plant diseases} & PLT\_VIL   &  72.0 $\pm$ 0.42 &  92.9 $\pm$ 0.17 &  80.7 $\pm$ 0.24 &  90.2 $\pm$ 0.17 &  85.3 $\pm$ 0.21 \\
& MED\_LF &  82.6 $\pm$ 0.32 &  83.7 $\pm$ 0.30 &  81.5 $\pm$ 0.23 &  84.4 $\pm$ 0.25 &  93.1 $\pm$ 0.16 \\
& PLT\_DOC      &  23.5 $\pm$ 0.21 &  31.9 $\pm$ 0.25 &  31.8 $\pm$ 0.21 &  30.7 $\pm$ 0.23 &  32.2 $\pm$ 0.25 \\
\midrule 
\multirow{3}{*}{Microscopy} & BCT      &  53.6 $\pm$ 0.42 &  80.8 $\pm$ 0.35 &  85.7 $\pm$ 0.28 &  87.0 $\pm$ 0.30 &  82.4 $\pm$ 0.40 \\
& PNU       &  29.7 $\pm$ 0.26 &  23.5 $\pm$ 0.20 &  29.8 $\pm$ 0.21 &  27.2 $\pm$ 0.28 &  38.0 $\pm$ 0.26 \\
& PRT           &  26.1 $\pm$ 0.25 &  34.2 $\pm$ 0.26 &  36.3 $\pm$ 0.26 &  36.8 $\pm$ 0.27 &  32.7 $\pm$ 0.25 \\
\midrule 
\multirow{3}{*}{Remote sensing} & RESISC        &  42.8 $\pm$ 0.36 &  65.7 $\pm$ 0.34 &  64.7 $\pm$ 0.31 &  67.2 $\pm$ 0.31 &  65.0 $\pm$ 0.39 \\
& RSICB         &  61.5 $\pm$ 0.49 &  89.2 $\pm$ 0.21 &  88.9 $\pm$ 0.22 &  90.9 $\pm$ 0.19 &  91.2 $\pm$ 0.22 \\
& RSD          &  40.1 $\pm$ 0.35 &  65.9 $\pm$ 0.37 &  63.3 $\pm$ 0.36 &  67.6 $\pm$ 0.35 &  58.8 $\pm$ 0.28 \\
\midrule 
\multirow{3}{*}{Vehicles} & CRS          &  30.1 $\pm$ 0.32 &  73.1 $\pm$ 0.39 &  76.4 $\pm$ 0.37 &  76.1 $\pm$ 0.35 &  75.8 $\pm$ 0.44 \\
& APL     &  41.5 $\pm$ 0.38 &  52.7 $\pm$ 0.32 &  57.3 $\pm$ 0.27 &  57.9 $\pm$ 0.33 &  52.1 $\pm$ 0.32 \\
& BTS         &  29.1 $\pm$ 0.30 &  36.9 $\pm$ 0.26 &  36.3 $\pm$ 0.24 &  40.1 $\pm$ 0.27 &  32.5 $\pm$ 0.24 \\
\midrule 
\multirow{3}{*}{Manufacturing} & TEX     &  71.6 $\pm$ 0.59 &  95.4 $\pm$ 0.19 &  94.7 $\pm$ 0.21 &  95.1 $\pm$ 0.21 &  94.7 $\pm$ 0.22 \\
& TEX\_DTD   &  27.1 $\pm$ 0.28 &  46.2 $\pm$ 0.35 &  39.4 $\pm$ 0.31 &  41.7 $\pm$ 0.32 &  44.1 $\pm$ 0.29 \\
& TEX\_ALOT  &  70.4 $\pm$ 0.58 &  98.3 $\pm$ 0.11 &  98.9 $\pm$ 0.10 &  98.8 $\pm$ 0.08 &  98.6 $\pm$ 0.11 \\
\midrule 
\multirow{3}{*}{Human actions} & SPT           &  44.4 $\pm$ 0.43 &  68.4 $\pm$ 0.39 &  66.1 $\pm$ 0.37 &  69.8 $\pm$ 0.38 &  63.5 $\pm$ 0.43 \\
& ACT\_40         &  26.8 $\pm$ 0.23 &  32.4 $\pm$ 0.23 &  33.6 $\pm$ 0.22 &  37.1 $\pm$ 0.24 &  29.5 $\pm$ 0.22 \\
& ACT\_410        &  42.5 $\pm$ 0.36 &  47.5 $\pm$ 0.30 &  40.7 $\pm$ 0.23 &  50.7 $\pm$ 0.29 &  55.9 $\pm$ 0.31 \\
\midrule 
\multirow{3}{*}{OCR} & MD\_MIX        &  19.5 $\pm$ 0.15 &  52.3 $\pm$ 0.77 &  94.6 $\pm$ 0.19 &  96.7 $\pm$ 0.15 &  20.4 $\pm$ 0.14 \\
& MD\_5\_BIS      &  20.2 $\pm$ 0.17 &  71.8 $\pm$ 0.61 &  96.3 $\pm$ 0.17 &  96.7 $\pm$ 0.15 &  21.1 $\pm$ 0.21 \\
& MD\_6          &  21.4 $\pm$ 0.19 &  93.2 $\pm$ 0.34 &  96.5 $\pm$ 0.14 &  98.0 $\pm$ 0.09 &  45.9 $\pm$ 1.58 \\
\bottomrule
\end{tabular}
\end{adjustbox}
\end{table}
 
\begin{table}[b]
 \centering
 
 \caption{The average rankings over all data sets and running times for all techniques on 5-way [1, 5, 10, 20]-shot image classification. The 95\% confidence intervals are computed at dataset level over 30 datasets.}
 \label{tab:apprankings}
  \begin{adjustbox}{width=\linewidth}
 \begin{tabular}{l@{\extracolsep{2pt}}ll ll ll ll}
\toprule
& \multicolumn{2}{c}{1-shot} & \multicolumn{2}{c}{5-shot} & \multicolumn{2}{c}{10-shot} & \multicolumn{2}{c}{20-shot} \\ 
 \cline{2-3}  \cline{4-5} \cline{6-7}  \cline{8-9} \\
& AR & Time & AR & Time & AR & Time & AR & Time \\
\midrule
TrainFromScratch         &  4.6 $\pm$ 0.35 & 1h29min &  4.7 $\pm$ 0.28 & 1h29min & 4.8 $\pm$ 0.25 &  1h29min & 4.9 $\pm$ 0.16 & 1h29min \\
Finetuning  &  2.4 $\pm$ 0.39 & 3h31min &  2.5 $\pm$ 0.36 & 3h37min & 2.5 $\pm$ 0.36 & 3h41min & 2.7 $\pm$ 0.36 & 4h18min \\
MatchingNet &  3.2 $\pm$ 0.36 & 2h50min & 3.3 $\pm$ 0.46 & 3h19min & 3.1 $\pm$ 0.45 & 3h40min &  2.8 $\pm$ 0.43 & 6h52min \\
ProtoNet    &  1.8 $\pm$ 0.43 & 2h52min &  1.7 $\pm$ 0.41 & 3h15min & 1.5 $\pm$ 0.29 & 3h40min &  1.5 $\pm$ 0.32 & 6h58min \\
FO-MAML        &  2.9 $\pm$ 0.48 & 6h52min & 2.9 $\pm$ 0.38 & 9h12min &  3.1 $\pm$ 0.40 & 11h24min & 3.1 $\pm$ 0.41 & 17h8min \\
\bottomrule
\end{tabular}
 \end{adjustbox}
 \end{table}
 
\clearpage
\newpage
\section{Cross-Domain Few-shot learning: additional results}
\label{appendix:cd-fs-additional}

\begin{figure}[t]
 \centering
 \begin{subfigure}{0.48\textwidth}
     \centering
     \includegraphics[width=\textwidth]{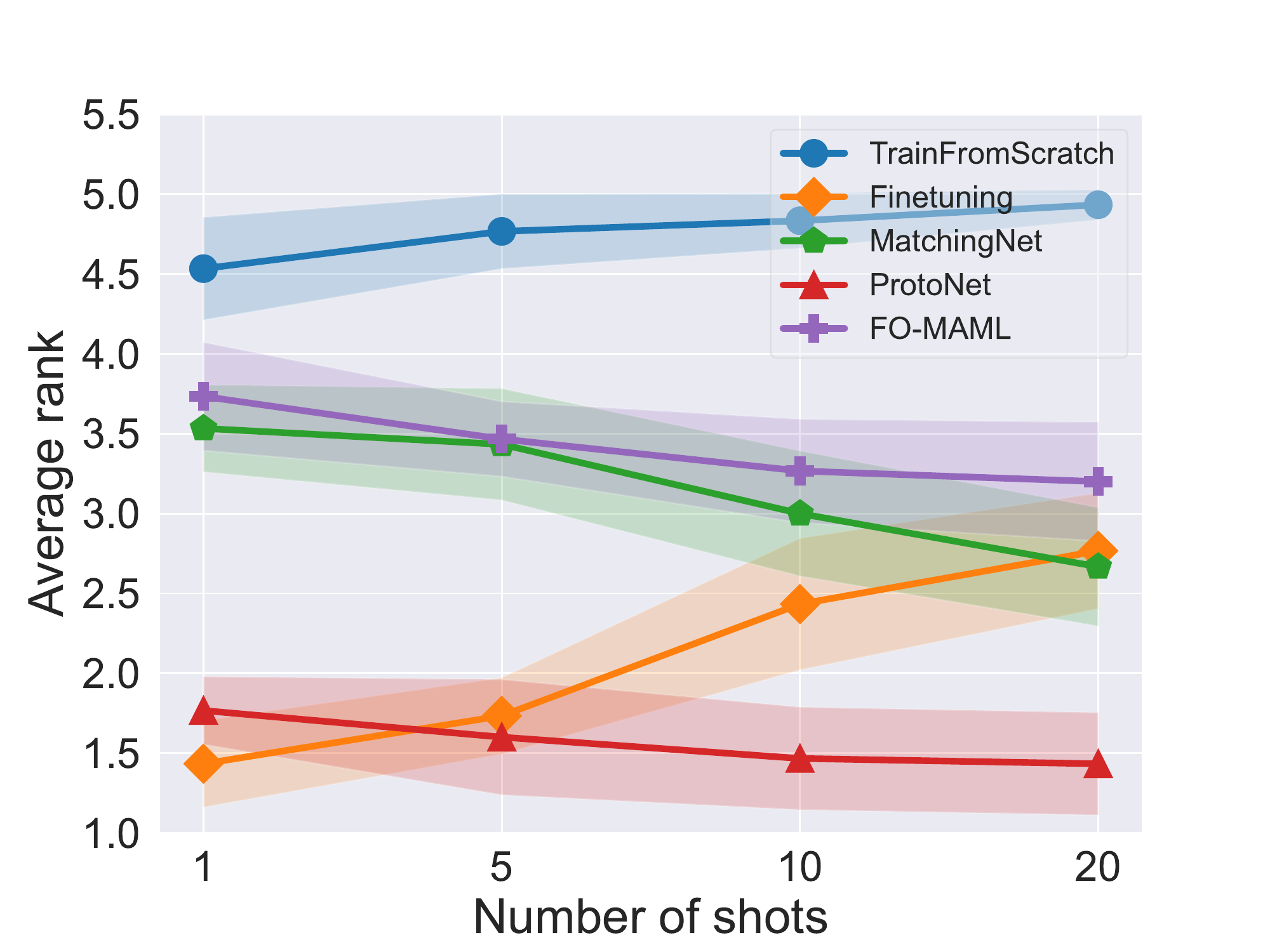}
     \caption{Average rankings}
 \end{subfigure}
 \hfill
 \begin{subfigure}{0.48\textwidth}
     \centering
     \includegraphics[width=\textwidth]{fig/cross-fsl/cfsl-avgaccs.pdf}
     \caption{Average accuracy}
 \end{subfigure}
 \caption{\label{fig:cdfsl-plots} {\bf Cross-domain few-shot learning.} Averages over 30 released Meta-Album datasets of: (a) algorithm rank (smaller is better). The 95\% confidence intervals are computed at dataset level over 30 datasets. (b) classification accuracy. The 95\% confidence intervals are computed at per-task level over 3 runs per dataset with 600 tasks per run (total tasks: $3 \times 600 \times 30 = 54\,000$).}
\end{figure}

Here, we show additional results for the cross-domain few-shot learning experiments from Section 3.2. Table~\ref{tab:cdfsl1shot}, Table~\ref{tab:cdfsl5shot}, Table~\ref{tab:cdfsl10shot}, and Table~\ref{tab:cdfsl20shot} display the average accuracy per technique and dataset in the 5-way [1, 5, 10, 20]-shot settings. Moreover, Figure~\ref{fig:cdfsl-plots} displays the average rank per method and average accuracy for the same settings. 

\begin{figure}[t]
 \centering
 \includegraphics[width=0.6\textwidth]{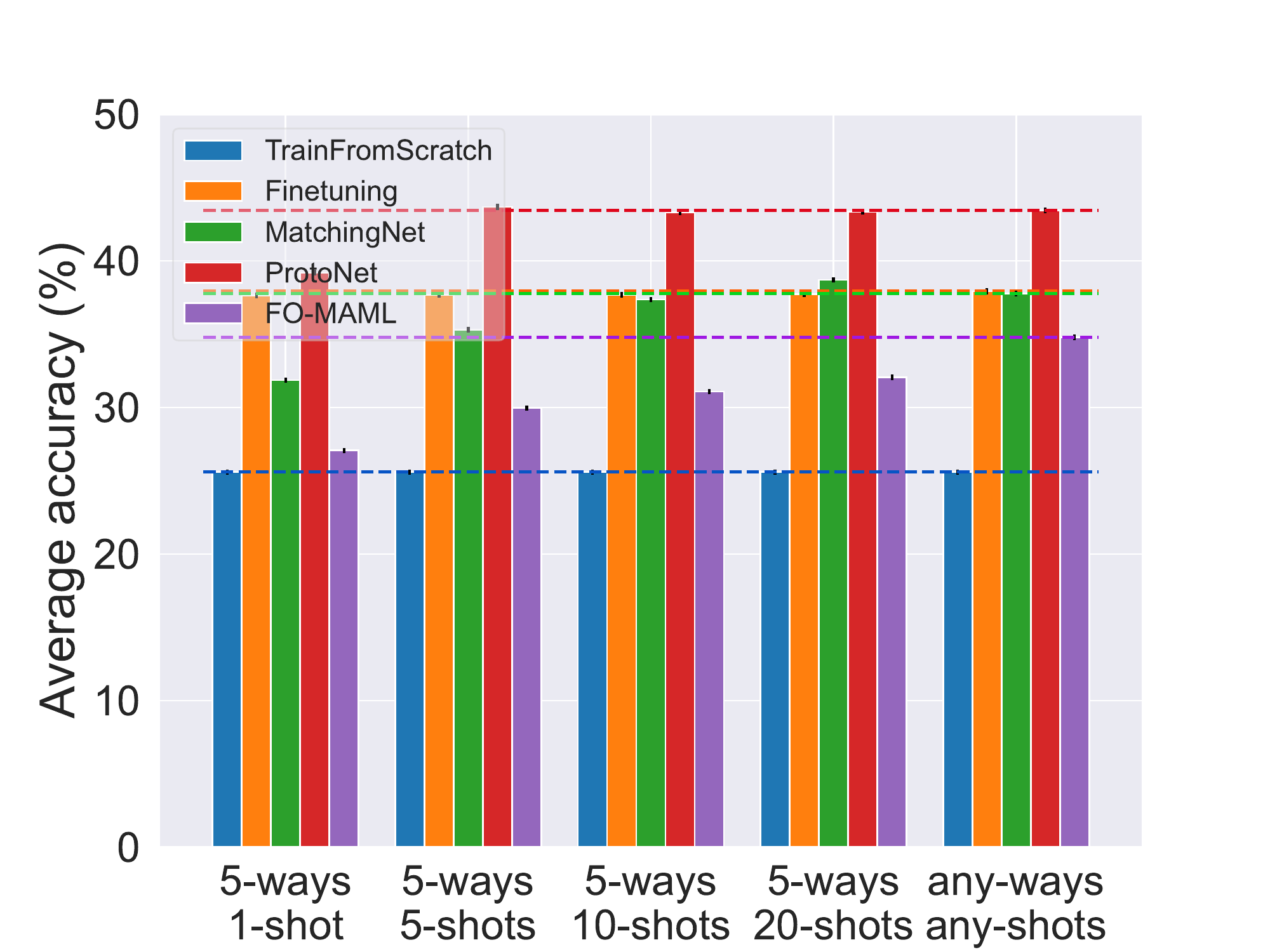}
 \caption{\label{fig:cdfsl-trained-fixed-tested-any} {\bf Comparison of ``cross-domain'' few-shot learning trained on different settings but tested on the any-way any-shot setting.} We plot few-shot learning meta-test mean task accuracy, averaged over test tasks drawn from the 30 released Meta-Album datasets. The 95\% confidence intervals are computed at per-task level over 3 runs per dataset with 600 tasks per run (total tasks: $3 \times 600 \times 30 = 54\,000$).}
\end{figure}

Table~\ref{tab:cdfslAnyshot} displays the average accuracy per technique and dataset in the any-way any-shot setting. Table~\ref{tab:cdfslWays} and
Table~\ref{tab:cdfslShots} display the results of the any-way any-shot setting but grouped by $N$-way and $k$-shot, respectively. Figure~\ref{fig:cdfsl-trained-fixed-tested-any} shows a performance comparison of the different techniques when trained on each of the evaluated settings and tested on the novel any-way any-shot setting. From these results, we can observe that most of the techniques are not benefited from training with a variable number of shots; only MAML shows a performance improvement by this approach.

Table~\ref{tab:cdfslTime} displays the running times per technique in the different cross-domain settings. This time is separated in meta-training, meta-validation and meta-testing. Note that the meta-training and meta-validation time of TrainFromScratch is 0 because this baseline learns every task starting from a random initialization at meta-test time.

\begin{table}[!ht]
    \centering
    \caption{Full performance results for cross-domain 5-way 1-shot image classification on all datasets. The 95\% confidence intervals are computed at per-task level over 3 runs per dataset with 600 tasks per run (total tasks: $3\times600=1\,800$).}
    \label{tab:cdfsl1shot}
     \begin{adjustbox}{width=\linewidth}
    \begin{tabular}{lllllll}
    \toprule
    Domain & Dataset ID  &           TrainFromScratch &       Finetuning &      MatchingNet &         ProtoNet &             FO-MAML \\
    \midrule 
    \multirow{3}{*}{Large animals} &  BRD & $31.45 \pm 0.38$ & $42.93 \pm 0.43$ & $37.53 \pm 0.48$ & $40.07 \pm 0.44$ & $31.59 \pm 0.35$ \\
    &      DOG & $23.79 \pm 0.27$ & $26.38 \pm 0.28$ & $25.17 \pm 0.29$ & $26.15 \pm 0.29$ & $24.41 \pm 0.29$ \\
    &      AWA & $25.75 \pm 0.31$ & $36.69 \pm 0.44$ & $32.16 \pm 0.43$ & $36.83 \pm 0.46$ & $32.12 \pm 0.44$ \\
    \midrule 
    \multirow{3}{*}{Small animals}     &      PLK & $42.15 \pm 0.49$ & $50.72 \pm 0.52$ & $34.28 \pm 0.47$ & $44.78 \pm 0.52$ & $42.08 \pm 0.48$ \\
    &    INS\_2 & $23.81 \pm 0.28$ & $29.10 \pm 0.32$ & $27.03 \pm 0.34$ & $28.22 \pm 0.33$ & $25.48 \pm 0.31$ \\
    & INS & $23.00 \pm 0.25$ & $31.01 \pm 0.34$ & $27.07 \pm 0.35$ & $30.57 \pm 0.34$ & $27.10 \pm 0.34$ \\
    \midrule 
    \multirow{3}{*}{Plants} & FLW & $38.47 \pm 0.48$ & $49.48 \pm 0.51$ & $50.23 \pm 0.59$ & $53.57 \pm 0.57$ & $42.91 \pm 0.50$ \\
    &  PLT\_NET & $24.57 \pm 0.28$ & $32.80 \pm 0.36$ & $29.88 \pm 0.37$ & $30.78 \pm 0.35$ & $27.87 \pm 0.34$ \\
    &      FNG & $22.04 \pm 0.24$ & $26.92 \pm 0.29$ & $25.93 \pm 0.33$ & $27.11 \pm 0.31$ & $26.73 \pm 0.34$ \\
    \midrule 
    \multirow{3}{*}{Plant diseases} & PLT\_VIL & $37.49 \pm 0.45$ & $52.47 \pm 0.51$ & $42.18 \pm 0.48$ & $50.03 \pm 0.53$ & $44.61 \pm 0.45$ \\
    &   MED\_LF & $54.54 \pm 0.56$ & $68.19 \pm 0.54$ & $56.02 \pm 0.59$ & $64.76 \pm 0.54$ & $50.27 \pm 0.61$ \\
    &  PLT\_DOC & $23.61 \pm 0.28$ & $30.54 \pm 0.32$ & $26.92 \pm 0.33$ & $33.77 \pm 0.38$ & $27.61 \pm 0.34$ \\
    \midrule 
    \multirow{3}{*}{Microscopy} & BCT & $32.26 \pm 0.41$ & $40.16 \pm 0.44$ & $34.64 \pm 0.46$ & $37.24 \pm 0.43$ & $35.40 \pm 0.42$ \\
    &      PNU & $22.61 \pm 0.26$ & $23.71 \pm 0.24$ & $23.01 \pm 0.26$ & $23.83 \pm 0.25$ & $22.58 \pm 0.25$ \\
    &      PRT & $21.78 \pm 0.24$ & $23.56 \pm 0.24$ & $24.36 \pm 0.27$ & $25.08 \pm 0.28$ & $25.10 \pm 0.29$ \\
    \midrule 
    \multirow{3}{*}{Remote sensing} & RESISC & $33.61 \pm 0.44$ & $42.84 \pm 0.45$ & $41.76 \pm 0.52$ & $44.37 \pm 0.48$ & $38.83 \pm 0.47$ \\
    &    RSICB & $45.48 \pm 0.55$ & $66.90 \pm 0.51$ & $59.01 \pm 0.62$ & $62.77 \pm 0.57$ & $60.44 \pm 0.65$ \\
    &      RSD & $31.74 \pm 0.40$ & $46.82 \pm 0.49$ & $37.97 \pm 0.47$ & $48.21 \pm 0.50$ & $38.64 \pm 0.47$ \\
    \midrule 
    \multirow{3}{*}{Vehicles} &      CRS & $24.03 \pm 0.26$ & $27.10 \pm 0.31$ & $25.75 \pm 0.35$ & $26.70 \pm 0.33$ & $23.73 \pm 0.26$ \\
    & APL & $30.76 \pm 0.36$ & $34.67 \pm 0.37$ & $28.29 \pm 0.34$ & $32.81 \pm 0.36$ & $26.90 \pm 0.33$ \\
    &      BTS & $22.77 \pm 0.26$ & $26.85 \pm 0.29$ & $25.30 \pm 0.32$ & $25.80 \pm 0.28$ & $23.30 \pm 0.26$ \\
    \midrule 
    \multirow{3}{*}{Manufacturing} & TEX & $53.41 \pm 0.65$ & $72.43 \pm 0.58$ & $71.29 \pm 0.63$ & $70.04 \pm 0.65$ & $62.54 \pm 0.64$ \\
    &  TEX\_DTD & $24.37 \pm 0.30$ & $33.19 \pm 0.37$ & $26.70 \pm 0.35$ & $30.25 \pm 0.38$ & $26.95 \pm 0.35$ \\
    & TEX\_ALOT & $55.62 \pm 0.59$ & $83.82 \pm 0.44$ & $74.34 \pm 0.60$ & $84.43 \pm 0.44$ & $77.78 \pm 0.62$ \\
    \midrule 
    \multirow{3}{*}{Human actions}     &      SPT & $32.08 \pm 0.40$ & $38.68 \pm 0.39$ & $32.18 \pm 0.39$ & $36.02 \pm 0.41$ & $33.96 \pm 0.39$ \\
    &   ACT\_40 & $23.23 \pm 0.26$ & $28.83 \pm 0.30$ & $26.31 \pm 0.31$ & $27.19 \pm 0.30$ & $25.69 \pm 0.32$ \\
    & ACT\_410 & $28.08 \pm 0.35$ & $34.61 \pm 0.37$ & $31.00 \pm 0.40$ & $35.52 \pm 0.40$ & $32.05 \pm 0.39$ \\
    \midrule 
    \multirow{3}{*}{OCR} & MD\_MIX & $19.93 \pm 0.19$ & $35.68 \pm 0.40$ & $19.63 \pm 0.20$ & $19.79 \pm 0.19$ & $19.68 \pm 0.20$ \\
    & MD\_5\_BIS & $20.46 \pm 0.21$ & $26.66 \pm 0.28$ & $20.21 \pm 0.21$ & $21.35 \pm 0.22$ & $20.06 \pm 0.21$ \\
    &     MD\_6 & $21.09 \pm 0.22$ & $50.13 \pm 0.48$ & $20.94 \pm 0.23$ & $28.75 \pm 0.47$ & $20.52 \pm 0.22$ \\
    \bottomrule
\end{tabular}
\end{adjustbox}
\end{table}

\begin{table}[!ht]
    \centering
    \caption{Full performance results for cross-domain 5-way 5-shot image classification on all datasets. The 95\% confidence intervals are computed at per-task level over 3 runs per dataset with 600 tasks per run (total tasks: $3\times600=1\,800$).}
    \label{tab:cdfsl5shot}
     \begin{adjustbox}{width=\linewidth}
    \begin{tabular}{lllllll}
    \toprule
    Domain & Dataset ID  &           TrainFromScratch &       Finetuning &      MatchingNet &         ProtoNet &             FO-MAML \\
    \midrule 
    \multirow{3}{*}{Large animals} & BRD & $45.83 \pm 0.43$ & $57.30 \pm 0.44$ & $50.10 \pm 0.44$ & $59.77 \pm 0.41$ & $46.19 \pm 0.42$ \\
    &      DOG & $28.70 \pm 0.29$ & $32.15 \pm 0.30$ & $32.16 \pm 0.30$ & $32.04 \pm 0.30$ & $31.55 \pm 0.32$ \\
    &      AWA & $32.75 \pm 0.37$ & $47.31 \pm 0.48$ & $44.43 \pm 0.46$ & $49.63 \pm 0.48$ & $44.06 \pm 0.47$ \\
    \midrule 
    \multirow{3}{*}{Small animals}     &      PLK & $55.01 \pm 0.52$ & $62.53 \pm 0.49$ & $50.65 \pm 0.43$ & $63.35 \pm 0.49$ & $53.75 \pm 0.54$ \\
    &    INS\_2 & $27.91 \pm 0.30$ & $37.71 \pm 0.33$ & $34.64 \pm 0.34$ & $35.32 \pm 0.33$ & $34.73 \pm 0.35$ \\
    & INS & $27.15 \pm 0.28$ & $40.09 \pm 0.38$ & $36.29 \pm 0.35$ & $39.90 \pm 0.38$ & $35.62 \pm 0.36$ \\
    \midrule 
    \multirow{3}{*}{Plants} & FLW & $53.29 \pm 0.51$ & $64.07 \pm 0.53$ & $63.37 \pm 0.49$ & $70.72 \pm 0.47$ & $60.74 \pm 0.49$ \\
    &  PLT\_NET & $29.74 \pm 0.31$ & $42.85 \pm 0.37$ & $40.03 \pm 0.37$ & $38.92 \pm 0.34$ & $40.89 \pm 0.36$ \\
    &      FNG & $25.41 \pm 0.26$ & $32.64 \pm 0.32$ & $32.28 \pm 0.35$ & $33.56 \pm 0.34$ & $32.43 \pm 0.36$ \\
    \midrule 
    \multirow{3}{*}{Plant diseases} & PLT\_VIL & $55.92 \pm 0.49$ & $67.84 \pm 0.49$ & $58.64 \pm 0.43$ & $69.99 \pm 0.47$ & $61.85 \pm 0.47$ \\
    &   MED\_LF & $72.79 \pm 0.47$ & $81.77 \pm 0.42$ & $70.24 \pm 0.46$ & $81.89 \pm 0.39$ & $76.26 \pm 0.45$ \\
    &  PLT\_DOC & $27.52 \pm 0.28$ & $39.41 \pm 0.35$ & $35.70 \pm 0.34$ & $46.88 \pm 0.38$ & $36.01 \pm 0.34$ \\
    \midrule 
    \multirow{3}{*}{Microscopy} & BCT & $43.48 \pm 0.47$ & $49.64 \pm 0.45$ & $49.37 \pm 0.48$ & $54.46 \pm 0.49$ & $46.42 \pm 0.44$ \\
    &      PNU & $25.12 \pm 0.26$ & $25.75 \pm 0.23$ & $25.55 \pm 0.25$ & $25.34 \pm 0.24$ & $25.42 \pm 0.26$ \\
    &      PRT & $25.71 \pm 0.26$ & $26.99 \pm 0.26$ & $29.81 \pm 0.28$ & $29.89 \pm 0.28$ & $29.23 \pm 0.29$ \\
    \midrule 
    \multirow{3}{*}{Remote sensing} & RESISC & $41.74 \pm 0.43$ & $56.74 \pm 0.44$ & $54.84 \pm 0.44$ & $57.62 \pm 0.45$ & $52.88 \pm 0.47$ \\
    &    RSICB & $57.07 \pm 0.51$ & $82.30 \pm 0.39$ & $75.38 \pm 0.43$ & $77.57 \pm 0.42$ & $77.56 \pm 0.46$ \\
    &      RSD & $39.09 \pm 0.40$ & $62.31 \pm 0.46$ & $53.84 \pm 0.43$ & $67.28 \pm 0.46$ & $56.12 \pm 0.46$ \\
    \midrule 
    \multirow{3}{*}{Vehicles} &      CRS & $29.54 \pm 0.32$ & $32.82 \pm 0.33$ & $31.28 \pm 0.35$ & $34.66 \pm 0.37$ & $29.70 \pm 0.33$ \\
    & APL & $41.97 \pm 0.47$ & $48.65 \pm 0.42$ & $36.68 \pm 0.37$ & $49.66 \pm 0.40$ & $39.36 \pm 0.39$ \\
    &      BTS & $26.62 \pm 0.28$ & $33.48 \pm 0.31$ & $30.02 \pm 0.28$ & $35.56 \pm 0.32$ & $30.62 \pm 0.28$ \\
    \midrule 
    \multirow{3}{*}{Manufacturing} & TEX & $69.12 \pm 0.60$ & $81.83 \pm 0.50$ & $81.44 \pm 0.46$ & $85.84 \pm 0.40$ & $78.84 \pm 0.52$ \\
    &  TEX\_DTD & $27.09 \pm 0.29$ & $45.20 \pm 0.39$ & $36.66 \pm 0.36$ & $43.37 \pm 0.39$ & $36.16 \pm 0.36$ \\
    & TEX\_ALOT & $68.29 \pm 0.55$ & $93.65 \pm 0.26$ & $89.34 \pm 0.34$ & $95.36 \pm 0.20$ & $89.77 \pm 0.39$ \\
    \midrule 
    \multirow{3}{*}{Human actions}     &      SPT & $43.78 \pm 0.43$ & $51.71 \pm 0.43$ & $45.05 \pm 0.41$ & $50.85 \pm 0.41$ & $45.40 \pm 0.43$ \\
    &   ACT\_40 & $28.33 \pm 0.29$ & $35.41 \pm 0.33$ & $34.48 \pm 0.33$ & $31.62 \pm 0.31$ & $33.23 \pm 0.33$ \\
    & ACT\_410 & $38.91 \pm 0.40$ & $47.89 \pm 0.40$ & $44.65 \pm 0.41$ & $51.08 \pm 0.40$ & $45.87 \pm 0.41$ \\
    \midrule 
    \multirow{3}{*}{OCR} & MD\_MIX & $19.78 \pm 0.18$ & $48.39 \pm 0.41$ & $19.65 \pm 0.20$ & $20.92 \pm 0.21$ & $19.25 \pm 0.20$ \\
    & MD\_5\_BIS & $20.21 \pm 0.19$ & $33.26 \pm 0.33$ & $21.40 \pm 0.21$ & $51.36 \pm 0.48$ & $22.12 \pm 0.23$ \\
    &     MD\_6 & $22.28 \pm 0.22$ & $65.88 \pm 0.45$ & $23.84 \pm 0.25$ & $52.77 \pm 0.57$ & $22.96 \pm 0.23$ \\
    \bottomrule
\end{tabular}
\end{adjustbox}
\end{table}

\begin{table}[!ht]
    \centering
    \caption{Full performance results for cross-domain 5-way 10-shot image classification on all datasets. The 95\% confidence intervals are computed at per-task level over 3 runs per dataset with 600 tasks per run (total tasks: $3\times600=1\,800$).}
    \label{tab:cdfsl10shot}
     \begin{adjustbox}{width=\linewidth}
    \begin{tabular}{lllllll}
    \toprule
    Domain & Dataset ID  &           TrainFromScratch &       Finetuning &      MatchingNet &         ProtoNet &             FO-MAML \\
    \midrule 
    \multirow{3}{*}{Large animals} & BRD & $48.48 \pm 0.46$ & $61.79 \pm 0.43$ & $57.60 \pm 0.40$ & $62.38 \pm 0.40$ & $51.58 \pm 0.40$ \\
    &      DOG & $29.52 \pm 0.29$ & $34.00 \pm 0.30$ & $35.91 \pm 0.32$ & $36.10 \pm 0.31$ & $34.48 \pm 0.32$ \\
    &      AWA & $33.74 \pm 0.38$ & $50.16 \pm 0.49$ & $50.50 \pm 0.45$ & $54.60 \pm 0.48$ & $47.58 \pm 0.48$ \\
    \midrule 
    \multirow{3}{*}{Small animals}     &      PLK & $56.40 \pm 0.52$ & $64.48 \pm 0.49$ & $57.23 \pm 0.43$ & $70.63 \pm 0.45$ & $63.18 \pm 0.44$ \\
    &    INS\_2 & $28.03 \pm 0.30$ & $40.23 \pm 0.34$ & $38.74 \pm 0.35$ & $40.13 \pm 0.35$ & $38.89 \pm 0.36$ \\
    & INS & $27.76 \pm 0.30$ & $41.61 \pm 0.39$ & $41.40 \pm 0.35$ & $45.36 \pm 0.38$ & $39.95 \pm 0.38$ \\
    \midrule 
    \multirow{3}{*}{Plants} & FLW & $55.70 \pm 0.52$ & $67.72 \pm 0.50$ & $68.24 \pm 0.45$ & $71.31 \pm 0.46$ & $66.97 \pm 0.48$ \\
    &  PLT\_NET & $31.02 \pm 0.32$ & $45.71 \pm 0.36$ & $45.40 \pm 0.37$ & $43.16 \pm 0.33$ & $45.49 \pm 0.35$ \\
    &      FNG & $26.50 \pm 0.27$ & $34.09 \pm 0.33$ & $35.66 \pm 0.35$ & $37.73 \pm 0.33$ & $34.33 \pm 0.33$ \\
    \midrule 
    \multirow{3}{*}{Plant diseases} & PLT\_VIL & $59.48 \pm 0.51$ & $70.68 \pm 0.48$ & $66.44 \pm 0.44$ & $74.39 \pm 0.43$ & $70.83 \pm 0.43$ \\
    &   MED\_LF & $76.44 \pm 0.47$ & $83.83 \pm 0.41$ & $71.73 \pm 0.40$ & $86.76 \pm 0.34$ & $80.80 \pm 0.40$ \\
    &  PLT\_DOC & $27.95 \pm 0.29$ & $41.36 \pm 0.35$ & $43.32 \pm 0.34$ & $50.87 \pm 0.38$ & $38.73 \pm 0.34$ \\
    \midrule 
    \multirow{3}{*}{Microscopy} & BCT & $46.23 \pm 0.48$ & $51.23 \pm 0.46$ & $57.60 \pm 0.43$ & $54.61 \pm 0.45$ & $51.40 \pm 0.46$ \\
    &      PNU & $26.17 \pm 0.25$ & $26.03 \pm 0.23$ & $28.53 \pm 0.26$ & $27.12 \pm 0.25$ & $28.64 \pm 0.26$ \\
    &      PRT & $26.65 \pm 0.28$ & $28.15 \pm 0.26$ & $31.40 \pm 0.30$ & $32.87 \pm 0.27$ & $32.55 \pm 0.29$ \\
    \midrule 
    \multirow{3}{*}{Remote sensing} & RESISC & $42.49 \pm 0.42$ & $60.34 \pm 0.44$ & $60.00 \pm 0.41$ & $61.27 \pm 0.43$ & $57.14 \pm 0.46$ \\
    &    RSICB & $57.65 \pm 0.54$ & $84.49 \pm 0.36$ & $79.37 \pm 0.39$ & $83.72 \pm 0.36$ & $81.53 \pm 0.40$ \\
    &      RSD & $39.90 \pm 0.42$ & $66.35 \pm 0.46$ & $64.29 \pm 0.41$ & $72.30 \pm 0.42$ & $61.07 \pm 0.47$ \\
    \midrule 
    \multirow{3}{*}{Vehicles} &      CRS & $31.39 \pm 0.35$ & $34.76 \pm 0.36$ & $35.05 \pm 0.36$ & $37.47 \pm 0.37$ & $32.20 \pm 0.31$ \\
    & APL & $43.64 \pm 0.51$ & $52.67 \pm 0.43$ & $41.73 \pm 0.42$ & $56.30 \pm 0.39$ & $42.43 \pm 0.40$ \\
    &      BTS & $27.44 \pm 0.28$ & $35.51 \pm 0.31$ & $31.82 \pm 0.28$ & $40.35 \pm 0.33$ & $33.93 \pm 0.29$ \\
    \midrule 
    \multirow{3}{*}{Manufacturing} & TEX & $71.33 \pm 0.60$ & $82.25 \pm 0.50$ & $85.79 \pm 0.38$ & $86.20 \pm 0.41$ & $84.93 \pm 0.44$ \\
    &  TEX\_DTD & $27.09 \pm 0.28$ & $49.24 \pm 0.37$ & $40.48 \pm 0.35$ & $49.45 \pm 0.38$ & $39.60 \pm 0.36$ \\
    & TEX\_ALOT & $69.09 \pm 0.56$ & $94.96 \pm 0.23$ & $92.44 \pm 0.26$ & $97.04 \pm 0.15$ & $92.97 \pm 0.32$ \\
    \midrule 
    \multirow{3}{*}{Human actions}     &      SPT & $45.98 \pm 0.44$ & $55.94 \pm 0.42$ & $51.93 \pm 0.39$ & $54.67 \pm 0.42$ & $52.74 \pm 0.42$ \\
    &   ACT\_40 & $29.32 \pm 0.28$ & $37.45 \pm 0.34$ & $37.96 \pm 0.34$ & $35.27 \pm 0.32$ & $37.15 \pm 0.35$ \\
    & ACT\_410 & $40.49 \pm 0.40$ & $51.72 \pm 0.41$ & $50.43 \pm 0.39$ & $58.72 \pm 0.39$ & $52.12 \pm 0.41$ \\
    \midrule 
    \multirow{3}{*}{OCR} & MD\_MIX & $19.51 \pm 0.17$ & $50.45 \pm 0.41$ & $19.68 \pm 0.21$ & $35.54 \pm 0.58$ & $18.78 \pm 0.20$ \\
    & MD\_5\_BIS & $20.37 \pm 0.18$ & $38.49 \pm 0.34$ & $22.84 \pm 0.23$ & $53.14 \pm 0.39$ & $22.75 \pm 0.23$ \\
    &     MD\_6 & $22.24 \pm 0.22$ & $69.41 \pm 0.42$ & $32.29 \pm 0.31$ & $51.61 \pm 0.46$ & $24.98 \pm 0.25$ \\
    \bottomrule
\end{tabular}
\end{adjustbox}
\end{table}

\begin{table}[!ht]
    \centering
    \caption{Full performance results for cross-domain 5-way 20-shot image classification on all datasets. The 95\% confidence intervals are computed at per-task level over 3 runs per dataset with 600 tasks per run (total tasks: $3\times600=1\,800$).}
    \label{tab:cdfsl20shot}
     \begin{adjustbox}{width=\linewidth}
    \begin{tabular}{lllllll}
    \toprule
    Domain & Dataset ID  &           TrainFromScratch &       Finetuning &      MatchingNet &         ProtoNet &             FO-MAML \\
    \midrule 
    \multirow{3}{*}{Large animals} & BRD & $48.94 \pm 0.48$ & $63.79 \pm 0.43$ & $62.04 \pm 0.37$ & $67.99 \pm 0.36$ & $58.02 \pm 0.41$ \\
    &      DOG & $29.73 \pm 0.31$ & $35.63 \pm 0.32$ & $40.19 \pm 0.32$ & $40.05 \pm 0.32$ & $36.24 \pm 0.31$ \\
    &      AWA & $33.87 \pm 0.38$ & $51.84 \pm 0.49$ & $54.40 \pm 0.43$ & $58.49 \pm 0.46$ & $50.15 \pm 0.48$ \\
    \midrule 
    \multirow{3}{*}{Small animals}     &      PLK & $56.86 \pm 0.54$ & $65.72 \pm 0.49$ & $61.97 \pm 0.41$ & $76.19 \pm 0.41$ & $63.71 \pm 0.52$ \\
    &    INS\_2 & $27.98 \pm 0.31$ & $42.31 \pm 0.35$ & $43.40 \pm 0.35$ & $44.28 \pm 0.36$ & $40.70 \pm 0.36$ \\
    & INS & $27.75 \pm 0.30$ & $43.22 \pm 0.39$ & $46.51 \pm 0.37$ & $49.40 \pm 0.40$ & $43.69 \pm 0.38$ \\
    \midrule 
    \multirow{3}{*}{Plants} & FLW & $56.41 \pm 0.52$ & $69.22 \pm 0.51$ & $68.95 \pm 0.45$ & $75.35 \pm 0.43$ & $71.06 \pm 0.46$ \\
    &  PLT\_NET & $31.26 \pm 0.31$ & $47.83 \pm 0.38$ & $49.95 \pm 0.34$ & $47.25 \pm 0.34$ & $48.10 \pm 0.35$ \\
    &      FNG & $26.39 \pm 0.27$ & $35.44 \pm 0.34$ & $39.99 \pm 0.34$ & $42.02 \pm 0.34$ & $36.88 \pm 0.34$ \\
    \midrule 
    \multirow{3}{*}{Plant diseases} & PLT\_VIL & $60.83 \pm 0.52$ & $72.18 \pm 0.48$ & $70.51 \pm 0.39$ & $78.77 \pm 0.39$ & $73.50 \pm 0.43$ \\
    &   MED\_LF & $77.76 \pm 0.46$ & $85.50 \pm 0.38$ & $77.28 \pm 0.38$ & $88.53 \pm 0.29$ & $83.06 \pm 0.38$ \\
    &  PLT\_DOC & $28.05 \pm 0.28$ & $43.92 \pm 0.36$ & $48.06 \pm 0.34$ & $54.50 \pm 0.38$ & $42.12 \pm 0.33$ \\
    \midrule 
    \multirow{3}{*}{Microscopy} & BCT & $46.74 \pm 0.47$ & $52.04 \pm 0.46$ & $58.27 \pm 0.44$ & $58.56 \pm 0.47$ & $58.92 \pm 0.54$ \\
    &      PNU & $26.19 \pm 0.25$ & $26.61 \pm 0.24$ & $29.93 \pm 0.27$ & $29.88 \pm 0.25$ & $30.87 \pm 0.26$ \\
    &      PRT & $26.93 \pm 0.28$ & $28.71 \pm 0.26$ & $34.81 \pm 0.27$ & $35.98 \pm 0.28$ & $33.15 \pm 0.27$ \\
    \midrule 
    \multirow{3}{*}{Remote sensing} & RESISC & $42.93 \pm 0.44$ & $62.09 \pm 0.44$ & $61.86 \pm 0.42$ & $64.37 \pm 0.40$ & $60.75 \pm 0.44$ \\
    &    RSICB & $57.97 \pm 0.54$ & $86.17 \pm 0.35$ & $82.72 \pm 0.34$ & $86.73 \pm 0.32$ & $83.35 \pm 0.40$ \\
    &      RSD & $39.29 \pm 0.40$ & $68.37 \pm 0.45$ & $68.91 \pm 0.41$ & $76.32 \pm 0.40$ & $65.66 \pm 0.43$ \\
    \midrule 
    \multirow{3}{*}{Vehicles} &      CRS & $31.53 \pm 0.36$ & $35.88 \pm 0.37$ & $36.38 \pm 0.35$ & $42.31 \pm 0.37$ & $35.64 \pm 0.36$ \\
    & APL & $43.64 \pm 0.53$ & $55.19 \pm 0.45$ & $48.82 \pm 0.37$ & $62.64 \pm 0.40$ & $45.55 \pm 0.39$ \\
    &      BTS & $27.31 \pm 0.29$ & $36.60 \pm 0.32$ & $36.96 \pm 0.30$ & $46.13 \pm 0.33$ & $35.53 \pm 0.28$ \\
    \midrule 
    \multirow{3}{*}{Manufacturing} & TEX & $70.89 \pm 0.60$ & $82.78 \pm 0.51$ & $86.10 \pm 0.37$ & $89.39 \pm 0.31$ & $88.69 \pm 0.37$ \\
    &  TEX\_DTD & $26.78 \pm 0.29$ & $51.60 \pm 0.38$ & $44.25 \pm 0.35$ & $50.70 \pm 0.37$ & $42.53 \pm 0.36$ \\
    & TEX\_ALOT & $69.64 \pm 0.56$ & $95.71 \pm 0.22$ & $94.65 \pm 0.22$ & $97.73 \pm 0.13$ & $94.94 \pm 0.24$ \\
    \midrule 
    \multirow{3}{*}{Human actions}     &      SPT & $46.76 \pm 0.44$ & $58.44 \pm 0.42$ & $55.69 \pm 0.37$ & $59.35 \pm 0.39$ & $55.40 \pm 0.41$ \\
    &   ACT\_40 & $29.43 \pm 0.30$ & $39.43 \pm 0.35$ & $41.35 \pm 0.36$ & $39.34 \pm 0.35$ & $39.66 \pm 0.35$ \\
    & ACT\_410 & $40.34 \pm 0.42$ & $54.19 \pm 0.42$ & $56.65 \pm 0.38$ & $64.87 \pm 0.37$ & $55.20 \pm 0.41$ \\
    \midrule 
    \multirow{3}{*}{OCR} & MD\_MIX & $19.54 \pm 0.16$ & $52.72 \pm 0.41$ & $20.39 \pm 0.22$ & $46.29 \pm 0.37$ & $18.30 \pm 0.19$ \\
    & MD\_5\_BIS & $20.32 \pm 0.17$ & $40.75 \pm 0.37$ & $25.81 \pm 0.29$ & $56.00 \pm 0.37$ & $22.34 \pm 0.23$ \\
    &     MD\_6 & $21.78 \pm 0.20$ & $70.69 \pm 0.40$ & $44.25 \pm 0.37$ & $64.42 \pm 0.39$ & $26.73 \pm 0.26$ \\
    \bottomrule
\end{tabular}
\end{adjustbox}
\end{table}

\begin{table}[!ht]
    \centering
    \caption{Full performance results for cross-domain any-way any-shot image classification on all datasets. The 95\% confidence intervals are computed at per-task level over 3 runs per dataset with 600 tasks per run (total tasks: $3 \times 600=1\,800$).}
    \label{tab:cdfslAnyshot}
     \begin{adjustbox}{width=\linewidth}
    \begin{tabular}{lllllll}
    \toprule
    Domain & Dataset ID  &           TrainFromScratch &       Finetuning &      MatchingNet &         ProtoNet &             FO-MAML \\
    \midrule 
    \multirow{3}{*}{Large animals} & BRD & $29.95 \pm 0.92$ & $37.96 \pm 0.91$ & $43.08 \pm 0.76$ & $51.93 \pm 0.72$ & $34.40 \pm 0.82$ \\
    &      DOG & $19.58 \pm 0.73$ & $22.95 \pm 0.76$ & $25.76 \pm 0.71$ & $27.05 \pm 0.72$ & $22.69 \pm 0.70$ \\
    &      AWA & $21.53 \pm 0.74$ & $33.64 \pm 0.83$ & $38.01 \pm 0.73$ & $42.67 \pm 0.76$ & $35.19 \pm 0.73$ \\
    \midrule 
    \multirow{3}{*}{Small animals}     &      PLK & $36.18 \pm 0.97$ & $47.26 \pm 0.90$ & $42.89 \pm 0.80$ & $52.86 \pm 0.73$ & $45.29 \pm 0.84$ \\
    &    INS\_2 & $18.01 \pm 0.70$ & $27.07 \pm 0.77$ & $28.12 \pm 0.71$ & $29.37 \pm 0.70$ & $25.22 \pm 0.73$ \\
    & INS & $17.88 \pm 0.69$ & $25.93 \pm 0.78$ & $29.86 \pm 0.71$ & $33.81 \pm 0.72$ & $27.91 \pm 0.71$ \\
    \midrule 
    \multirow{3}{*}{Plants} & FLW & $34.11 \pm 0.98$ & $42.86 \pm 0.90$ & $51.66 \pm 0.75$ & $60.11 \pm 0.72$ & $46.61 \pm 0.85$ \\
    &  PLT\_NET & $20.30 \pm 0.74$ & $30.94 \pm 0.82$ & $33.53 \pm 0.73$ & $32.49 \pm 0.74$ & $30.26 \pm 0.75$ \\
    &      FNG & $17.58 \pm 0.67$ & $23.26 \pm 0.75$ & $26.94 \pm 0.70$ & $28.65 \pm 0.71$ & $25.05 \pm 0.69$ \\
    \midrule 
    \multirow{3}{*}{Plant diseases} & PLT\_VIL & $38.25 \pm 1.02$ & $49.60 \pm 0.97$ & $51.78 \pm 0.77$ & $61.95 \pm 0.72$ & $50.69 \pm 0.92$ \\
    &   MED\_LF & $54.99 \pm 0.93$ & $63.73 \pm 0.84$ & $62.36 \pm 0.63$ & $75.24 \pm 0.55$ & $58.56 \pm 0.86$ \\
    &  PLT\_DOC & $18.06 \pm 0.67$ & $27.55 \pm 0.77$ & $30.39 \pm 0.70$ & $39.09 \pm 0.73$ & $28.13 \pm 0.70$ \\
    \midrule 
    \multirow{3}{*}{Microscopy} & BCT & $29.68 \pm 0.95$ & $35.47 \pm 0.90$ & $41.98 \pm 0.80$ & $42.60 \pm 0.78$ & $34.13 \pm 0.87$ \\
    &      PNU & $17.70 \pm 0.64$ & $18.50 \pm 0.64$ & $21.52 \pm 0.62$ & $19.42 \pm 0.60$ & $18.47 \pm 0.61$ \\
    &      PRT & $17.80 \pm 0.70$ & $18.10 \pm 0.66$ & $22.03 \pm 0.67$ & $22.10 \pm 0.68$ & $21.40 \pm 0.67$ \\
    \midrule 
    \multirow{3}{*}{Remote sensing} & RESISC & $27.05 \pm 0.87$ & $39.70 \pm 0.86$ & $43.55 \pm 0.73$ & $49.34 \pm 0.71$ & $39.34 \pm 0.80$ \\
    &    RSICB & $36.19 \pm 1.03$ & $67.64 \pm 0.74$ & $66.59 \pm 0.65$ & $70.96 \pm 0.62$ & $65.20 \pm 0.76$ \\
    &      RSD & $25.61 \pm 0.88$ & $46.74 \pm 0.90$ & $50.28 \pm 0.75$ & $59.66 \pm 0.73$ & $47.44 \pm 0.77$ \\
    \midrule 
    \multirow{3}{*}{Vehicles} &      CRS & $20.10 \pm 0.72$ & $22.36 \pm 0.74$ & $24.12 \pm 0.70$ & $27.77 \pm 0.70$ & $21.02 \pm 0.69$ \\
    & APL & $26.67 \pm 0.91$ & $37.08 \pm 0.86$ & $31.05 \pm 0.70$ & $40.28 \pm 0.73$ & $27.91 \pm 0.78$ \\
    &      BTS & $17.96 \pm 0.68$ & $24.09 \pm 0.76$ & $24.28 \pm 0.69$ & $30.36 \pm 0.71$ & $21.75 \pm 0.67$ \\
    \midrule 
    \multirow{3}{*}{Manufacturing} & TEX & $45.63 \pm 1.11$ & $66.63 \pm 0.82$ & $73.21 \pm 0.60$ & $79.37 \pm 0.53$ & $66.78 \pm 0.82$ \\
    &  TEX\_DTD & $17.84 \pm 0.65$ & $34.59 \pm 0.78$ & $30.79 \pm 0.69$ & $32.95 \pm 0.70$ & $28.23 \pm 0.68$ \\
    & TEX\_ALOT & $44.15 \pm 1.11$ & $83.10 \pm 0.56$ & $85.45 \pm 0.43$ & $91.05 \pm 0.39$ & $81.62 \pm 0.74$ \\
    \midrule 
    \multirow{3}{*}{Human actions}     &      SPT & $28.81 \pm 0.89$ & $35.82 \pm 0.89$ & $37.01 \pm 0.75$ & $43.60 \pm 0.74$ & $35.15 \pm 0.82$ \\
    &   ACT\_40 & $19.11 \pm 0.70$ & $24.74 \pm 0.77$ & $27.81 \pm 0.70$ & $27.96 \pm 0.70$ & $23.55 \pm 0.68$ \\
    & ACT\_410 & $25.03 \pm 0.80$ & $33.79 \pm 0.83$ & $40.50 \pm 0.70$ & $46.14 \pm 0.70$ & $36.83 \pm 0.73$ \\
    \midrule 
    \multirow{3}{*}{OCR} & MD\_MIX & $13.42 \pm 0.54$ & $37.76 \pm 0.79$ & $12.99 \pm 0.55$ & $14.00 \pm 0.56$ & $12.88 \pm 0.55$ \\
    & MD\_5\_BIS & $13.73 \pm 0.54$ & $28.78 \pm 0.77$ & $15.34 \pm 0.58$ & $24.55 \pm 0.77$ & $14.73 \pm 0.57$ \\
    &     MD\_6 & $15.16 \pm 0.59$ & $50.02 \pm 0.92$ & $20.00 \pm 0.65$ & $46.12 \pm 0.82$ & $16.85 \pm 0.60$ \\
    \bottomrule
\end{tabular}
\end{adjustbox}
\end{table}

\begin{table}[!ht]
\centering
\caption{Full performance results for cross-domain any-way any-shot image classification per number of ways. The 95\% confidence intervals are computed at per-task level using the number of task shown in the table.}
\label{tab:cdfslWays}
 \begin{adjustbox}{width=\linewidth}
\begin{tabular}{lllllll}
\toprule
 \textit{N}-way &  Evaluated Tasks & TrainFromScratch &      Finetuning &     MatchingNet &        ProtoNet &         FO-MAML \\
\midrule
      2 &             2\,925 &  $0.70 \pm 0.01$ & $0.78 \pm 0.01$ & $0.73 \pm 0.01$ & $0.77 \pm 0.01$ & $0.72 \pm 0.01$ \\
      3 &             2\,708 &  $0.55 \pm 0.01$ & $0.66 \pm 0.01$ & $0.61 \pm 0.01$ & $0.65 \pm 0.01$ & $0.58 \pm 0.01$ \\
      4 &             2\,910 &  $0.46 \pm 0.01$ & $0.58 \pm 0.01$ & $0.54 \pm 0.01$ & $0.59 \pm 0.01$ & $0.52 \pm 0.01$ \\
      5 &             2\,704 &  $0.38 \pm 0.01$ & $0.53 \pm 0.01$ & $0.48 \pm 0.01$ & $0.54 \pm 0.01$ & $0.46 \pm 0.01$ \\
      6 &             2\,856 &  $0.33 \pm 0.01$ & $0.48 \pm 0.01$ & $0.44 \pm 0.01$ & $0.50 \pm 0.01$ & $0.43 \pm 0.01$ \\
      7 &             2\,683 &  $0.30 \pm 0.01$ & $0.44 \pm 0.01$ & $0.42 \pm 0.01$ & $0.47 \pm 0.01$ & $0.40 \pm 0.01$ \\
      8 &             3\,025 &  $0.26 \pm 0.00$ & $0.40 \pm 0.01$ & $0.39 \pm 0.01$ & $0.45 \pm 0.01$ & $0.37 \pm 0.01$ \\
      9 &             2\,864 &  $0.24 \pm 0.00$ & $0.37 \pm 0.01$ & $0.37 \pm 0.01$ & $0.43 \pm 0.01$ & $0.35 \pm 0.01$ \\
     10 &             2\,943 &  $0.22 \pm 0.00$ & $0.35 \pm 0.01$ & $0.35 \pm 0.01$ & $0.41 \pm 0.01$ & $0.33 \pm 0.01$ \\
     11 &             2\,892 &  $0.19 \pm 0.00$ & $0.33 \pm 0.01$ & $0.33 \pm 0.01$ & $0.40 \pm 0.01$ & $0.31 \pm 0.01$ \\
     12 &             2\,983 &  $0.18 \pm 0.00$ & $0.31 \pm 0.01$ & $0.32 \pm 0.01$ & $0.38 \pm 0.01$ & $0.29 \pm 0.01$ \\
     13 &             3\,076 &  $0.17 \pm 0.00$ & $0.29 \pm 0.01$ & $0.31 \pm 0.01$ & $0.37 \pm 0.01$ & $0.28 \pm 0.01$ \\
     14 &             2\,952 &  $0.15 \pm 0.00$ & $0.28 \pm 0.01$ & $0.30 \pm 0.01$ & $0.36 \pm 0.01$ & $0.26 \pm 0.01$ \\
     15 &             2\,811 &  $0.14 \pm 0.00$ & $0.26 \pm 0.01$ & $0.29 \pm 0.01$ & $0.35 \pm 0.01$ & $0.25 \pm 0.01$ \\
     16 &             2\,682 &  $0.13 \pm 0.00$ & $0.24 \pm 0.01$ & $0.27 \pm 0.01$ & $0.33 \pm 0.01$ & $0.23 \pm 0.01$ \\
     17 &             2\,842 &  $0.12 \pm 0.00$ & $0.24 \pm 0.01$ & $0.27 \pm 0.01$ & $0.33 \pm 0.01$ & $0.22 \pm 0.01$ \\
     18 &             2\,779 &  $0.12 \pm 0.00$ & $0.23 \pm 0.01$ & $0.26 \pm 0.01$ & $0.32 \pm 0.01$ & $0.21 \pm 0.01$ \\
     19 &             2\,711 &  $0.11 \pm 0.00$ & $0.21 \pm 0.01$ & $0.25 \pm 0.01$ & $0.31 \pm 0.01$ & $0.20 \pm 0.01$ \\
     20 &             2\,654 &  $0.10 \pm 0.00$ & $0.21 \pm 0.01$ & $0.25 \pm 0.01$ & $0.31 \pm 0.01$ & $0.19 \pm 0.01$ \\
\bottomrule
\end{tabular}
\end{adjustbox}
\end{table}

\begin{table}[!ht]
\centering
\caption{Full performance results for cross-domain any-way any-shot image classification per number of shots. The 95\% confidence intervals are computed at per-task level using the number of task shown in the table.}
\label{tab:cdfslShots}
 \begin{adjustbox}{width=\linewidth}
\begin{tabular}{lllllll}
\toprule
 \textit{k}-shot &  Evaluated Tasks & TrainFromScratch &      Finetuning &     MatchingNet &        ProtoNet &         FO-MAML \\
\midrule
       1 &             2\,635 &  $0.21 \pm 0.01$ & $0.30 \pm 0.01$ & $0.29 \pm 0.01$ & $0.30 \pm 0.01$ & $0.25 \pm 0.01$ \\
       2 &             2\,577 &  $0.24 \pm 0.01$ & $0.35 \pm 0.01$ & $0.33 \pm 0.01$ & $0.35 \pm 0.01$ & $0.29 \pm 0.01$ \\
       3 &             2\,571 &  $0.25 \pm 0.01$ & $0.36 \pm 0.01$ & $0.34 \pm 0.01$ & $0.38 \pm 0.01$ & $0.31 \pm 0.01$ \\
       4 &             2\,803 &  $0.26 \pm 0.01$ & $0.37 \pm 0.01$ & $0.36 \pm 0.01$ & $0.40 \pm 0.01$ & $0.33 \pm 0.01$ \\
       5 &             2\,737 &  $0.26 \pm 0.01$ & $0.38 \pm 0.01$ & $0.37 \pm 0.01$ & $0.41 \pm 0.01$ & $0.34 \pm 0.01$ \\
       6 &             2\,748 &  $0.25 \pm 0.01$ & $0.37 \pm 0.01$ & $0.36 \pm 0.01$ & $0.41 \pm 0.01$ & $0.33 \pm 0.01$ \\
       7 &             2\,761 &  $0.26 \pm 0.01$ & $0.38 \pm 0.01$ & $0.37 \pm 0.01$ & $0.43 \pm 0.01$ & $0.34 \pm 0.01$ \\
       8 &             2\,761 &  $0.25 \pm 0.01$ & $0.38 \pm 0.01$ & $0.37 \pm 0.01$ & $0.43 \pm 0.01$ & $0.34 \pm 0.01$ \\
       9 &             2\,718 &  $0.26 \pm 0.01$ & $0.39 \pm 0.01$ & $0.38 \pm 0.01$ & $0.44 \pm 0.01$ & $0.35 \pm 0.01$ \\
      10 &             2\,791 &  $0.27 \pm 0.01$ & $0.40 \pm 0.01$ & $0.40 \pm 0.01$ & $0.46 \pm 0.01$ & $0.37 \pm 0.01$ \\
      11 &             2\,850 &  $0.26 \pm 0.01$ & $0.39 \pm 0.01$ & $0.39 \pm 0.01$ & $0.45 \pm 0.01$ & $0.36 \pm 0.01$ \\
      12 &             2\,747 &  $0.26 \pm 0.01$ & $0.39 \pm 0.01$ & $0.39 \pm 0.01$ & $0.46 \pm 0.01$ & $0.36 \pm 0.01$ \\
      13 &             2\,580 &  $0.26 \pm 0.01$ & $0.39 \pm 0.01$ & $0.40 \pm 0.01$ & $0.46 \pm 0.01$ & $0.37 \pm 0.01$ \\
      14 &             2\,607 &  $0.26 \pm 0.01$ & $0.39 \pm 0.01$ & $0.40 \pm 0.01$ & $0.46 \pm 0.01$ & $0.37 \pm 0.01$ \\
      15 &             2\,683 &  $0.26 \pm 0.01$ & $0.39 \pm 0.01$ & $0.40 \pm 0.01$ & $0.47 \pm 0.01$ & $0.37 \pm 0.01$ \\
      16 &             2\,616 &  $0.27 \pm 0.01$ & $0.40 \pm 0.01$ & $0.40 \pm 0.01$ & $0.47 \pm 0.01$ & $0.37 \pm 0.01$ \\
      17 &             2\,632 &  $0.27 \pm 0.01$ & $0.41 \pm 0.01$ & $0.41 \pm 0.01$ & $0.48 \pm 0.01$ & $0.38 \pm 0.01$ \\
      18 &             2\,703 &  $0.25 \pm 0.01$ & $0.38 \pm 0.01$ & $0.40 \pm 0.01$ & $0.47 \pm 0.01$ & $0.36 \pm 0.01$ \\
      19 &             2\,753 &  $0.25 \pm 0.01$ & $0.39 \pm 0.01$ & $0.40 \pm 0.01$ & $0.47 \pm 0.01$ & $0.37 \pm 0.01$ \\
      20 &             2\,727 &  $0.25 \pm 0.01$ & $0.39 \pm 0.01$ & $0.40 \pm 0.01$ & $0.47 \pm 0.01$ & $0.37 \pm 0.01$ \\
\bottomrule
\end{tabular}
\end{adjustbox}
\end{table}

\begin{table}[b]
\centering
\caption{The running times for all techniques on all cross-domain settings over all datasets. The number of tasks in each phase is: 540\,000 during meta-training, 129\,600 during meta-validation, and 54\,000 during meta-testing.}
\label{tab:cdfslTime}
 \begin{adjustbox}{width=\linewidth}
\begin{tabular}{lllllll}
\toprule
Setting &           Phase & TrainFromScratch & Finetuning & MatchingNet & ProtoNet & FO-MAML \\
\midrule
\multirow{3}{*}{5-way 1-shot} &      Meta-train &             0h0m &      3h23m &       6h33m &    6h47m &  17h40m \\
& Meta-validation &             0h0m &     24h53m &       2h48m &    2h45m &   8h52m \\
&       Meta-test &           29h29m &      9h21m &       0h18m &    0h19m &   2h48m \\
\midrule
\multirow{3}{*}{5-way 5-shot} &      Meta-train &             0h0m &      3h30m &       7h13m &     7h0m &   18h2m \\
 & Meta-validation &             0h0m &     25h41m &       3h24m &    3h16m &   9h32m \\
 &       Meta-test &           29h24m &      9h22m &       0h19m &    0h20m &   2h58m \\
\midrule
\multirow{3}{*}{5-way 10-shot} &      Meta-train &             0h0m &      3h25m &       7h25m &    7h10m &  19h28m \\
 & Meta-validation &             0h0m &     25h45m &        4h2m &    3h55m &  10h30m \\
 &       Meta-test &            30h7m &      9h26m &       0h20m &    0h21m &   3h16m \\
\midrule
\multirow{3}{*}{5-way 20-shot} &      Meta-train &             0h0m &      3h24m &        8h3m &    8h20m &  22h10m \\
 & Meta-validation &             0h0m &     27h11m &       5h21m &    5h23m &  13h18m \\
 &       Meta-test &           29h56m &      9h22m &       0h24m &    0h25m &   3h39m \\
\midrule
\multirow{3}{*}{any-way any-shot} &      Meta-train &             0h0m &      3h20m &        7h4m &    7h59m &  19h37m \\
 & Meta-validation &             0h0m &     30h11m &       8h27m &    8h58m &  17h21m \\
 &       Meta-test &            30h2m &      9h32m &       0h32m &    0h34m &   4h19m \\
\bottomrule
\end{tabular}
\end{adjustbox}
\end{table}

\clearpage
\newpage
\section{Benchmarks/Datasets comparison}
\label{appendix:benchmark_comparison}

We compare the \textbf{Meta-Album meta-dataset} with other (meta-)datasets in 
Table~\ref{tab:datasets_comparison_2}. Each row shows a (meta-)dataset while each column shows features. The last three rows show versions of the newly proposed meta-dataset, \ie Meta-Album. 
The table describes the following features, divided in quantitative and qualitative features:

\subsection*{Quantitative features}
\begin{itemize}
  \item number of domains
  \item number of datasets
  \item number of total images
  \item min/max number of classes per domain
  \item min/max number of images per class
  \item size on disk (in GB)
\end{itemize}

\subsection*{Qualitative features}
\begin{itemize}
  \item Whether the (meta-)datasets has datasets from multiple domains
  \item Whether the (meta-)dataset is lightweight (easy to store on most computers, \ie < 20 GB)
  \item Whether the (meta-)dataset has a uniform number of images per class
  \item Whether the (meta-)dataset has images of uniform size
  \item Whether the (meta-)dataset is designed to be extended with more datasets or domains. 
\end{itemize}

\begin{table}[b]
  \caption{Feature comparison between Meta-Album and other large-scale or (meta-)datasets
  }
 
  \label{tab:datasets_comparison_2}
  \centering
  \begin{adjustbox}{width=\linewidth}
  \begin{tabular}{l r r r r r l c c c c c}
    \toprule
    \makecell{Dataset/\\Meta-Dataset}  
    & \rot{\# of domains} & \rot{\# of datasets}  
    & \rot{\# of images}  & \rot{\makecell{min/max classes\\per domain}}  
    & \rot{\makecell{min/max images\\per class}} & \rot{size on disk} 
    & \rot{multi-domain} & \rot{\makecell{lightweight\\($<$20GB)}} 
    & \rot{\makecell{uniform \# of\\images per class}} 
    & \rot{\makecell{uniform\\image size}} & \rot{extendable} 
    \\
    \midrule
    
    Meta-Dataset    & 7 & 10 & 53\,068\,000       
    & 43/1\,696 & 3/140\,000 & 210 GB & \cmark & \xmark & \xmark & \xmark  & \xmark
    \\
    VTAB    & 3 & 19 & 2\,244\,000     & 2/397       
    & 40/1\,000 & 100 GB & \cmark & \xmark & \xmark & \xmark & \xmark
    \\
    MS-COCO & 1 & 1 & 328\,000 & 80/80 & 9/10\,777 & 44 GB
    & \xmark & \xmark & \xmark & \xmark & \xmark
    \\
    Mini Imagenet   & 1 & 1 & 60\,000 & 100/100 & 600/600   
    & 1 GB & \xmark & \cmark & \cmark & \cmark & \xmark
    \\
    Omniglot    & 1 & 1 & 32\,000    & 1\,623/1\,623 & 20/20      
    & 148 MB & \xmark & \cmark & \cmark & \cmark & \xmark
    \\
    CUB-200  & 1 & 1 & 6\,000         & 200/200 & 20/39    
    & 647 MB & \xmark & \cmark & \xmark & \xmark & \xmark
    \\
    CIFAR-100   & 3 & 1 & 60\,000        & 15/50 & 600/600   
    & 161 MB & \xmark & \cmark & \cmark & \cmark & \xmark
    \\
    
    \midrule
    \textbf{Meta-Album {\it Micro}} & \textbf{10} & \textbf{40} & \textbf{32\,000} & \textbf{19/20} & \textbf{40/40} 
    & \textbf{380 MB} & \cmark & \cmark & \cmark & \cmark & \cmark
    \\
    \textbf{Meta-Album {\it Mini}} & \textbf{10} & \textbf{40} & \textbf{220\,950} & \textbf{19/706} & \textbf{40/40} 
    & \textbf{3.9 GB} & \cmark & \cmark & \cmark & \cmark & \cmark
    \\
    \textbf{Meta-Album {\it Extended}} & \textbf{10} & \textbf{40} & \textbf{1\,583\,624} & \textbf{19/706} & \textbf{1/187\,384} 
    & \textbf{15 GB} & \cmark & \cmark & \xmark & \cmark & \cmark
    \\
    \bottomrule
    
  \end{tabular}
  \end{adjustbox}
\end{table}

We discuss the number of domains and extensibility in more detail.

\subsection*{Dataset domains}
We could identify the following domains in Meta-Dataset, VTAB, and CIFAR-100. For other datasets, we have not been able to find an adequate description of the domains, and as such we refrain from this.

\paragraph{Meta-Dataset}
The meta-dataset is divided into total 7 domains as listed below:
\begin{enumerate}
  \item Organisms, \eg birds, fungi, flowers.
  \item Common objects, \eg from ImageNet, ILSVRC, COCO.
  \item Characters, \ie Omniglot.
  \item Textures, \ie DTD.
  \item Vehicles, \ie FGVC-Aircraft.
  \item Drawings, \ie Quick, Draw!
  \item Traffic signs, \ie the German Traffic Sign Recognition Benchmark.
\end{enumerate}

\paragraph{CIFAR-100}
The CIFAR-100 is divided into 3 domains:
\begin{enumerate}
  \item people, reptiles, small mammals, aquatic mammals, fish, insects, large carnivores, large omnivores and herbivores, medium-sized mammals, non-insect invertebrates
  \item trees, flowers, fruit and vegetables
  \item vehicles 1, vehicles 2, large natural outdoor scenes, large man-made outdoor things, household furniture, household electrical devices, food containers	
\end{enumerate}

\paragraph{VTAB}
VTAB is divided into 3 domains:
\begin{enumerate}
  \item \textit{Natural} image tasks include images of the natural world captured through standard cameras, representing generic objects, fine-grained classes, or abstract concepts.
  \item \textit{Specialized} tasks utilize images captured using specialist equipment, such as medical images or remote sensing. 
  \item \textit{Structured} tasks often derive from artificial environments that target understanding of specific changes between images, such as predicting the distance to an object in a 3D scene, counting objects, or detecting orientation.
\end{enumerate}

\subsection*{Extensibility}
Meta-Album is the only one among these datasets that has been designed explicitly to be extended. We are inviting the community to contribute more domains and datasets. We provide open-source code to prepare more datasets following the Meta-Album format, and we have a review process so that newly added datasets are checked for quality and subsequently included.

\clearpage
\newpage
\section*{Important Links}
\label{appenidx:links}
Meta-Album Website : \\\url{https://meta-album.github.io/}
\\~\\
Meta-Album GitHub repo : \\\url{https://github.com/ihsanullah2131/meta-album}
\\~\\
NeurIPS Cross-Domain MetaDL Competition 2022 : \\\url{https://github.com/DustinCarrion/cd-metadl}
\\~\\
NeurIPS MetaDL Challenge 2021 : \\\url{https://autodl.lri.fr/competitions/210}
\\~\\
MetaDL Self Service : \\\url{https://competitions.codalab.org/competitions/31280}
\\~\\
Factsheets repo : \\\url{https://github.com/ihsanullah2131/meta-album/tree/master/Factsheets}
\\~\\
Data Format : \\\url{https://github.com/ihsanullah2131/meta-album/tree/master/DataFormat}
\\~\\
Data Format Check : \\\url{https://github.com/ihsanullah2131/meta-album/blob/master/DataFormat/check_data_format.py}
\\~\\
Factsheet Report Script : \\\url{https://github.com/ihsanullah2131/meta-album/blob/master/Factsheets/generate_pdf_report.py}
\\~\\
Factsheet Report Template : \\\url{https://github.com/ihsanullah2131/meta-album/blob/master/Factsheets/template.html}
\\~\\

\subsection*{Contact}
For any query about the Meta-Album meta-dataset, reach us out by email \href{mailto:meta-album@chalearn.org}{meta-album@chalearn.org}.

\end{document}